\documentclass{article}
\usepackage{microtype}
\usepackage{graphicx}
\usepackage{subfigure}
\usepackage{booktabs} 
\usepackage{amsmath}
\usepackage{amssymb}
\usepackage{mathtools}
\usepackage{amsthm}
\usepackage{amsfonts}

\usepackage[backref=page,hidelinks]{hyperref}
\usepackage[capitalize,noabbrev]{cleveref}
\usepackage[accepted]{icml2023}
\usepackage[toc,page,header]{appendix}
\usepackage{minitoc}

\newcommand{\overbar}[1]{\mkern 1.2mu\overline{\mkern-1.2mu#1\mkern-1.2mu}\mkern 1.2mu}

\usepackage{color, colortbl}
\definecolor{LightCyan}{rgb}{0.88,1,1}

\theoremstyle{plain}
\newtheorem{theorem}{Theorem}[section]
\newtheorem{proposition}[theorem]{Proposition}
\newtheorem{lemma}[theorem]{Lemma}
\newtheorem{corollary}[theorem]{Corollary}
\theoremstyle{definition}

\theoremstyle{remark}

\usepackage{microtype}
\usepackage{amssymb,amsmath, amsthm}
\usepackage{wrapfig, blindtext}
\usepackage{booktabs} 
\usepackage{amssymb}
\usepackage{framed}
\usepackage{mathrsfs}

\usepackage{multirow}
\usepackage{enumerate,pifont,bigdelim,bbding}
\usepackage{multirow} 
\usepackage{bm}
\usepackage{cite}
\usepackage{tikz}
\definecolor{gray}{rgb}{0.9, 0.9, 0.9}
\usepackage{enumitem}

\usepackage[export]{adjustbox} 
\usepackage{tabularx}

\newcommand{\inclu}[0] {\ar@{^{(}->}}

\newcommand{\cA}{\mathcal{A}}

\newcommand{\KL}{\mathrm{KL}}

\newcommand{\E}{\mathbb E}

\newcommand{\BR}{\mathfrak{BR}}

\newcommand{\argmin}{\operatornamewithlimits{argmin}}
\newcommand{\argmax}{\operatornamewithlimits{argmax}}

\newcommand{\algo}{\texttt{STEERING}}

\newcommand{\cD}{\mathcal{D}}

\newcommand{\cH}{\mathcal{H}}

\newcommand{\cS}{{\mathcal{S}}}

\newcommand{\KSD}{\text{KSD}}
\newcommand{\DSD}{\text{DSD}}
\usepackage{amssymb,amsmath,amsthm}
\usepackage{xcolor}
\newcommand{\mat}[1]{\mathbf{#1}}

\usepackage{soul}

\usepackage{algorithm} 
\usepackage{algorithmic}

\usepackage{mathtools}

\icmltitlerunning{\hfill Stein
Information Directed Exploration for Model-Based
Reinforcement Learning\hfill\thepage}

\begin{document}
\doparttoc 
\faketableofcontents 

\twocolumn[
\icmltitle{{\algo}: Stein Information  Directed Exploration for \\Model-Based Reinforcement Learning}
\begin{icmlauthorlist}
\icmlauthor{Souradip Chakraborty}{umd}
\icmlauthor{Amrit Singh Bedi}{umd}
\icmlauthor{Alec Koppel}{amz}
\icmlauthor{Mengdi Wang}{princew}
\icmlauthor{Furong Huang}{umd}
\icmlauthor{Dinesh Manocha}{umd}
\end{icmlauthorlist}

\icmlaffiliation{umd}{Department of Computer Science, University of Maryland, College Park, USA. }
\icmlaffiliation{princew}{Department of Electrical Engineering, Princeton University/Deepmind, Princeton, NJ, USA}
\icmlaffiliation{amz}{JP Morgan Chase AI Research,  USA. }

\icmlcorrespondingauthor{Amrit Singh Bedi}{amritbd@umd.edu}

\icmlkeywords{Machine Learning, ICML}

\vskip 0.3in
]



\printAffiliationsAndNotice{} 

\begin{abstract}
    Directed Exploration is a crucial challenge in reinforcement learning (RL), especially when rewards are sparse. Information-directed sampling (IDS), which optimizes the information ratio, seeks to do so by augmenting regret with information gain. However, estimating information gain is computationally intractable or relies on restrictive assumptions which prohibit its use in many practical instances. In this work, we posit an alternative exploration incentive in terms of the integral probability metric (IPM) between a current estimate of the transition model and the unknown optimal, which under suitable conditions, can be computed in closed form with the kernelized Stein discrepancy (KSD). Based on KSD, we develop a novel algorithm \algo: \textbf{STE}in information dir\textbf{E}cted exploration for model-based \textbf{R}einforcement Learn\textbf{ING}. To enable its derivation, we develop fundamentally new variants of KSD for discrete conditional distributions. {We further establish that {\algo} archives sublinear Bayesian regret, improving upon prior learning rates of information-augmented MBRL.} Experimentally,  we show that the proposed algorithm is computationally affordable and outperforms several prior approaches.

\end{abstract}
  \section{Introduction} 

 Exploring effectively is a major challenge in reinforcement learning (RL), particularly when the rewards are sparse \citep{sparse_rwd1, achiam_lb}. Recent research using model-based reinforcement learning (MBRL) with intrinsic curiosity \citep{curiosity1, curiosity2, curiosity3} has offered a potential solution, but its theoretical understanding is relatively immature. On the other hand, posterior sampling reinforcement learning (PSRL) offers an efficient framework for balancing exploration and exploitation that is conceptually well-substantiated \citep{osband2013, osband2014model}. However, PSRL may struggle in scenarios where many trajectories are uninformative, due to, e.g., reward sparsity \citep{russo2014learning}.
     \begin{figure}[t]
        \centering
        \includegraphics[width=0.73\columnwidth,clip = true]{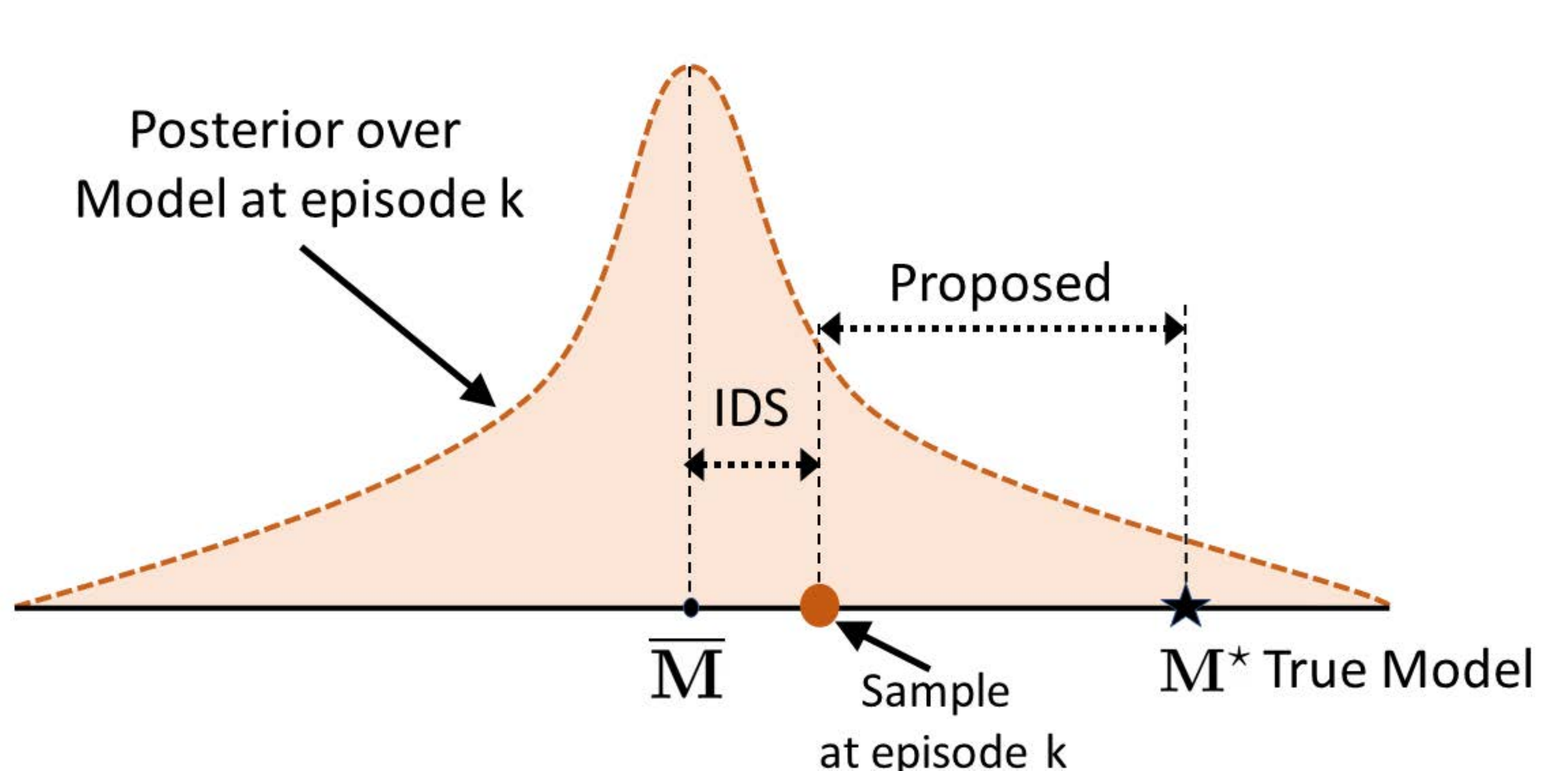}
        \caption{(Directed exploration) This figure illustrates that information-directed sampling (IDS) focuses on the distance between the current sample at episode $k$ and the mean of the posterior. In contrast, we focus on the distance to the true model.}
        \label{fig:conc_pp}
        \vspace{-0mm}
    \end{figure}
    
  To incentivize exploration in  MBRL, \citet{lu_info} proposed a design principle of information directed sampling (IDS), which optimizes the tradeoff between \emph{regret}  and \emph{information}. Tight information-theoretic Bayesian regret bounds for IDS are derived in \citep{lu_info, vanroy_bit}, under the specific choice of Dirichlet priors for the transition model, inspired by earlier work on bandits \citep{russo_info}.
  Recently,  \citet{hao2022regret} alleviated any requirements on the prior based upon the development of a surrogate environment estimation procedure via rate-distortion theory.
  
  Unfortunately, the existing IDS approaches {face two main challenges (1) they }are computationally intractable due to the need to estimate the information gain, {(2) they do not induce exploration directed towards the optimal true transition dynamics. By \emph{directed exploration}, we mean the algorithm moves in a direction toward the optimal transition dynamics rather than collecting all the information about the underlying environment, which is the focus of information gain-based exploration in IDS.} {The first challenge can be} partially addressed by instead optimizing the evidence lower-bound \citep{achiam_lb}, but ends up restricting focus to the posterior variance, which may be insufficiently informative about the underlying target distribution. 
        This motivates us to pose the following question:
        
        \emph{Can we develop a computationally tractable posterior sampling-based RL algorithm that exhibits efficient directed exploration {with provable guarantees}? }

        We provide an affirmative answer to this question by considering an alternative measure of information. That is, we propose \emph{Stein information gain}, which is the integral probability metric (IPM) difference between the estimated and true (unknown) transition dynamics \citep{sriperumbudur2012empirical}, {hence inducing \emph{directed exploration}}. Under the assumption that the transition model lies in the Stein class, we employ Stein's identity \citep{efron1973stein,james1992estimation} to evaluate this IPM between the true (unknown) and estimated transitions in closed-form using kernelized Stein discrepancy (KSD) \citep{gorham2015measuring, liu2016kernelized, hawkins2022online}. This is the key novelty that alleviates a major drawback of prior approaches that require evaluating mutual information. Thereby we introduce the Stein-information ratio, which may be seen as a modification of the information ratio in \citep{russo2018learning,vanroy_bit}, and incentivizes exploration. 
        We emphasize that our notion of        KSD-based Stein information gain empowers us to evaluate the distance to the true transition dynamics. Doing so permits us to derive the best-known prior-free information-theoretic Bayesian regret bounds.
        Towards this end, we also develop the first KSD for conditional discrete distributions and employ it in tabular RL settings. Appendix \ref{related_works} provides a detailed context of related works.

            \textbf{Contributions:} Our main contributions are as follows.
            \begin{list}{$\rhd$}{\topsep=0.ex \leftmargin=0.2in \rightmargin=0.in \itemsep =0.0in}
                \item We formalize the setting of model-based episodic RL with Bayesian regret incorporating a notion of distance to the ground-truth MDP with KSD, and hence achieves directed exploration towards the optima. 
                
                \item  We introduce discrete conditional KSD (DSD) in tabular RL for the first time and use it to analyze distributional distance, which empowers us to evaluate the exploration incentive towards the true transition dynamics. Through this definition, we introduce a specific exploration-incentivized modification of posterior sampling, called {\algo} (Algo. \ref{alg:MPC-PSRL}). 
                
                \item  We establish prior-free sublinear Bayesian regret bounds for {\algo} and provide certain regularity conditions on the RKHS under which the regret can be further improved.
                \item We provide extensive experimental evidence for the proposed {\algo} algorithm in sparse reward settings and show improvement via efficient directed exploration compared to all existing approaches. 
             \end{list}
	
  \section{Problem Formulation} 
   \begin{figure*}[t]
    \centering
    \hfil
	\subfigure[Dense rewards.]
        {\includegraphics[width=.31\textwidth,clip = true]{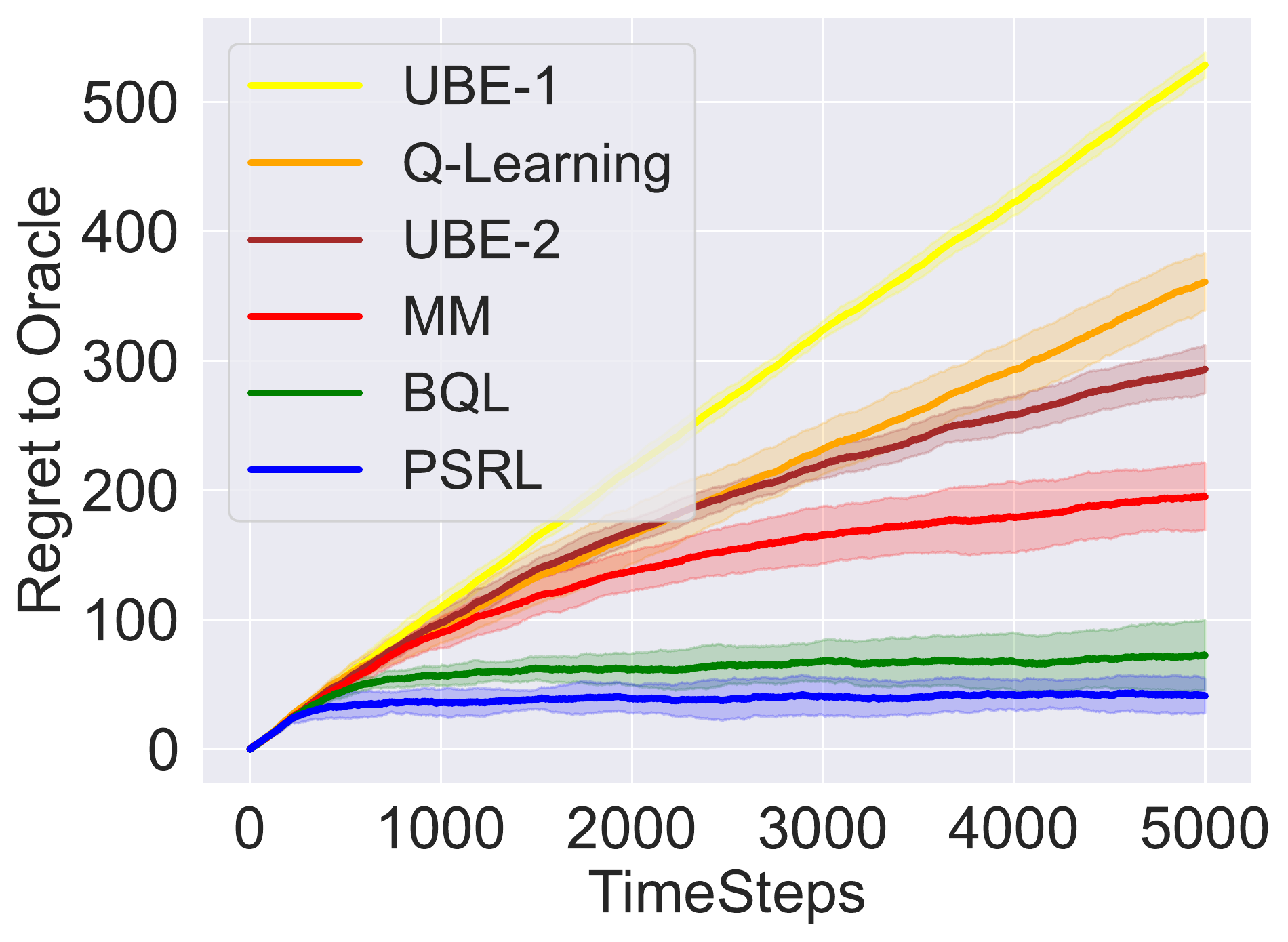}}
        \hfil
	\subfigure[Sparse rewards.]
        {\includegraphics[width=.31\textwidth,clip = true]{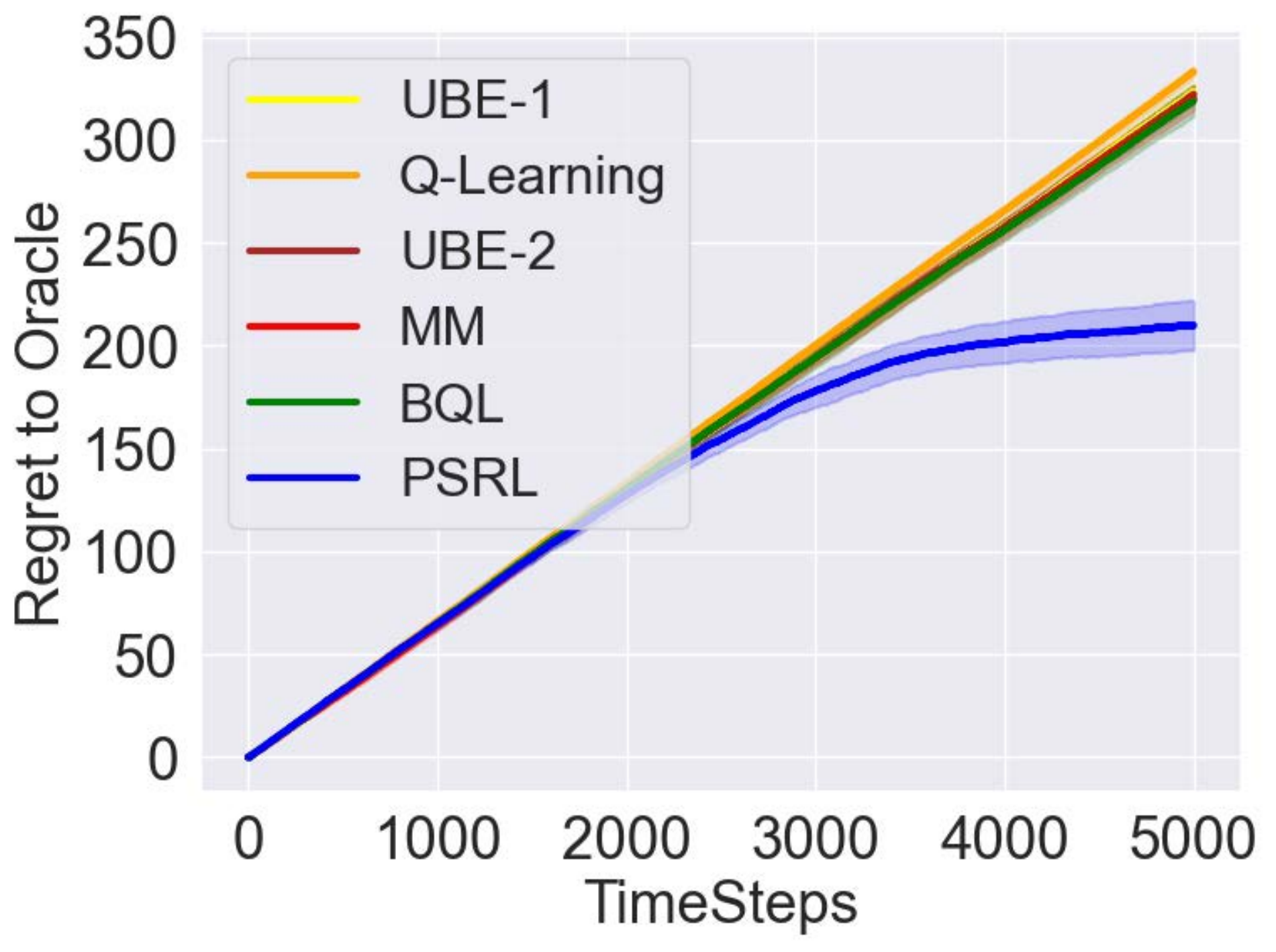}
  }
  \hfil
	\subfigure[Variable sparsity {(PSRL)}.]
    {\includegraphics[width=.31\textwidth,clip = true]{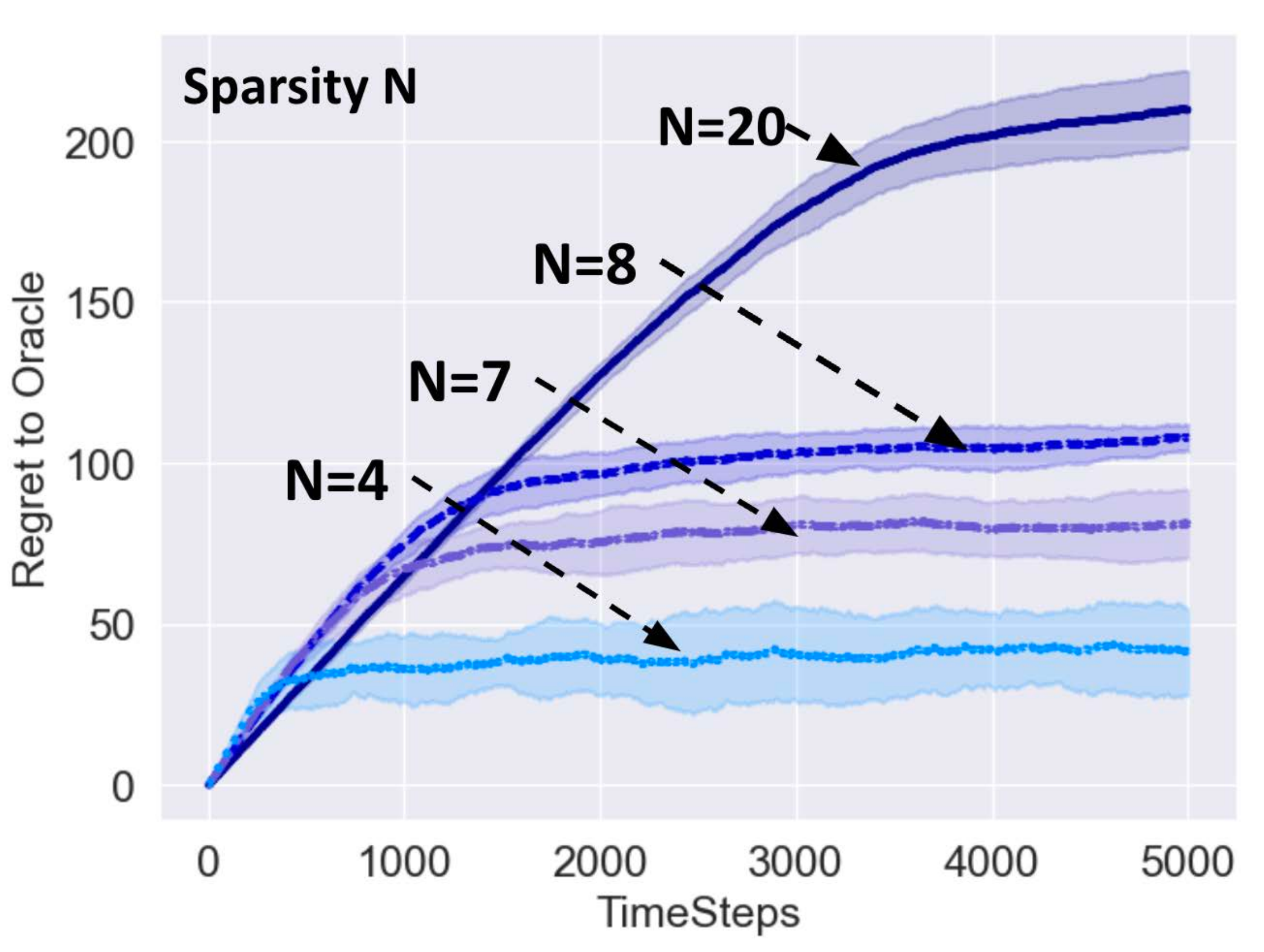}
    \label{fig:psrl_sparse}}
    \hfil
    \caption{This figure compares the performance of various RL algorithms for Dense reward (Fig. 2(a)) and Sparse reward (Fig. 2(b)) DeepSea environment (DSE) \citep{osband17a}. As we move from dense to sparse reward settings, we note that the regret becomes almost linear for all algorithms except PSRL \citep{osband2013}. We further show the performance degradation even for the PSRL algorithm with different sparsity levels in Fig. 2(c) achieved by varying $N$ (denotes the number of states in DSE).}
    \label{figure_sparse_rewards}
\end{figure*}

	  We consider the problem of learning transition dynamics in an episodic finite-horizon time-homogeneous tabular Markov Decision Process (MDP) setting. We define the unknown MDP as a random variable {$M :=\{\mathcal{S},\mathcal{A},R,P,H, R_{\text{max}},\rho\}$, where $\mathcal{S}$} is finite state-space with $S=|\mathcal{S}|$, $\mathcal{A}$ is the finite action space with $A=|\mathcal{A}|$, and $H$ is the episode length.	  Here, $P: \cS \times \cA \to \Delta_\cS$ represents the transition dynamics for the state-action transitions and {$R$  is the rewards distribution} where $\Delta_{\cS}$ denotes the set of probability distributions over a finite set $\cS$. After every episode of length $H$, state will reset according to the initial state distribution  $\rho$. At time step $i \in [H]$ within an episode, the agent observes state $s_i \in \mathcal{S}$, selects action $a_i \in \mathcal{A}$ according to a stochastic policy $\pi:\mathcal{S}\rightarrow \Delta_{\cA}$, receives reward  {$r_i \sim R(s_{i},a_i)$} and transitions to a new state {$s_{i+1} \sim P(\cdot|s_{i},a_i)$}. In this work, we consider $M$ as a random process, as is often the case in Bayesian RL \citep{russo2014learning}. 

   \textbf{Value Function and Bayesian Regret.} For a given MDP $M$, the value for time step $i$  is the reward accumulation during the episode, given by 
	  \begin{align}
		      V^{M}_{\pi,i}(s)=\mathbb{E}\Bigg[\sum_{j=i}^{H}[\bar{r}^M(s_{j},a_{j})|s_i = s, a_j\sim\pi(\cdot|s_j)]\Bigg],
		  \end{align}
	  where $j$ denotes the time-step within the episode  and $\bar{r}^M(s,a) = \mathbb{E}_{r \sim R^M(s,a)} [r]$. Without loss of generality, we assume $|\bar{r}^M(s,a)| \leq R_{\text{max}}$ for all $s\in \mathcal{S},a\in \mathcal{A}$, which implies that $|V(s)|\leq HR_{\text{max}}$, for all $s\in \mathcal{S}$. Next, for a given MDP $M$, an optimal policy $\pi^*$ is 
	  \begin{align}\label{optimal_policy}
		      \pi^* =  \argmax_{\pi} V^{M}_{\pi, 1}(s),
		  \end{align}
	  for all $s$ and $i\in [H]$. We emphasize that since $\pi^*$ is a function of $M$, it is also a random variable. An RL agent would interact with the environment over $K$ number of episodes with policy $\{\pi^k\}_{k=1}^{K}$. The performance of the learning agent with respect to environment $M$ can be quantified by {\textit{Bayesian regret}}:
	  \begin{align}\label{bayesian_Regret}
		    \BR_K:= \mathbb{E}\left[\sum_{k=1}^K ( {V_{1,{\pi}^*}^{{{M^*}}}(s_1^k)-V_{1,\pi^k}^{{M^*}}(s_1^k))} \right],
		  \end{align}
	  where the expectation is with respect to the randomness in the policy $\pi^k$ and the prior distribution of $M^*$, {where $M^*$ is the true MDP - a realization from the prior for an instantiation}. Typically, one focuses on ensuring the sublinear growth of Bayesian regret in \eqref{bayesian_Regret} as a way to quantify the learning performance of a given model-based estimate $M_k$ and the resultant policy $\pi^k:=\argmax_{\pi} V^{M_k}_{\pi}(s)$.  
   Posterior sampling-based reinforcement learning (PSRL) operates this way \citep{osband17a}, as does upper-confidence bound \citep{ayoub2020model}, and Gittin's index~\citep{https://doi.org/10.48550/arxiv.1909.05075}. However, regret [cf. \eqref{bayesian_Regret}] alone may under-perform in contexts where the expected value function is a sparse function of the initial state, which can occur when the reward function is sparse, which is true for different applications \citep{weerakoon_sparse,chakraborty2022dealing}.
   
   To emphasize this point, we consider the DeepSea environment of \citet{osband17a} and compare the performance of different existing methods in dense and sparse reward settings in Fig.~\ref{figure_sparse_rewards}. The performance degradation as we go from dense to sparse settings is evident (from Fig.~\ref{figure_sparse_rewards}(a) to Fig.~\ref{figure_sparse_rewards}(b)). We note that all the algorithms exhibit almost linear regret except PSRL, which establishes the efficient exploration aspect of PSRL. 
   But as we investigate PSRL further for different sparsity levels in Fig.~\ref{figure_sparse_rewards}(c), we conclude that even PSRL suffers badly {in very sparse reward settings}.  
   To deal with such challenges, information-directed sampling has been developed in the literature to jointly quantify value function sub-optimality with state space coverage.

	  \textbf{Information Directed Sampling (IDS)} augments PSRL by introducing the information ratio  at episode $k$ defined as 
	  \begin{align}\label{def:information_ratio_one}
		       {\Gamma_k(\pi, M^*):=\frac{\big(\mathbb E_k[V_{1,\pi^*}^{{M^*}}(s_1^k)-V_{1,\pi}^{M^*}(s_1^k)]\big)^2}{\mathbb I_k^{\pi}({M^*}; \cH_{k, H})}},
		  \end{align}
   whose numerator is the Bayesian regret and the denominator is the information gain as used in \citep{russo2014learning, hao2022regret}.
   The policy selection becomes $\pi_{\texttt{IDS}} =\arg\max_{\pi}\Gamma_k(\pi, M^*)$ which not only minimizes  regret but does so while maximizing per unit information gain, which can yield efficient exploration. Following \citet[Lemma A.1]{hao2022regret}, to gain further insight, it is useful to rewrite the information gain  as 
	  \begin{align}\label{informayion_gain_KL}
		    &\mathbb I_k^{\pi^{k}_{\texttt{TS}}}\left({M^*}; \cH_{k, H}\right)
		    \\
		    &\!\!\!=\!\!\sum_{h=1}^H \mathbb E_{k}\mathbb E_{\pi^{k}_{\texttt{TS}}}^{\overbar{M}_k}\left[D_{\text{KL}}\left(P^{M^*}(\cdot|s_h^k,a_h^k)||P^{\overbar{M}_k}(\cdot|s_h^k,a_h^k)\right)\right],  \nonumber
		  \end{align}
	  where $\mathbb E_{\pi^{k}_{\texttt{TS}}}^{\overbar{M}_k}$ is taken with respect to $s_h^k,a_h^k$, and $\mathbb E_{k}$ is with respect to $\pi$ and environment $M$. 
   The expression in \eqref{informayion_gain_KL} makes it clear that incorporating {$\mathbb I_k^{\pi}\left({M^*}; \cH_{k, H}\right)$} into the objective  motivates the agent to visit the state action region where {$D_{\KL}(P^{M^*}||{P}^{\overbar{M}_k})$} is higher which acts as an intrinsic reward to explore state action pairs with high uncertainty.

   \subsection{Limitations of IDS}\label{IDS_limitations}
The ratio objective in \eqref{def:information_ratio_one} exhibits practical limitations related to the fact that the KL divergence in \eqref{informayion_gain_KL} cannot be evaluated.  We next detail why this is so and how a modification can alleviate this issue.
    
	    \textbf{(L1) Computational Intractability:} A  major challenge in prior approaches including \citep{hao2022regret, lu_info} lies in an accurate estimation of mutual information. One of the first prior-free IDS analyses by \citet{hao2022regret} relies on constructing a covering set for KL divergence with cover radius $\epsilon=1/KH$. This implies that the information gain term grows unbounded as the number of episodes $K$ increases. Moreover, for practical implementation, to make the KL divergence in \eqref{informayion_gain_KL} tractable, \citet{hao2022regret} substitutes this quantity by its lower bound via Pinsker's inequality: $\mathbb E_{k}[D_{\KL}(P^{M^*}(\cdot|s_h^k,a_h^k)||{P}^ {\overbar{M}_k}(\cdot|s_h^k,a_h^k))] \geq \sum_{s'} \text{Var}(P^{M^*}(s'|s_h^k,a_h^k))$. 
        However, \citet{ozair_wasserstein} shows that any high-confidence lower bound requires exponential samples in the mutual information, which is a critical concern. Hence even the variance lower bound with Pinsker's inequality runs into the computational limits of estimating mutual information. 
            
	  \textbf{(L2) Not Truly A Directed Exploration:}  In the majority of the prior research on information-directed RL \citep{hao2022regret, lu_info}, the mutual information or KL divergence serves as the information-theoretic regularization or intrinsic curiosity to induce directed exploration to deal with the hard exploration challenges as detailed in \citep{thompson_tutorial}. However, as we note from \eqref{informayion_gain_KL}, since the mutual information must be substituted by the variance of the current estimate of the posterior distribution over $M$ for tractability purposes, it only encourages the agent to explore trajectories with high variance. Ideally, we would want our exploration to be directed towards the true MDP $M^*$ (see Fig. \ref{fig:conc_pp}) from which the data $(s,a,s')$ samples are collected in practice.  A directed exploration towards $M^*$ is crucial to avoid random wandering in the environment. For instance, consider a setting where we start with the strong belief prior, then, since the variance is already low, IDS based approach will not add any benefit on top of PSRL-based approaches. 

   To address the above limitations, we propose a  novel notion of \emph{Stein information gain} to achieve directed exploration in the next section. 
   	  \section{Proposed Approach and Algorithm}
		  {In this work, we  develop a novel Bayesian regret analysis that incorporates a notion of distance to the true optimal MDP and provides a computationally tractable alternative to the notion of information ratio in \citet{hao2022regret}. Before presenting the
proposed approach, let us discuss the technical development
as follows.
	\subsection{Kernelized Stein Discrepancy}\label{KSD_existing}
 Integral probability metrics (IPM) have gained traction in Bayesian inference and generative modeling \citep{arjovsky2017wasserstein}  for their ability to quantify the merit of a given posterior distribution with respect to an unknown target without specifically having knowledge of that target. In particular, when one suitably assumes the class of posteriors over which the search is conducted to the \emph{Stein class} \citep{liu2016kernelized}, IPMs admit a closed-form evaluation in terms of Stein discrepancies. 
   Please refer to Appendix \ref{KSD_existing_appenndix} for detailed discussion and derivation. To this end,  \citet{liu2016kernelized} define kernel Stein discrepancy (KSD) between two distributions $p$ and $q$ as 
	  \begin{align}\label{jordan_stein}
		   {\KSD}(p,q)  =  \mathbb E_{x,x'\sim p } \big [  u_q(x,x') \big],
		 \end{align}
	  where $ u_q(x,x')$ is the Stein kernel defined as 
	  \begin{align}\label{stein_kernel}
		  u_q & (x,x') := s_q(x)^\top \kappa(x,x') s_q(x')  +   s_q(x)^\top \nabla_{x'}\kappa(x,x')  
\nonumber
\\  &+\nabla_{x}\kappa(x,x')^\top   s_q(x') + trace(\nabla_{x,x'}\kappa(x,x')),
		  \end{align}
    where $\kappa(x,x')$ is the base kernel (any positive definite kernel, for instance, Hamming Kernel for discrete rvs)
    The Stein kernel in \eqref{stein_kernel} measures the similarity between two samples $x$ and $x'$, which comes from $p$, using the score function of $q$.
    For the setting in this work, we have $p=P^{M^*}$ (transition dynamics corresponding to true model $M^*$) and $q=P^{{M_k}}$ (transition dynamics corresponding to posterior $M_k \sim \phi(\cdot|\mathcal{H}_k)$). Interestingly, KSD empowers us to evaluate the distance $\KSD(P^{M_k}, P^{M^*})$.  In the next subsection, we present the main idea of this work. 
	\subsection{ Stein Information Directed Sampling}
	  Now, after the introduction of KSD, we note that the distributional distance to the unknown target can be computed in closed form. We use this fact to address the limitations of IDS discussed in Sec. \ref{IDS_limitations}, and  propose to replace information gain $\mathbb I_k^{\pi}({M^*}; \cH_{k, H})$ in the denominator of \eqref{def:information_ratio_one} with what we call \textit{Stein information gain} $\mathbb K_k^{\pi}\left(M^*; \cH_{k, H}\right)$ given by
    \begin{align}\label{informayion_gain_KSD}
		      &\mathbb K_k^{\pi}\left(M^*; \cH_{k, H}\right)
        \\
		      &:=\sum_{h=1}^H \mathbb E_{k}\mathbb E_{\pi}^{ {M}^*}\left[\KSD\left({P}^{M_k}(\cdot|s_h^k,a_h^k), {P}^{M^*}(\cdot|s_h^k,a_h^k)\right)\right].\nonumber
		  \end{align}
  {where $\mathbb E_{\pi}^{{M^*}}$ is taken with respect to $(s_h^k,a_h^k)$, and $\mathbb E_{k}$ is with respect to $\pi$ and environment $M_k$. There are two main differences here as compared to information gain defined in \eqref{informayion_gain_KL}. First, we use KSD to characterize if two transition models are close or not. Second, we use the distributional distance to true MDP $M^*$ in \eqref{informayion_gain_KSD} in contrast to posterior variance in \eqref{informayion_gain_KL}.} Hence, the ratio objective in \eqref{def:information_ratio_one} would modify to Stein information ratio as 
	  \begin{align}\label{def:information_ratio}
		       {\Gamma_k^{{\KSD}}(\pi):=\frac{(\mathbb E_k[V_{1,\pi^*}^{M^*}(s_1^k)-V_{1,\pi}^{M^*}(s_1^k)])^2}{ \mathbb K_k^{\pi}\left( M^*; \cH_{k, H}\right)}\,,}
		  \end{align}
	and we select $\pi^k_{\texttt{SIDS}}=\argmin_{\pi} \Gamma_k^{\text{KSD}}(\pi)$.  We remark that \eqref{informayion_gain_KSD}-\eqref{def:information_ratio}  are the point of departure from the existing IDS-based methods \citep{hao2022regret}. Stein information gain term of \eqref{informayion_gain_KSD} is different from the reduction in entropy (as in information gain) but instead characterizes the distributional distance to the true MDP transition dynamics $P^{M^*}$. So it forces the algorithm to move the model estimate towards the optimal than focusing on the coverage of space induced by information gain term in \eqref{def:information_ratio_one} (addressing L2). Another interesting aspect of the modified ratio objective in \eqref{def:information_ratio} is that it is computationally tractable due to the use of KSD and we no longer need to utilize Pinsker's inequality to approximate uncertainty via posterior variance which is unavoidable in IDS based approaches for practical implementation \citep{hao2022regret} (addressing L1) . 

   Unfortunately, although KSD is well-established, the pre-existing machinery \eqref{KSD_existing}  does not directly apply to our setting in \eqref{informayion_gain_KSD}. The impediments are twofold: firstly, \eqref{jordan_stein} holds for the continuous smooth densities, but our setting is tabular MDP;
   secondly, \eqref{jordan_stein} applies to estimating unconditional target distributions, but in an MDP context, we require conditional distributions. To adapt Stein discrepancy to our setting, we need to first address both of these issues next. 

		  \subsection{Discrete Conditional KSD}\label{DSD_section}
	 In this subsection, we introduce the kernelized conditional discrete stein operator and \textit{Discrete conditional kernelized Stein Discrepancy (DSD)} for analyzing the distributional distance between $P^{M_k}$ where $M_k \sim \phi(\cdot|\mathcal{H}_k)$ and $P^{M^*}$. This is a unique contribution of this work which may be of independent interest in mathematical statistics. In tabular RL setting, if we are in state $(s,a)$, then we know $s'\sim P^{M^*}(\cdot | s,a)$, and  up to $k^{th}$ episode, we collect samples in  dictionary $\mathcal{D}_k := \{((s_1,a_1),s'_1), ((s_2,a_2),s'_2), \cdots ((s_k,a_k),s'_k)\}$. 
  Now, the objective is to derive DSD between {$P^{M_k}(\cdot|s,a)$ and ground truth $P^{M^*}(\cdot|s,a)$} leveraging the recent literature on conditional independence testing with Kernel Stein's method \citep{Wittawat} (defined only for continuous smooth densities). For simplicity and analysis in this subsection, let us denote the state-action pair $(s,a) \rightarrow {x} \in \mathcal{X}:=\mathcal{S}\times \mathcal{A}$ and the corresponding next state $s'  \rightarrow {y} \in \mathcal{S}$. 

    To write DSD, we start by defining the Stein operator (cf. Appendix \ref{KSD_existing_appenndix} for unconditional case) as
        \begin{align} \label{our_kernel}
            \! \! \! \! \kappa_{M_k}(( x, y), \cdot) = G(x, \cdot) \big[s_{P^{M_k}}( y) l( y, \cdot) - \triangle^{\ast} l( y,\cdot)\big],
        \end{align}
   where $M_k$ signifies the dependence on $P^{M_k}$, function $l: \mathcal{S}\times \mathcal{S} \rightarrow \mathbb{R}$ is a positive definite kernels, and $G(x, \cdot)$ is a real-valued kernel associated with the RKHS $\mathcal{F}_{k}$ and explicitly defined in Proposition \ref{lemma_KCDSD}. 
   Further in \eqref{our_kernel}, we define score function $s_{P^{M_k}}(y)$ for $P^{M_k}$ and $\triangle^{\ast}$ as the difference operator w.r.t inverse permutation denoted by $\land$ for the set $\mathcal{S}$. 
   We denote  $\triangle$ as the 
   cyclic permutation $\lor$ for the set $\mathcal{S}$. 
   For example,  with $\mathcal{S}= \{+1,-1\}$, $\lor s = -s$, $\forall s \in \mathcal{S}$. On the other hand, inverse permutation satisfies $\lor(\land (s)) = \land(\lor (s)) = s$ (see \citep{yang18c} more details). Therefore, we expand the terms in  \eqref{our_kernel} as 
  \begin{align}\label{def_disc_op}
      s_{P^{M_k}}( y)_i =& \frac{\triangle_{ y_i} P^{M_k}( y| x)}{P^{M_k}( y| x)} = 1 - \frac{P^{M_k}( \lor_i  y)}{P^{M_k}( y| x)
      }\\
      \triangle^{\ast}_{y_i} l(y, \cdot) =& l(y, \cdot) - l(\land_i y, \cdot),
      \end{align}
for $i = 1,2, \cdots, S$. Next, we provide the definition of DSD between two conditional pmfs in Proposition \ref{lemma_KCDSD}.  
	  \begin{proposition}\label{lemma_KCDSD}
   \normalfont
		  Let $G{{(x,\cdot)}} \colon\mathcal{X}\times\mathcal{X}\to\mathbb{R}$ be a real-valued kernel associated with the RKHS $\mathcal{F}_{k}$   
    and $l\colon\mathcal{S}\times\mathcal{S}\to\mathbb{R}$ be positive definite kernels. Assume $G_{{x}}({x},{x}'):=k({x},{x}')I$
		  for a positive definite kernel $k\colon\mathcal{X}\times\mathcal{X}\to\mathbb{R}$.
		  Then, DSD between  {$P^{M_k}(s'|s,a)$ and $P^{M^*}(s'|s,a) $} is given by 
		  \begin{align}\label{eq:kssd_pop_ustat}
			   \text{DSD}(P^{M_k}, P^*) =\mathbb{E}_{[{(x, y)}, {(x', y')}]} [\kappa_{M_k}(({x},{y}),({x}',{y}'))], 			  
			  \end{align}
     where ${(x, y)}$ and ${(x', y')}$ are samples from the joint distribution $P^{M^*}$. {For simplicity of notations, we will denote $\text{DSD}(P^{M_k}, P^{M^*}) \rightarrow \text{DSD}(P^{k})$ which quantifies the Stein distance of $M_k$ from the true MDP $M^*$.}
		  \end{proposition}

 The proof of Proposition \ref{lemma_KCDSD} is provided in Appendix \ref{appendix_lemma_1_proof} and we show that $\DSD(P^{M^*}(\cdot|s,a))  = 0$ iff $P^{M_k}(\cdot|s,a) = P^{M^*}(\cdot|s,a)$. 
	  In the tabular RL setting, since both $ x \in \mathcal{X}$ and $ y \in \mathcal{S}$ are discrete random variables, we consider both $l( y,   y') =  \exp\{-H( y,  y')\}$ and  $k( x,  x') = \exp\{-H( x,  x')\}$ as the exponential Hamming kernel which is a positive definite kernel. However, the specific selection of kernels and kernel design in tabular RL is the scope of future work. We remark that since now DSD is defined, we will use DSD in place of KSD in all the places moving forward.

    \subsection{{\algo}: Proposed  Algorithm}
	  Now, we are ready to present the proposed STEin information dirEcted
sampling for model-based Reinforcement LearnING ({\algo}) algorithm for tabular settings, summarized in Algorithm \ref{alg:MPC-PSRL}. We begin {\algo} by assuming a prior over the transition, and the rewards model denoted by $\phi_{\mathcal{D}_{1}}=\{\mathcal{P}_{\mathcal{D}_{1}},\mathcal{R}_{\mathcal{D}_{1}}\}$. 
%
At episode $k$, {\algo} first samples a transition model {$P^{M_k}$} and rewards model {$R^{M_k}$} from the posterior distribution $\phi_{\mathcal{D}_{k}}=\{\mathcal{P}_{\mathcal{D}_{k}},\mathcal{R}_{\mathcal{D}_{k}}\}$. It then optimizes the policy under these sampled models by minimizing the proposed Stein information ratio $\argmin_\pi \Gamma_k^{\text{DSD}}(\pi)$ as in \eqref{def:information_ratio}. This inner optimization procedure is a key aspect of {\algo}, as it uses the Stein information to guide exploration. Finally, the agent interacts with the real environment using the resulting policy $\pi^k_{\texttt{IDS}}(a|s)$ to collect new samples at episode $k$ and store them in $\mathcal{C}$. {Instead of just appending $\mathcal{C}$ to dictionary $\mathcal{D}_k$, we propose to use an intelligent sample selection  procedure (similar to SPMCMC \citep{stein_point_Markov}) on the collected samples in each episode $k$ before updating the dictionary $\mathcal{D}_{k}$. This procedure selects new samples by minimizing their similarity to existing samples in the dictionary $\mathcal{D}_{k}$ through a local optimization procedure, as outlined in Appendix \ref{proof_lemma_KCSD} starting from equation \eqref{here_12}. This selection procedure helps to derive tighter convergence rates in the next section.

   \begin{algorithm}[ht]
		    \caption{\algo: \textbf{STE}in information dir\textbf{E}cted sampling
for model-based \textbf{R}einforcement Learn\textbf{ING} }
		    \label{alg:MPC-PSRL}
		  \begin{algorithmic}[1]
			    \STATE \textbf{Input} : Episode length $H$, Total timesteps $T$, Dictionary $\mathcal{D}$, prior distribution $\phi=\{\mathcal{P},\mathcal{R}\}$ , policy $\pi(a|s)$, hyperparameter $\gamma$.
			   \\
			    \STATE \textbf{Initialization} : Initialize dictionary  $\mathcal{D}_1$ with random $\pi_0(a|s)$, posterior $\phi_{\mathcal{D}_{1}}=\{\mathcal{P}_{\mathcal{D}_{1}},\mathcal{R}_{\mathcal{D}_{1}}\}$, policy $\pi_{1}(a|s)$.
			    \FOR{{Episodes} $k=1$ to $K$}
			    %
			    \STATE \textbf{Sample} a transition {$P^{M_k} \sim \mathcal{P}_{\mathcal{D}_{k}}$ and reward model $R^{M_k} \sim \mathcal{R}_{\mathcal{D}_{k}}$  and initialize empty $\mathcal{C}=[\ ]$ }
			    \STATE \textbf{Estimate} the optimal policy by minimizing the \textit{Stein information ratio} : $\pi^{k}_{\texttt{IDS}} (a|s) \leftarrow \argmin_\pi \Gamma_k^{\text{DSD}}(\pi, M^*)$ as in {\eqref{def:information_ratio}}
			    \STATE \textbf{Interact} with the environment using the optimal policy  $\pi^{k}_{\texttt{IDS}} (a|s)$ to gather and initialize empty $\mathcal{C}= \{s_{k,1},a_{k,1},r_{k,1},\cdots,s_{k,H},a_{k,H},r_{k,H}\}$
                    \STATE \textbf{Select} the subset of samples $\mathcal{C'} \in \mathcal{C}$ with least similarity to sample in samples in $\mathcal{D}_{k}$ in terms of Stein kernel
		       \STATE \textbf{Update} dictionary ${\mathcal{D}}_{k+1} \leftarrow {\mathcal{D}}_{k} \cup {\mathcal{C'}}$ and update the posteriors $\mathcal{P}_{\mathcal{D}_{k}}, \mathcal{R}_{\mathcal{D}_{k}}$
		
			    \ENDFOR
			  \end{algorithmic}
		  \end{algorithm}	
	
  \section{Regret Analysis}\label{regret_analysis}
	
	  In this section, we derive the merits of incorporating Stein's method into the Bayesian regret of an MBRL method for the first time. We start by deriving our key result in Theorem {\ref{theorem_0}}, which connects Bayesian regret (cf. \eqref{bayesian_Regret})  with Stein information gain (cf. \eqref{informayion_gain_KSD}). Then, we analyze the evolution of the model-based estimates in terms of DSD [cf. Sec. \ref{DSD_section}], which decreases with the number of samples processed (Lemma  \ref{lemma_KSD}). Next, we connect this bound to the Stein information ratio objective and the total Stein information gain in Lemma \ref{lemma_1}. Finally, we utilize Theorem \ref{theorem_0}, Lemma \ref{lemma_KSD}, and Lemma \ref{lemma_1}  to derive Bayesian regret in terms of state and action space cardinality in Theorem \ref{theorem} . We also extend our analysis to regularized settings in Theorem \ref{theorem_regularized_regret}. 
 Next, we present our first Bayesian regret as follows. 
	  \begin{theorem}(Stein Information Theoretic Regret)\label{theorem_0}
	       		  		 When Algorithm \ref{alg:MPC-PSRL} is run for $K$ episodes of horizon length $H$, it achieves the following Bayesian regret:
		  \begin{align}\label{final_bayes_regret}
			      \BR_K\leq \sqrt{\mathbb E[{\Gamma}^{*}]  K \sum_{k=1}^K \mathbb E [\mathbb K_k^{\pi}\left( M^*; \cH_{k, H}\right)]}\,,
			  \end{align}
    {where $\mathbb K_k^{\pi}\left(M^*; \cH_{k, H}\right)$ is the Stein information gain (cf. \eqref{informayion_gain_KSD}) and $\Gamma^*$ is the worst case Stein information ratio such that $\Gamma_k^{{\DSD}}(\pi) \leq \Gamma^*$ for any $k \in K$ and $\pi$.}
	  \end{theorem}
	 The proof of Theorem \ref{theorem_0} is provided in Appendix \ref{proof_theorem_0}.  We call the regret in Theorem \ref{theorem_0} as the \emph{Stein information theoretic regret}  because  it upper bounds the Bayesian regret in terms of Stein information gain (cf. \eqref{informayion_gain_KSD}), which is a DSD between the estimated model and the true model. This is the main point of departure as compared to information-theoretic regret derived in \citet{osband2017posterior,hao2022regret}, which eventuates in substitution of the information gain by the posterior variance as its uncertainty quantifier due to the computational effort required to estimate information gain. By contrast, we consider this distributional distance instead in terms of integral probability metrics, which under some hypotheses, are computable as DSD (cf. Sec. \ref{DSD_section}).  To the best of our knowledge, this is the first time this notion of distance to ground truth, which is typical of frequentist analysis of Bayesian methods, has been incorporated into the Bayesian regret. 
  
  Next, we proceed toward deriving an absolute upper bound on the regret in terms of $S$, $A$, and $H$. To achieve that, we present two intermediate results in Lemma \ref{lemma_KSD}-\ref{lemma_1}. 
\begin{lemma}(\normalfont{DSD Convergence Rate})\label{lemma_KSD}
     \normalfont
		With Algorithm \ref{alg:MPC-PSRL}, we collect dictionary $\mathcal{D}_k$ for which it holds that
	  \begin{align}
		      \mathbb{E}_k\left[\text{DSD}(P^{k};{\mathcal{D}}_{k})^2\right] & = \mathcal{O} \left(\frac{S^2A}{k}\right)\label{lemma_KCDSD_main_body},
		  \end{align}	 
    for all $k$. Here $\text{DSD}(P^{k}) := \text{DSD}((P^{M^*}, P^{M^k})$
    In \eqref{lemma_KCDSD_main_body}, and $\text{DSD}(P^k;{\mathcal{D}}_{k})^2$ denotes the empirical approximation using dictionary $\mathcal{D}_{k}$ of discretized conditional kernelized Stein discrepancy (cf. \eqref{eq:kssd_pop_ustat}) between $P^{M_k}$ and true  $P^{M^*}$. {An improved bound of order $\mathcal{O} (\frac{SA}{k})$, can be derived under certain boundedness and regularity conditions on the RKHS.}
		  \end{lemma}
	  The proof of Lemma \ref{lemma_KSD} is provided in {Appendix \ref{proof_lemma_KCSD}}. 
Lemma \ref{lemma_KSD} establishes the convergence of the transition model estimation $P^{M_k}$ to the true model $P^{M^*}$, which is an important result to prove next Lemma \ref{lemma_1}.
    \begin{figure*}[ht]
                \centering
                \hfil
            \subfigure[Low sparsity.]{
                    \includegraphics[width=0.65\columnwidth,clip = true]{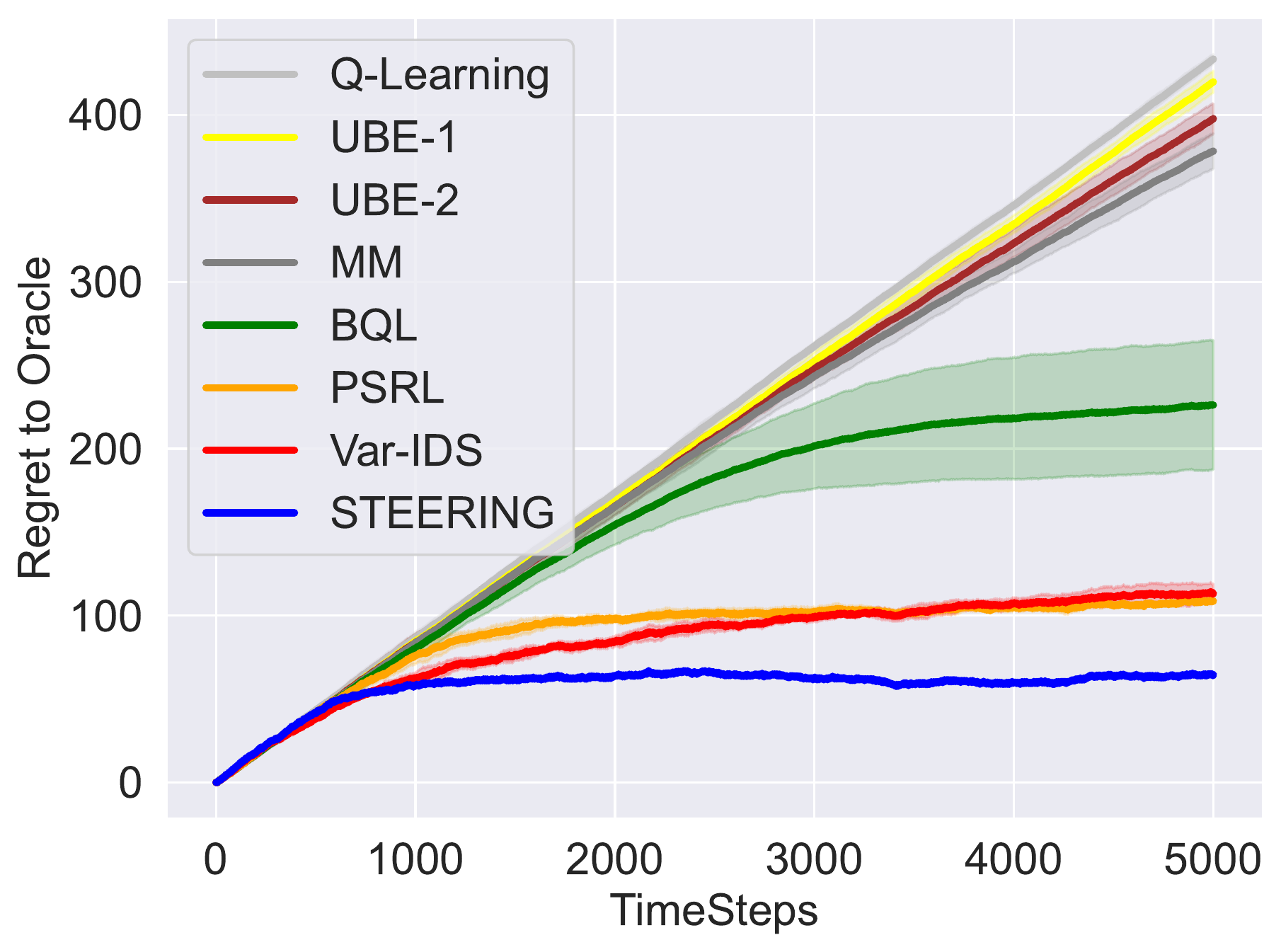}
                 }
                 \hfil
        \subfigure[Medium sparsity.]{
                \includegraphics[width=0.65\columnwidth,clip = true]{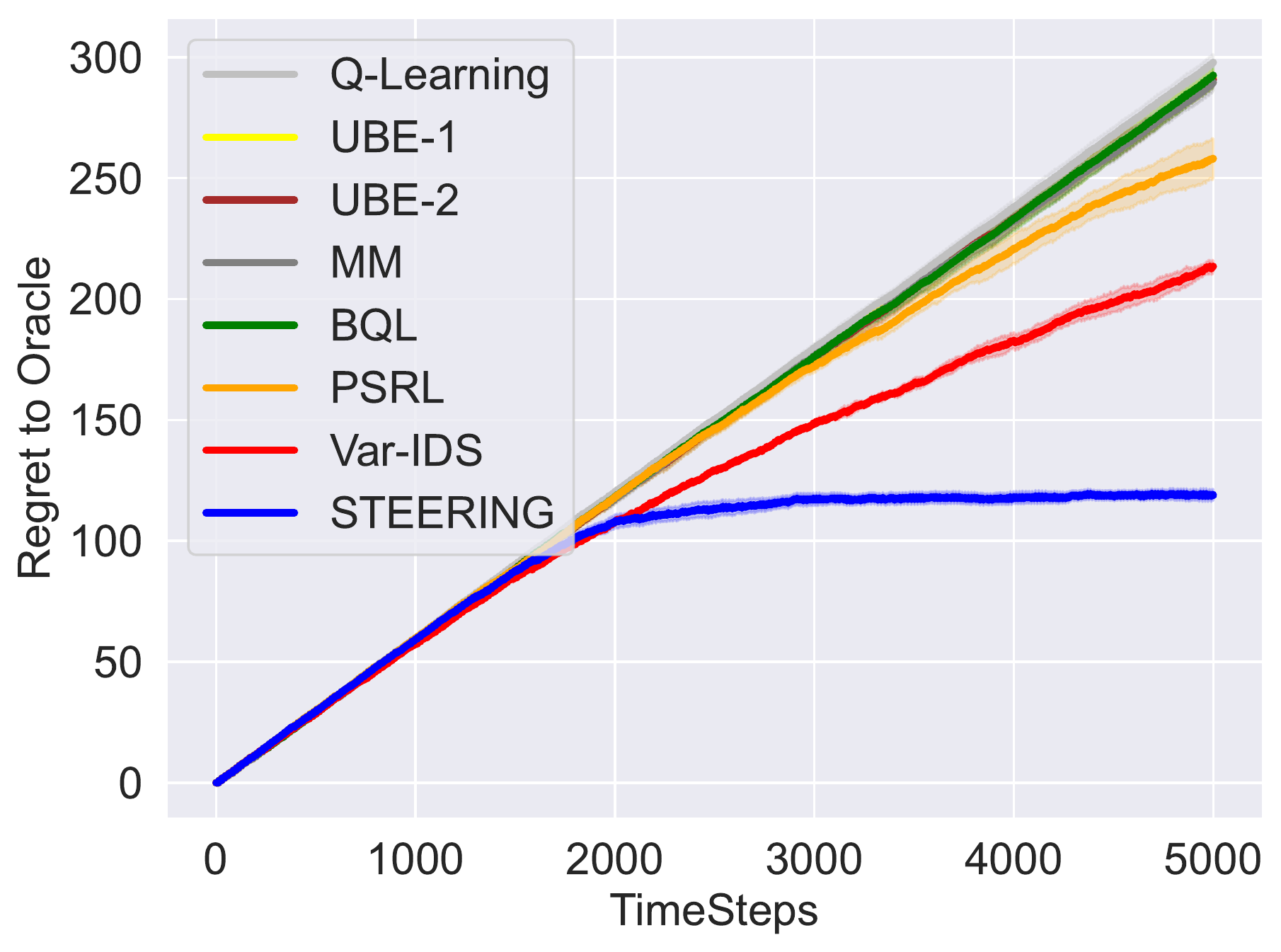}
             }
             \hfil
        \subfigure[High sparsity.]{
                \includegraphics[width=0.65\columnwidth,clip = true]{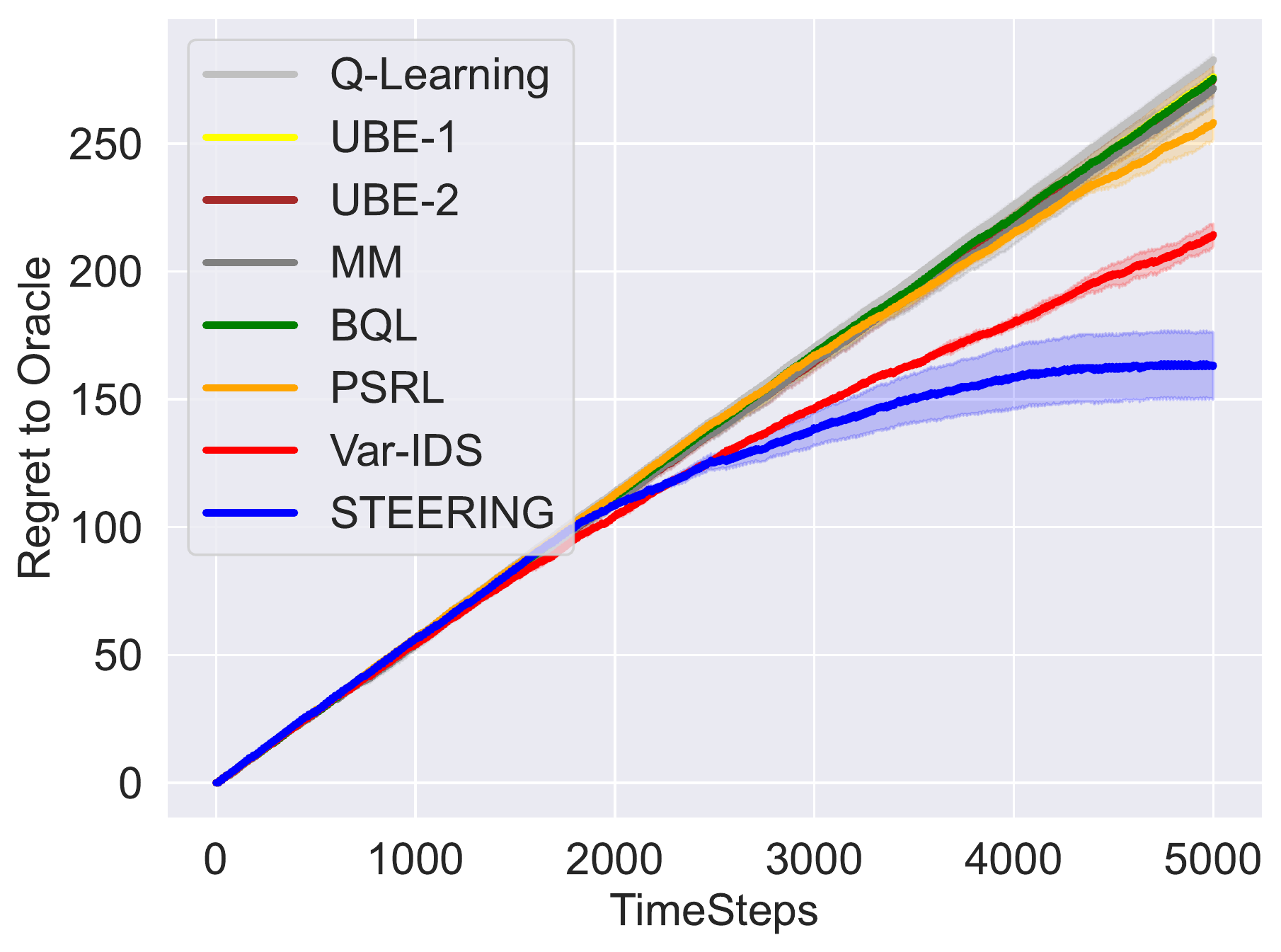}
        }
        \hfil
            \caption{This figure compares the performance of {\algo} on DeepSea enviornment \citep{osband17a} against the existing RL baselines vanilla Q-learning with $\epsilon-$greedy action selection \citep{Watkins1992}, Bayesian Q-learning (BQL) \citep{dearden_bql}, Uncertainty Bellman Equation (UBE) \citep{donoghue_ube}, Moment matching (MM) across Bellman equation \citep{markou_mm}, Posterior Sampling RL (PSRL) \citep{osband2013}, and IDS \citep{hao2022regret}. We present results for three sparsity levels and  observe {\algo} outperforms existing baselines. Interestingly, the performance of {\algo} is comparable (still much better shown in Fig.~\ref{final_figure_sota}(a)-(b) to PSRL or IDS with low or medium sparsity, but for high sparsity in Fig.~\ref{final_figure_sota}(c), {\algo} significantly outperforms the other methods.}
            \label{final_figure_sota}
    \end{figure*}
	  \begin{lemma}\label{lemma_1}
   \normalfont
		 For Algorithm \ref{alg:MPC-PSRL}, after $K$ episodes of horizon length $H$, it holds that 
  
			      (1) Stein information ratio (cf. \eqref{def:information_ratio}) is upper bounded as $\mathbb E[\Gamma_k^{\text{DSD}}(\pi)]\leq SAH^3$ for all $k$.
         
			      (2) Total Stein information gain (cf. \eqref{informayion_gain_KSD}) is bounded as 
			      \begin{align*}
				    \mathbb\sum_{k=1}^K \mathbb E [\mathbb K_k^{\pi}\left( M^*; \cH_{k, H}\right)]\leq  HS^2A(\log K).     \qed
				  \end{align*}  
		  \end{lemma}
	  The proof is provided in Appendix \ref{proof_lemma_1}. The upper bounds established in Lemma \ref{lemma_1} are crucial to specialize the general regret bounds developed in Theorem \ref{theorem_0} in terms of state-action cardinalities, as follows in Theorem \ref{theorem}. 
     
	
	  \begin{theorem}\label{theorem}
		  		 When Algorithm \ref{alg:MPC-PSRL} is run for $K$ episodes of horizon length $H$, it achieves the following performance in terms of Bayesian regret:
		  \begin{align}\label{theorem_statement}
			      \BR_K=\tilde{\mathcal{O}}\left(\sqrt{H^4 S^3A^2 K }\right)\,,
			  \end{align}
    where $\tilde{\mathcal{O}}$ absorbs the log factors, $S$ and $A$ are the state and action space cardinalities, respectively. 
		  \end{theorem}
The proof of Theorem \ref{theorem} is provided in Appendix \ref{proof_of_main_theorem}. Theorem \ref{theorem} states that {\algo}  achieves  Bayesian regret, which is sublinear in terms of episode index $K$ and the number of actions $A$, and linear in terms of the number of states $S$.

\textbf{Remark 1 (Proof Sketch):} Here, we provide  insights regarding the regret proof of {\algo}. One key factor in our analysis is the use of distributional directed sampling achieved via the integration of DSD and a point selection strategy inspired by SPMCMC \citep{stein_point_Markov}. By utilizing an intelligent point selection method  (cf. proof of Lemma \ref{lemma_KSD} in Appendix \ref{proof_lemma_KCSD}), we can establish convergence to underlying true distributions, as previously demonstrated in different contexts by \citep{koppel2021consistent, stein_point_Markov}. Specifically, our point selection approach (Appendix \ref{proof_lemma_KCSD} Eq. \ref{here_12}) $\inf_{(x,y)} \sum_{(x_i, y_i) \in {\mathcal{D}}_{k-1}}\kappa_{M_k}((x_i, y_i),(x, y))$ involves choosing a new sample $(x, y)$ as the point that minimizes the similarity (in RKHS) to the current samples in the dictionary ${\mathcal{D}}_{k-1}$ resulting in directed exploration. 
}

\textbf{Remark 2 (Improved Regret):} Here, we provided insights to obtain an improved version of regret provided in Theorem \ref{theorem}. We start by mentioning that the final bound in Theorem \ref{theorem} depends upon the Stein information gain bound obtained in Lemma \ref{lemma_1}.  We can obtain an improved bound on the Stein information gain term as summarized in Corollary \ref{improvement_corollary}. 
\begin{corollary}\label{improvement_corollary}
    Under additional assumptions on the structure of the Stein kernel, the Stein information gain upper bound (cf. Lemma \ref{lemma_1}) can be improved to $HSA(\log K)$. 
\end{corollary}
The details are provided in Appendix \ref{proof_lemma_KCSD} and \ref{proof_lemma_1}. The result in Corollary \ref{improvement_corollary} would lead to an improved regret bound of $\tilde{\mathcal{O}}(\sqrt{H^4 S^2 A^2 K})$.

	  \textbf{Regularized {\algo}.} We also consider a regularized Stein information gain-based sampling objective and optimize the policy as 
	  \begin{align}\label{def:regularized_stein}
		       \pi^k_{\texttt{r-IDS}}= \argmax_{\pi}  \mathbb E_k[V_{1,\mu}^{M^*}(s_1^k)]) + \lambda
		       \mathbb K_k^{\pi}\left(M^*; \cH_{k, H}\right)\,,
		  \end{align}
		  where $\lambda$ is the unknown regularization parameter. 
	Next, we prove that the new regularized objective incurs the same Bayesian regret as the Stein information ratio in \eqref{def:information_ratio}. 
		  \begin{theorem}(Regularized  Regret)\label{theorem_regularized_regret}
	       		  		 When Algorithm \ref{alg:MPC-PSRL} (after replacing step 5 with \eqref{def:regularized_stein}) is run for $K$ episodes of horizon length $H$,  it achieves the following performance in terms of Bayesian regret
		  %
		  %
		     $\BR_K \leq  \sqrt{\frac{3}{2} \mathbb E[{\Gamma}^*]  K \sum_{k=1}^K \mathbb E [\mathbb K_k^{\pi}\left( M^*; \cH_{k, H}\right)]}$,
   for 		     $\lambda = \sqrt{K\mathbb E[\Gamma^*]/\mathbb \sum_{k=1}^K \mathbb E [\mathbb K_k^{\pi}\left( M^*; \cH_{k, H}\right)}$.
	  \end{theorem}
  The proof of Theorem \ref{theorem_regularized_regret} is provided in Appendix \ref{proof_regularized_regret}.
  \section{Experiments}
	  In this section, we evaluate the performance of the proposed {\algo} algorithm and compare it with other existing state-of-the-art algorithms. Since we are interested in developing algorithms for efficient directed exploration for an RL agent, we consider a challenging tabular sparse reward environment of DeepSea Exploration (DSE) introduced in \citet{osband17a} (Fig. \ref{fig:new_architecure}).  
   \begin{figure}[t]
        \centering
        \vspace{-0mm}
        \includegraphics[width=0.9\columnwidth, clip=true]{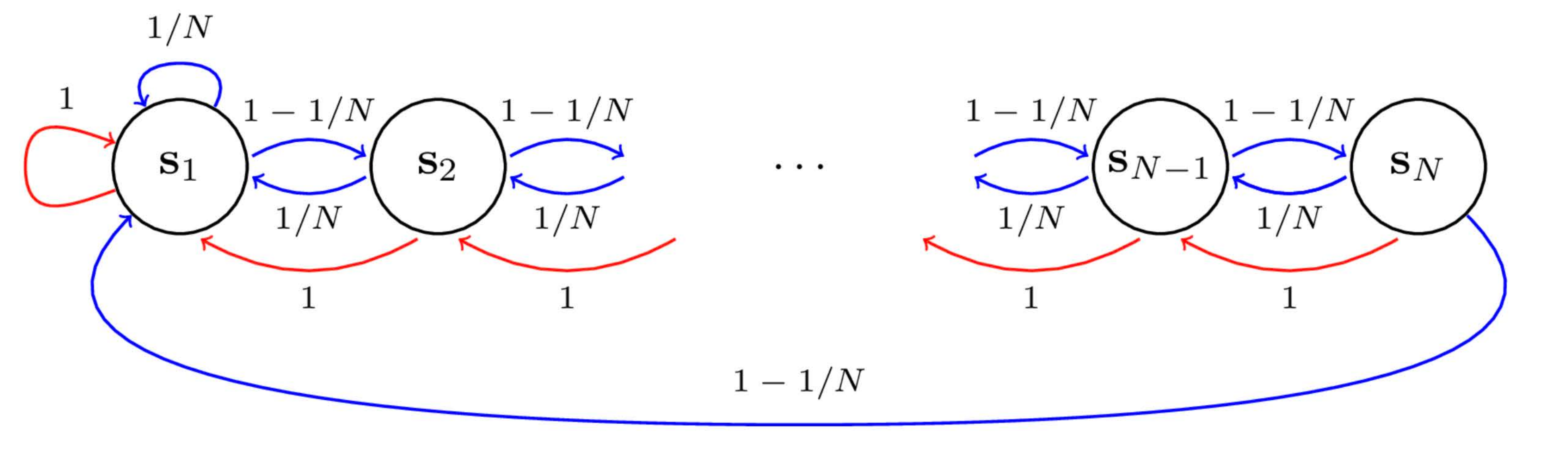}
        \caption{DeepSea Exploration Environment.}
                \vspace{-0mm}
        \label{fig:new_architecure}
    \end{figure}
     The DSE environment tests the agent's capability of directed and sustained exploration. The agent starts from the left-most state and can swim left or right from each of the $N$ states in the environment with {near zero rewards everywhere except a reward of $r = 1$ only on a successful swim-right to $s = N$ (see Appendix \ref{environment_Details} for more details)}. Hence, increasing the number of states $N$ induces more sparsity in the environment, making it extremely hard for the agent to explore without directed exploration, as verified in Fig. \ref{fig:psrl_sparse}. This motivates us to present improvements in the DSE environment. Next, we test {\algo} on four different aspects of performance:  (1) Regret to the Oracle, (2) Directed exploration, (3) Robustness to  prior belief, and  (4) Convergence to optimal value function. 
        \textbf{(1) Regret to the Oracle:} First, in Fig.~\ref{final_figure_sota},  we compare {\algo} with other Bayesian/ non-Bayesian RL algorithms in terms of the cumulative regret accumulated with respect to an oracle agent following the optimal policy (refer to Appendix \ref{additonal_experiments} for details).   We present results in Fig.~\ref{final_figure_sota} for three different levels of sparsity: \emph{low} ($N=8$), \emph{medium} ($N=14$), and \emph{high} ($N=15$).  The results validate our hypothesis that under highly sparse environments, general existing RL methods
        fail to explore efficiently, resulting in higher regret but  {\algo} outperforms. 
       \begin{figure}[t]
        \centering
        \includegraphics[width=\columnwidth]{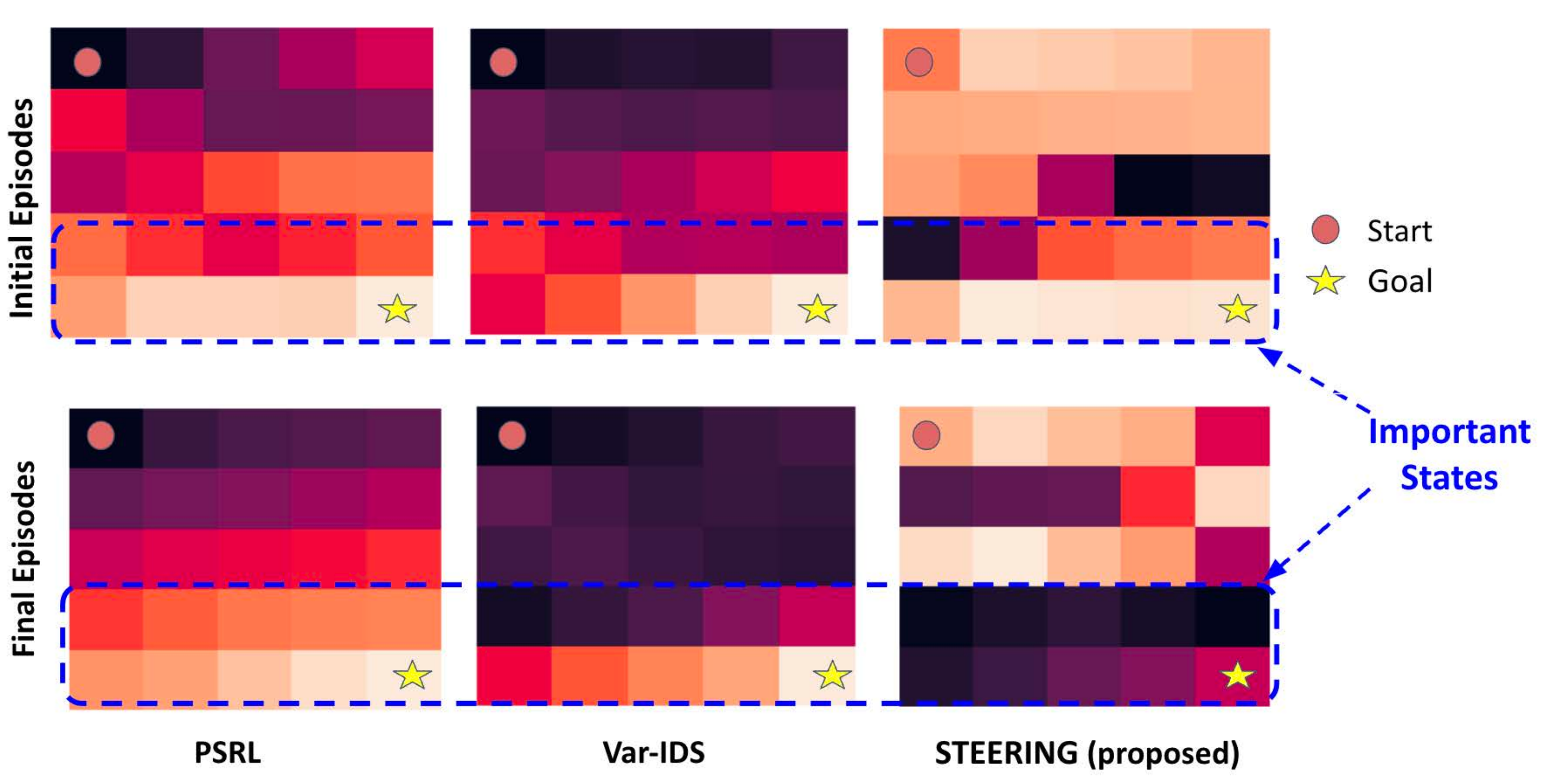}
        \caption{(Directed Exploration)  We plot heatmaps showing the distribution of state-action visits in the initial and final episodes of training for PSRL, IDS, and {\algo}.  The results reveal that {\algo} can effectively identify and visit the most important states indicated by the dark colors in the bottom row for {\algo}. By important states, we mean the states closer to the goal state.  In contrast, PSRL and IDS exhibit less directed exploration (see lighter colors in the bottom row for PSRL and IDS), with their heatmaps showing less concentration on important states. These findings demonstrate the superior performance of {\algo} in guiding the exploration process towards optimality.} 
        \label{fig:occupancy}
    \end{figure}
	
        \textbf{(2)  Directed Exploration by {\algo}}: To emphasize the nature of effective directed exploration provided by {\algo}, we analyze its state-action space coverage in Fig.~ \ref{fig:occupancy}. We compare the heatmaps of state-action occupancy measure of the initial (top row) and final episodes (bottom row) for PSRL,  IDS, and {\algo}.
 
        \textbf{(3) Robustness to Prior Belief}: The performance of Bayesian algorithms depends upon the prior, which might be a confident but mis-specified belief in practical settings. To test against such scenarios, we perform an ablation study in Fig.~\ref{conc_plot} to validate the robustness of {\algo} to different levels of confidence of the prior belief over the MDP. 
        \begin{figure}[t]
        \centering
        \hfil
\subfigure[Strong belief prior.]
    {\includegraphics[width=.49\columnwidth,clip = true]{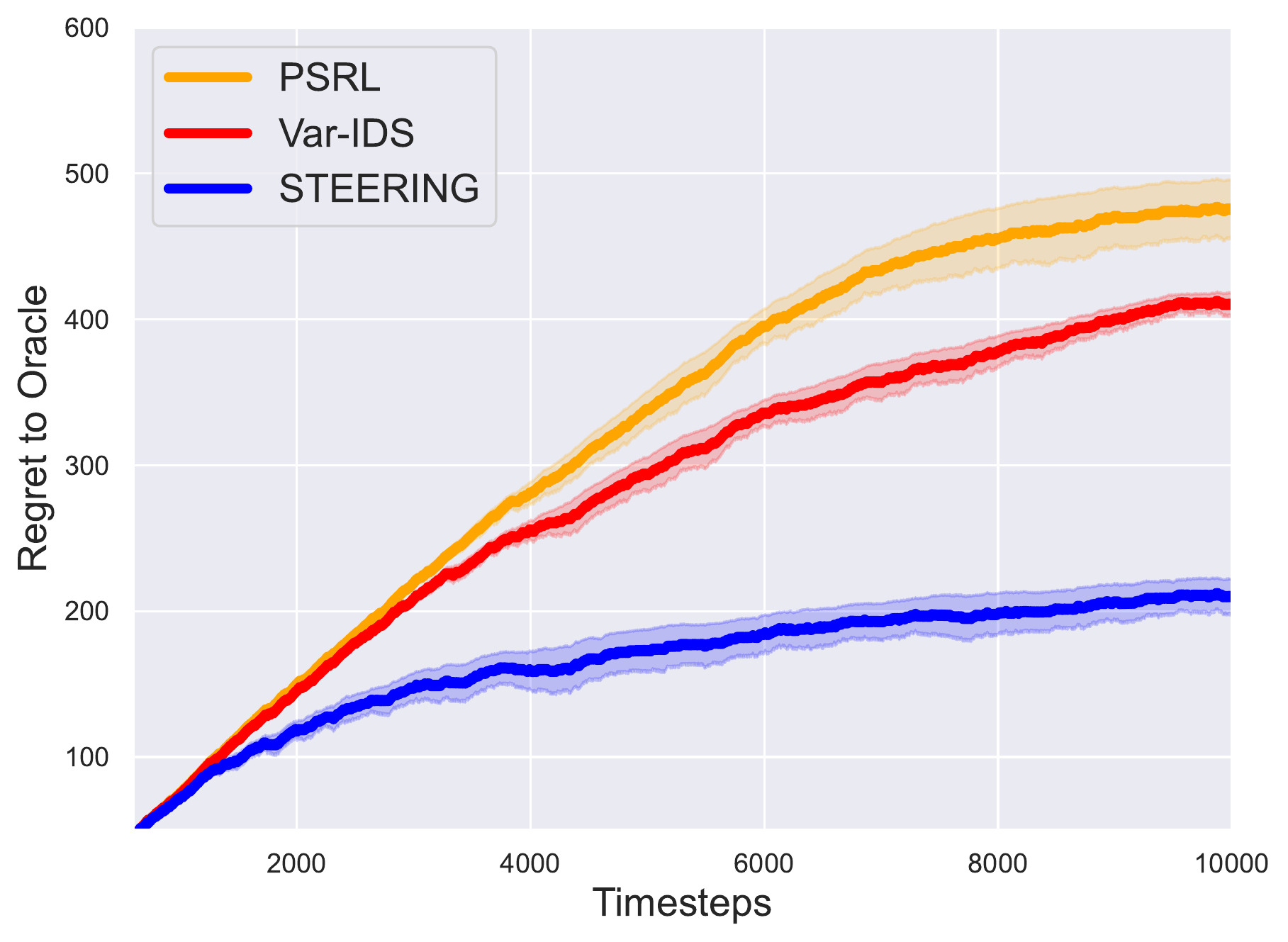}}
    \hfil
\subfigure[Weak belief prior.]
{\includegraphics[width=.49\columnwidth,clip = true]{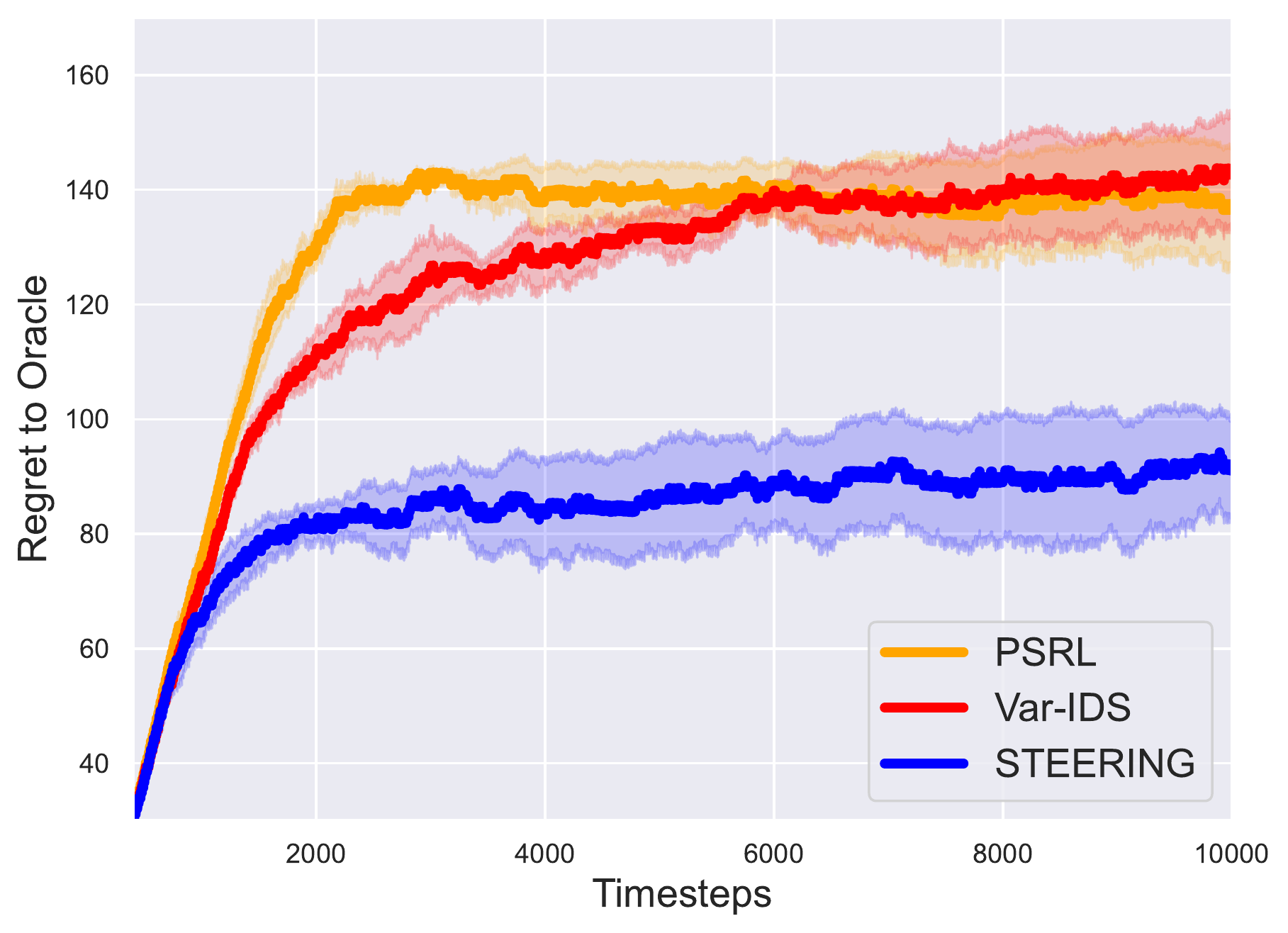}}
\hfil
\caption{ We consider two belief levels: \emph{weak} (higher variance) and \emph{strong} (small variance) 
        and compare the performances of PSRL, IDS, and {\algo}. We note in Fig.~\ref{conc_plot}(a), i.e., with strong prior belief, IDS provides  a marginal benefit over PSRL, but {\algo} is significantly better. This is due to the construction of {\algo} based on the notion of distance to true MDP and not entirely relying on the posterior variance. Whereas IDS, on the other hand, for computational tractability, relies on posterior variance via Pinsker inequality. For weak prior belief in Fig.~\ref{conc_plot}(b) also, {\algo} performs favorably. }
        \label{conc_plot}
        \end{figure}

        \textbf{(4) Convergence to Optimal Value function}: We perform additional experiments in Appendix \ref{convergence_Q} to analyze and compare the convergence of the predicted $\hat{Q}$ values for by {\algo}.

        In Fig. \ref{new_env}, we also compare {\algo} with baselines on WideNarrow MDP to validate its performance under factored posterior approximations and PriorMDP  to validate its performance in general and practical environments without special structures \citep{markou_mm}.        
   \begin{figure}[ht]
        \centering
        \hfil
        \subfigure[WideNarrow MDP.]{\includegraphics[width=.45\columnwidth,clip = true]{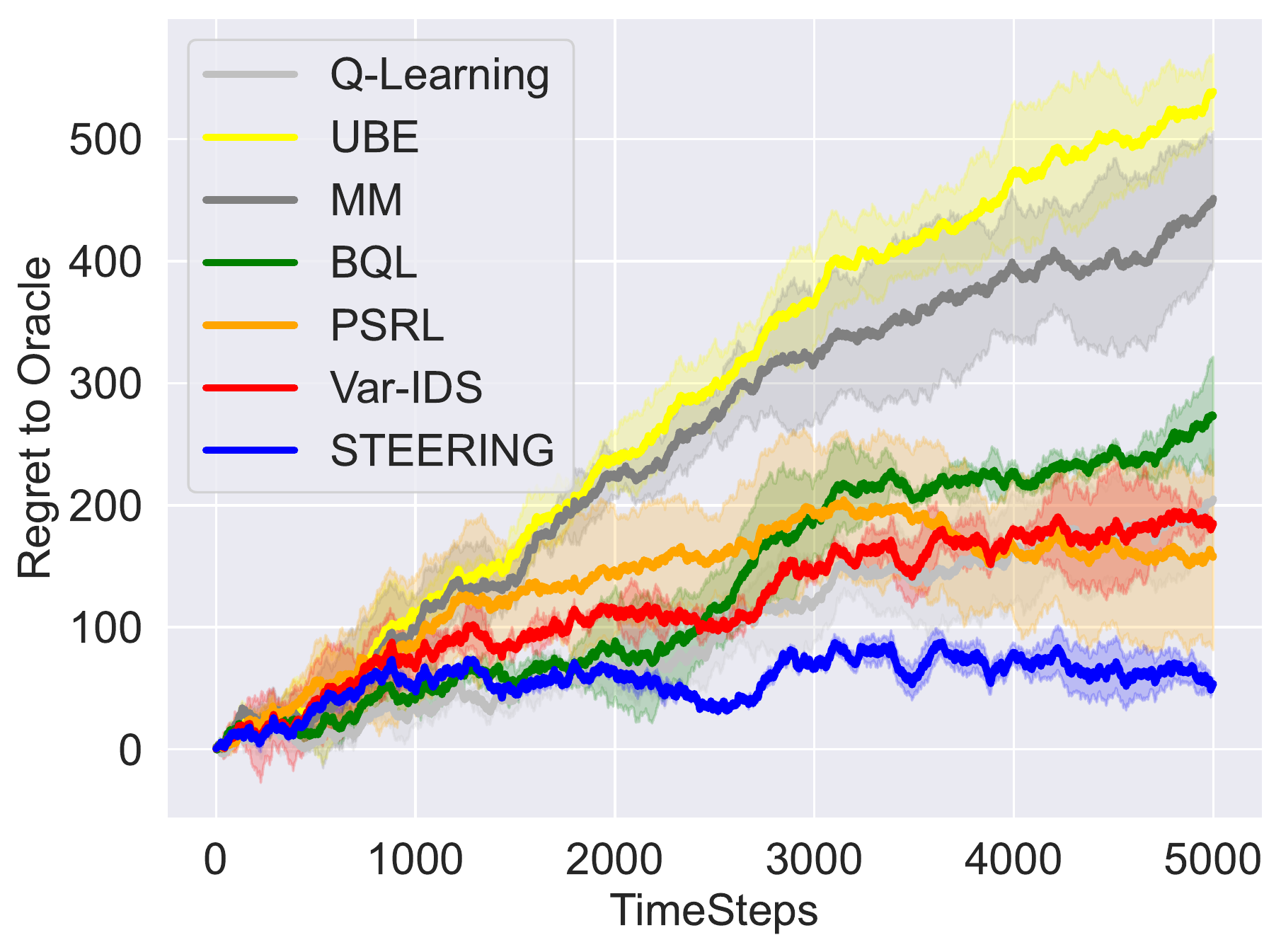}}
        \hfil
        \subfigure[Prior MDP.]{\includegraphics[width=.45\columnwidth,clip = true]{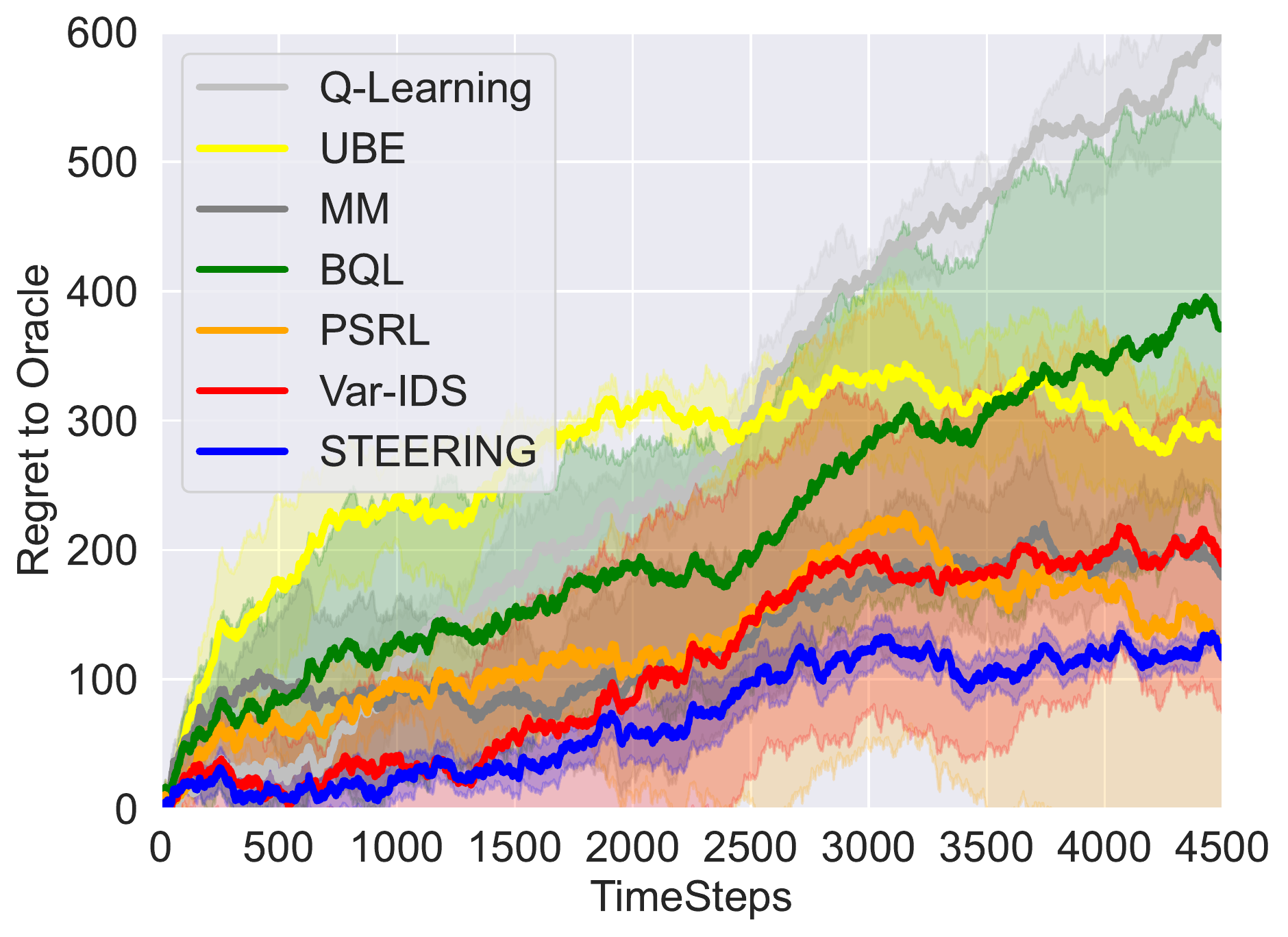}}
        \hfil
        \caption{Performance comparison of {\algo} and baseline algorithms in terms of Regret in two environments: WideNarrow MDP and PriorMDP.}
        \label{new_env}
        \end{figure}

   	  \section{Conclusions}
	  Information-directed sampling (IDS) provides a way to induce exploration incentives into model-based reinforcement learning (MBRL). But IDS approaches suffer from computational tractability issues. To make such ratio-based approaches computationally tractable and efficient, we propose a novel measure to quantify directed exploration through a distributional distance to the optimal model via kernelized Stein discrepancy. To this end, we introduced a novel notion of Stein information gain and Stein information-directed sampling in MBRL. We theoretically established prior-free sublinear Bayesian regret bounds and experimentally demonstrate favorable performance in practice.

\section{Acknowledgments}
Chakraborty and Huang are supported by National Science Foundation NSF-IIS-FAI program, DOD-ONR-Office of Naval Research, DOD Air Force Office of Scientific Research, DOD-DARPA-Defense Advanced Research Projects Agency Guaranteeing AI Robustness against Deception (GARD), Adobe, Capital One, and JP Morgan faculty fellowships. Bedi and Manocha
would like to acknowledge the support from Army Cooperative Agreement W911NF2120076 and
Amazon Research Award 2022. Mengdi Wang acknowledges the support by NSF grants DMS-1953686, IIS-2107304, CMMI-1653435, ONR grant 1006977, and C3.AI.

\bibliography{icml_2023}

\begin{thebibliography}{75}
\providecommand{\natexlab}[1]{#1}
\providecommand{\url}[1]{\texttt{#1}}
\expandafter\ifx\csname urlstyle\endcsname\relax
  \providecommand{\doi}[1]{doi: #1}\else
  \providecommand{\doi}{doi: \begingroup \urlstyle{rm}\Url}\fi

\bibitem[Achiam \& Sastry(2017)Achiam and Sastry]{achiam_lb}
Achiam, J. and Sastry, S.
\newblock Surprise-based intrinsic motivation for deep reinforcement learning,
  2017.
\newblock URL \url{https://arxiv.org/abs/1703.01732}.

\bibitem[Ahmed et~al.(2018)Ahmed, Roux, Norouzi, and
  Schuurmans]{zafarali_entropy}
Ahmed, Z., Roux, N.~L., Norouzi, M., and Schuurmans, D.
\newblock Understanding the impact of entropy on policy optimization, 2018.
\newblock URL \url{https://arxiv.org/abs/1811.11214}.

\bibitem[Amortila et~al.(2020)Amortila, Precup, Panangaden, and
  Bellemare]{amortila2020distributional}
Amortila, P., Precup, D., Panangaden, P., and Bellemare, M.~G.
\newblock A distributional analysis of sampling-based reinforcement learning
  algorithms.
\newblock In \emph{International Conference on Artificial Intelligence and
  Statistics}, pp.\  4357--4366. PMLR, 2020.

\bibitem[Arjovsky et~al.(2017)Arjovsky, Chintala, and
  Bottou]{arjovsky2017wasserstein}
Arjovsky, M., Chintala, S., and Bottou, L.
\newblock Wasserstein generative adversarial networks.
\newblock In \emph{International conference on machine learning}, pp.\
  214--223. PMLR, 2017.

\bibitem[Ayoub et~al.(2020)Ayoub, Jia, Szepesvari, Wang, and
  Yang]{ayoub2020model}
Ayoub, A., Jia, Z., Szepesvari, C., Wang, M., and Yang, L.
\newblock Model-based reinforcement learning with value-targeted regression.
\newblock In \emph{International Conference on Machine Learning}, pp.\
  463--474. PMLR, 2020.

\bibitem[Bedi et~al.(2022)Bedi, Chakraborty, Parayil, Sadler, Tokekar, and
  Koppel]{sparse3}
Bedi, A.~S., Chakraborty, S., Parayil, A., Sadler, B.~M., Tokekar, P., and
  Koppel, A.
\newblock On the hidden biases of policy mirror ascent in continuous action
  spaces.
\newblock In Chaudhuri, K., Jegelka, S., Song, L., Szepesvari, C., Niu, G., and
  Sabato, S. (eds.), \emph{Proceedings of the 39th International Conference on
  Machine Learning}, volume 162 of \emph{Proceedings of Machine Learning
  Research}, pp.\  1716--1731. PMLR, 17--23 Jul 2022.
\newblock URL \url{https://proceedings.mlr.press/v162/bedi22a.html}.

\bibitem[Berlinet \& Thomas-Agnan(2011)Berlinet and
  Thomas-Agnan]{berlinet2011reproducing}
Berlinet, A. and Thomas-Agnan, C.
\newblock \emph{Reproducing kernel Hilbert spaces in probability and
  statistics}.
\newblock Springer Science \& Business Media, 2011.

\bibitem[Borkar \& Meyn(2002)Borkar and Meyn]{borkar2002risk}
Borkar, V.~S. and Meyn, S.~P.
\newblock Risk-sensitive optimal control for markov decision processes with
  monotone cost.
\newblock \emph{Mathematics of Operations Research}, 27\penalty0 (1):\penalty0
  192--209, 2002.

\bibitem[Burda et~al.(2018{\natexlab{a}})Burda, Edwards, Pathak, Storkey,
  Darrell, and Efros]{burda_curiosity}
Burda, Y., Edwards, H., Pathak, D., Storkey, A., Darrell, T., and Efros, A.~A.
\newblock Large-scale study of curiosity-driven learning, 2018{\natexlab{a}}.
\newblock URL \url{https://arxiv.org/abs/1808.04355}.

\bibitem[Burda et~al.(2018{\natexlab{b}})Burda, Edwards, Pathak, Storkey,
  Darrell, and Efros]{curiosity2}
Burda, Y., Edwards, H., Pathak, D., Storkey, A., Darrell, T., and Efros, A.~A.
\newblock Large-scale study of curiosity-driven learning, 2018{\natexlab{b}}.
\newblock URL \url{https://arxiv.org/abs/1808.04355}.

\bibitem[Chakraborty et~al.(2022{\natexlab{a}})Chakraborty, Bedi, Koppel,
  Sadler, Huang, Tokekar, and Manocha]{chakraborty2022posterior}
Chakraborty, S., Bedi, A.~S., Koppel, A., Sadler, B.~M., Huang, F., Tokekar,
  P., and Manocha, D.
\newblock Posterior coreset construction with kernelized stein discrepancy for
  model-based reinforcement learning.
\newblock \emph{arXiv preprint arXiv:2206.01162}, 2022{\natexlab{a}}.

\bibitem[Chakraborty et~al.(2022{\natexlab{b}})Chakraborty, Bedi, Koppel,
  Tokekar, and Manocha]{chakraborty2022dealing}
Chakraborty, S., Bedi, A.~S., Koppel, A., Tokekar, P., and Manocha, D.
\newblock Dealing with sparse rewards in continuous control robotics via
  heavy-tailed policies.
\newblock \emph{arXiv preprint arXiv:2206.05652}, 2022{\natexlab{b}}.

\bibitem[Chakraborty et~al.(2022{\natexlab{c}})Chakraborty, Bedi, Koppel,
  Tokekar, and Manocha]{sparse4}
Chakraborty, S., Bedi, A.~S., Koppel, A., Tokekar, P., and Manocha, D.
\newblock Dealing with sparse rewards in continuous control robotics via
  heavy-tailed policies, 2022{\natexlab{c}}.

\bibitem[Chakraborty et~al.(2023{\natexlab{a}})Chakraborty, Bedi, Koppel,
  Sadler, Huang, Tokekar, and Manocha]{ksrl}
Chakraborty, S., Bedi, A.~S., Koppel, A., Sadler, B.~M., Huang, F., Tokekar,
  P., and Manocha, D.
\newblock Posterior coreset construction with kernelized stein discrepancy for
  model-based reinforcement learning, 2023{\natexlab{a}}.

\bibitem[Chakraborty et~al.(2023{\natexlab{b}})Chakraborty, Weerakoon, Poddar,
  Tokekar, Bedi, and Manocha]{sparse1}
Chakraborty, S., Weerakoon, K., Poddar, P., Tokekar, P., Bedi, A.~S., and
  Manocha, D.
\newblock Re-move: An adaptive policy design approach for dynamic environments
  via language-based feedback, 2023{\natexlab{b}}.

\bibitem[Chen et~al.(2019)Chen, Barp, Briol, Gorham, Girolami, Mackey, and
  Oates]{stein_point_Markov}
Chen, W.~Y., Barp, A., Briol, F.-X., Gorham, J., Girolami, M., Mackey, L., and
  Oates, C.
\newblock Stein point markov chain monte carlo.
\newblock In \emph{International Conference on Machine Learning}, pp.\
  1011--1021. PMLR, 2019.

\bibitem[Chowdhury \& Gopalan(2019)Chowdhury and Gopalan]{chowdhury2019online}
Chowdhury, S.~R. and Gopalan, A.
\newblock Online learning in kernelized markov decision processes.
\newblock In \emph{The 22nd International Conference on Artificial Intelligence
  and Statistics}, pp.\  3197--3205, 2019.

\bibitem[Chua et~al.(2018)Chua, Calandra, McAllister, and Levine]{chua2018deep}
Chua, K., Calandra, R., McAllister, R., and Levine, S.
\newblock Deep reinforcement learning in a handful of trials using
  probabilistic dynamics models.
\newblock In \emph{Advances in Neural Information Processing Systems}, pp.\
  4754--4765, 2018.

\bibitem[Dearden et~al.(1998)Dearden, Friedman, and Russell]{dearden_bql}
Dearden, R., Friedman, N., and Russell, S.
\newblock Bayesian q-learning.
\newblock In \emph{Proceedings of the Fifteenth National/Tenth Conference on
  Artificial Intelligence/Innovative Applications of Artificial Intelligence},
  AAAI '98/IAAI '98, pp.\  761–768, USA, 1998. American Association for
  Artificial Intelligence.
\newblock ISBN 0262510987.

\bibitem[Deisenroth \& Rasmussen(2011)Deisenroth and
  Rasmussen]{deisenroth2011pilco}
Deisenroth, M. and Rasmussen, C.~E.
\newblock Pilco: A model-based and data-efficient approach to policy search.
\newblock In \emph{Proceedings of the 28th International Conference on machine
  learning (ICML-11)}, pp.\  465--472, 2011.

\bibitem[Edwards(2019)]{https://doi.org/10.48550/arxiv.1909.05075}
Edwards, J.
\newblock Practical calculation of gittins indices for multi-armed bandits,
  2019.
\newblock URL \url{https://arxiv.org/abs/1909.05075}.

\bibitem[Efron \& Morris(1973)Efron and Morris]{efron1973stein}
Efron, B. and Morris, C.
\newblock Stein's estimation rule and its competitors—an empirical bayes
  approach.
\newblock \emph{Journal of the American Statistical Association}, 68\penalty0
  (341):\penalty0 117--130, 1973.

\bibitem[Eysenbach \& Levine(2021)Eysenbach and Levine]{eysenbach_entropy}
Eysenbach, B. and Levine, S.
\newblock Maximum entropy rl (provably) solves some robust rl problems, 2021.
\newblock URL \url{https://arxiv.org/abs/2103.06257}.

\bibitem[Fan \& Ming(2021)Fan and Ming]{pmlr-v139-fan21b}
Fan, Y. and Ming, Y.
\newblock Model-based reinforcement learning for continuous control with
  posterior sampling.
\newblock In Meila, M. and Zhang, T. (eds.), \emph{Proceedings of the 38th
  International Conference on Machine Learning}, volume 139 of
  \emph{Proceedings of Machine Learning Research}, pp.\  3078--3087. PMLR,
  18--24 Jul 2021.
\newblock URL \url{https://proceedings.mlr.press/v139/fan21b.html}.

\bibitem[Gelfand \& Mitter(1991)Gelfand and Mitter]{gelfand1991recursive}
Gelfand, S.~B. and Mitter, S.~K.
\newblock Recursive stochastic algorithms for global optimization in r\^{}d.
\newblock \emph{SIAM Journal on Control and Optimization}, 29\penalty0
  (5):\penalty0 999--1018, 1991.

\bibitem[Gorham \& Mackey(2015)Gorham and Mackey]{gorham2015measuring}
Gorham, J. and Mackey, L.
\newblock Measuring sample quality with stein's method.
\newblock \emph{Advances in Neural Information Processing Systems}, 28, 2015.

\bibitem[Haarnoja et~al.(2018)Haarnoja, Zhou, Abbeel, and
  Levine]{haarnoja2018soft}
Haarnoja, T., Zhou, A., Abbeel, P., and Levine, S.
\newblock Soft actor-critic: Off-policy maximum entropy deep reinforcement
  learning with a stochastic actor.
\newblock \emph{arXiv preprint arXiv:1801.01290}, 2018.

\bibitem[Hao \& Lattimore(2022)Hao and Lattimore]{hao2022regret}
Hao, B. and Lattimore, T.
\newblock Regret bounds for information-directed reinforcement learning.
\newblock \emph{arXiv preprint arXiv:2206.04640}, 2022.

\bibitem[Hawkins et~al.()Hawkins, Koppel, and Zhang]{hawkins2022online}
Hawkins, C., Koppel, A., and Zhang, Z.
\newblock Online, informative mcmc thinning with kernelized stein discrepancy.
\newblock \emph{arXiv preprint arXiv:2201.07130}.

\bibitem[Hawkins et~al.(2022)Hawkins, Koppel, and Zhang]{cole_koppel}
Hawkins, C., Koppel, A., and Zhang, Z.
\newblock Online, informative mcmc thinning with kernelized stein discrepancy,
  2022.
\newblock URL \url{https://arxiv.org/abs/2201.07130}.

\bibitem[James \& Stein(1992)James and Stein]{james1992estimation}
James, W. and Stein, C.
\newblock Estimation with quadratic loss.
\newblock In \emph{Breakthroughs in statistics}, pp.\  443--460. Springer,
  1992.

\bibitem[Janner et~al.(2019)Janner, Fu, Zhang, and Levine]{janner2019trust}
Janner, M., Fu, J., Zhang, M., and Levine, S.
\newblock When to trust your model: Model-based policy optimization.
\newblock In \emph{Advances in Neural Information Processing Systems}, pp.\
  12498--12509, 2019.

\bibitem[Jin et~al.(2018)Jin, Allen-Zhu, Bubeck, and Jordan]{jin2018q}
Jin, C., Allen-Zhu, Z., Bubeck, S., and Jordan, M.~I.
\newblock Is q-learning provably efficient?
\newblock In \emph{Advances in Neural Information Processing Systems}, pp.\
  4863--4873, 2018.

\bibitem[Jin et~al.(2020)Jin, Yang, Wang, and Jordan]{jin2020provably}
Jin, C., Yang, Z., Wang, Z., and Jordan, M.~I.
\newblock Provably efficient reinforcement learning with linear function
  approximation.
\newblock In \emph{Conference on Learning Theory}, pp.\  2137--2143, 2020.

\bibitem[Jitkrittum et~al.(2020)Jitkrittum, Kanagawa, and Schölkopf]{Wittawat}
Jitkrittum, W., Kanagawa, H., and Schölkopf, B.
\newblock Testing goodness of fit of conditional density models with kernels,
  2020.
\newblock URL \url{https://arxiv.org/abs/2002.10271}.

\bibitem[Koppel et~al.(2021)Koppel, Pradhan, and Rajawat]{koppel2021consistent}
Koppel, A., Pradhan, H., and Rajawat, K.
\newblock Consistent online gaussian process regression without the sample
  complexity bottleneck.
\newblock \emph{Statistics and Computing}, 31\penalty0 (6):\penalty0 1--18,
  2021.

\bibitem[Lai et~al.(1985)Lai, Robbins, et~al.]{lai1985asymptotically}
Lai, T.~L., Robbins, H., et~al.
\newblock Asymptotically efficient adaptive allocation rules.
\newblock \emph{Advances in applied mathematics}, 6\penalty0 (1):\penalty0
  4--22, 1985.

\bibitem[Lattimore \& Szepesvari(2019)Lattimore and
  Szepesvari]{Lattimore_minmax}
Lattimore, T. and Szepesvari, C.
\newblock An information-theoretic approach to minimax regret in partial
  monitoring, 2019.
\newblock URL \url{https://arxiv.org/abs/1902.00470}.

\bibitem[Li et~al.(2008)Li, Littman, and Walsh]{Lihong}
Li, L., Littman, M.~L., and Walsh, T.~J.
\newblock Knows what it knows: A framework for self-aware learning.
\newblock In \emph{Proceedings of the 25th International Conference on Machine
  Learning}, ICML '08, pp.\  568–575, New York, NY, USA, 2008. Association
  for Computing Machinery.
\newblock ISBN 9781605582054.
\newblock \doi{10.1145/1390156.1390228}.
\newblock URL \url{https://doi.org/10.1145/1390156.1390228}.

\bibitem[Lillicrap et~al.(2015)Lillicrap, Hunt, Pritzel, Heess, Erez, Tassa,
  Silver, and Wierstra]{lillicrap_ddpg}
Lillicrap, T.~P., Hunt, J.~J., Pritzel, A., Heess, N., Erez, T., Tassa, Y.,
  Silver, D., and Wierstra, D.
\newblock Continuous control with deep reinforcement learning, 2015.
\newblock URL \url{https://arxiv.org/abs/1509.02971}.

\bibitem[Littlestone(1987)]{Littlestone}
Littlestone, N.
\newblock Learning quickly when irrelevant attributes abound: A new
  linear-threshold algorithm.
\newblock In \emph{28th Annual Symposium on Foundations of Computer Science
  (sfcs 1987)}, pp.\  68--77, 1987.
\newblock \doi{10.1109/SFCS.1987.37}.

\bibitem[Liu et~al.(2019)Liu, Gu, and Liu]{liu_entropy}
Liu, J., Gu, X., and Liu, S.
\newblock Policy optimization reinforcement learning with entropy
  regularization, 2019.
\newblock URL \url{https://arxiv.org/abs/1912.01557}.

\bibitem[Liu et~al.(2016)Liu, Lee, and Jordan]{liu2016kernelized}
Liu, Q., Lee, J., and Jordan, M.
\newblock A kernelized stein discrepancy for goodness-of-fit tests.
\newblock In \emph{International conference on machine learning}, pp.\
  276--284. PMLR, 2016.

\bibitem[Lu \& Van~Roy(2019)Lu and Van~Roy]{lu_info}
Lu, X. and Van~Roy, B.
\newblock Information-theoretic confidence bounds for reinforcement learning,
  2019.
\newblock URL \url{https://arxiv.org/abs/1911.09724}.

\bibitem[Lu et~al.(2021)Lu, Van~Roy, Dwaracherla, Ibrahimi, Osband, and
  Wen]{vanroy_bit}
Lu, X., Van~Roy, B., Dwaracherla, V., Ibrahimi, M., Osband, I., and Wen, Z.
\newblock Reinforcement learning, bit by bit, 2021.
\newblock URL \url{https://arxiv.org/abs/2103.04047}.

\bibitem[Markou \& Rasmussen(2019)Markou and Rasmussen]{markou_mm}
Markou, E. and Rasmussen, C.~E.
\newblock Bayesian methods for efficient reinforcement learning in tabular
  problems, 2019.
\newblock URL
  \url{https://github.com/stratisMarkou/sample-efficient-bayesian-rl}.

\bibitem[Mnih et~al.(2013)Mnih, Kavukcuoglu, Silver, Graves, Antonoglou,
  Wierstra, and Riedmiller]{mnih_qlearn}
Mnih, V., Kavukcuoglu, K., Silver, D., Graves, A., Antonoglou, I., Wierstra,
  D., and Riedmiller, M.
\newblock Playing atari with deep reinforcement learning, 2013.
\newblock URL \url{https://arxiv.org/abs/1312.5602}.

\bibitem[O'Donoghue et~al.(2017)O'Donoghue, Osband, Munos, and
  Mnih]{donoghue_ube}
O'Donoghue, B., Osband, I., Munos, R., and Mnih, V.
\newblock The uncertainty bellman equation and exploration, 2017.
\newblock URL \url{https://arxiv.org/abs/1709.05380}.

\bibitem[Osband \& Van~Roy(2014)Osband and Van~Roy]{osband2014model}
Osband, I. and Van~Roy, B.
\newblock Model-based reinforcement learning and the eluder dimension.
\newblock In \emph{Advances in Neural Information Processing Systems}, pp.\
  1466--1474, 2014.

\bibitem[Osband \& Van~Roy(2017{\natexlab{a}})Osband and Van~Roy]{osband17a}
Osband, I. and Van~Roy, B.
\newblock Why is posterior sampling better than optimism for reinforcement
  learning?
\newblock In Precup, D. and Teh, Y.~W. (eds.), \emph{Proceedings of the 34th
  International Conference on Machine Learning}, pp.\  2701--2710,
  International Convention Centre, Sydney, Australia, 2017{\natexlab{a}}. PMLR.

\bibitem[Osband \& Van~Roy(2017{\natexlab{b}})Osband and
  Van~Roy]{osband2017posterior}
Osband, I. and Van~Roy, B.
\newblock Why is posterior sampling better than optimism for reinforcement
  learning?
\newblock In \emph{International conference on machine learning}, pp.\
  2701--2710. PMLR, 2017{\natexlab{b}}.

\bibitem[Osband et~al.(2013)Osband, Benjamin, and Daniel]{osband2013}
Osband, I., Benjamin, V.~R., and Daniel, R.
\newblock ({M}ore) efficient reinforcement learning via posterior sampling.
\newblock In \emph{Proceedings of the 26th International Conference on Neural
  Information Processing Systems - Volume 2}, NIPS'13, pp.\  3003--3011, USA,
  2013. Curran Associates Inc.

\bibitem[Osband et~al.(2019)Osband, Van~Roy, Russo, and Wen]{osband2019deep}
Osband, I., Van~Roy, B., Russo, D.~J., and Wen, Z.
\newblock Deep exploration via randomized value functions.
\newblock \emph{Journal of Machine Learning Research}, 20\penalty0
  (124):\penalty0 1--62, 2019.

\bibitem[Ozair et~al.(2019)Ozair, Lynch, Bengio, Oord, Levine, and
  Sermanet]{ozair_wasserstein}
Ozair, S., Lynch, C., Bengio, Y., Oord, A. v.~d., Levine, S., and Sermanet, P.
\newblock Wasserstein dependency measure for representation learning, 2019.
\newblock URL \url{https://arxiv.org/abs/1903.11780}.

\bibitem[Pathak et~al.(2017{\natexlab{a}})Pathak, Agrawal, Efros, and
  Darrell]{curiosity1}
Pathak, D., Agrawal, P., Efros, A.~A., and Darrell, T.
\newblock Curiosity-driven exploration by self-supervised prediction,
  2017{\natexlab{a}}.
\newblock URL \url{https://arxiv.org/abs/1705.05363}.

\bibitem[Pathak et~al.(2017{\natexlab{b}})Pathak, Agrawal, Efros, and
  Darrell]{pathak_curiosity}
Pathak, D., Agrawal, P., Efros, A.~A., and Darrell, T.
\newblock Curiosity-driven exploration by self-supervised prediction,
  2017{\natexlab{b}}.
\newblock URL \url{https://arxiv.org/abs/1705.05363}.

\bibitem[Pathak et~al.(2019)Pathak, Gandhi, and Gupta]{curiosity3}
Pathak, D., Gandhi, D., and Gupta, A.
\newblock Self-supervised exploration via disagreement, 2019.
\newblock URL \url{https://arxiv.org/abs/1906.04161}.

\bibitem[Raginsky et~al.(2017)Raginsky, Rakhlin, and
  Telgarsky]{raginsky2017non}
Raginsky, M., Rakhlin, A., and Telgarsky, M.
\newblock Non-convex learning via stochastic gradient langevin dynamics: a
  nonasymptotic analysis.
\newblock In \emph{Conference on Learning Theory}, pp.\  1674--1703. PMLR,
  2017.

\bibitem[Rengarajan et~al.(2022)Rengarajan, Vaidya, Sarvesh, Kalathil, and
  Shakkottai]{sparse_rwd1}
Rengarajan, D., Vaidya, G., Sarvesh, A., Kalathil, D., and Shakkottai, S.
\newblock Reinforcement learning with sparse rewards using guidance from
  offline demonstration, 2022.
\newblock URL \url{https://arxiv.org/abs/2202.04628}.

\bibitem[Russo \& Van~Roy(2014{\natexlab{a}})Russo and
  Van~Roy]{russo2014learning}
Russo, D. and Van~Roy, B.
\newblock Learning to optimize via posterior sampling.
\newblock \emph{Mathematics of Operations Research}, 39\penalty0 (4):\penalty0
  1221--1243, 2014{\natexlab{a}}.

\bibitem[Russo \& Van~Roy(2014{\natexlab{b}})Russo and Van~Roy]{russo_info}
Russo, D. and Van~Roy, B.
\newblock An information-theoretic analysis of thompson sampling,
  2014{\natexlab{b}}.
\newblock URL \url{https://arxiv.org/abs/1403.5341}.

\bibitem[Russo \& Van~Roy(2018)Russo and Van~Roy]{russo2018learning}
Russo, D. and Van~Roy, B.
\newblock Learning to optimize via information-directed sampling.
\newblock \emph{Operations Research}, 66\penalty0 (1):\penalty0 230--252, 2018.

\bibitem[Russo et~al.(2017)Russo, Van~Roy, Kazerouni, Osband, and
  Wen]{thompson_tutorial}
Russo, D., Van~Roy, B., Kazerouni, A., Osband, I., and Wen, Z.
\newblock A tutorial on thompson sampling.
\newblock 2017.
\newblock \doi{10.48550/ARXIV.1707.02038}.
\newblock URL \url{https://arxiv.org/abs/1707.02038}.

\bibitem[Schulman et~al.(2017)Schulman, Wolski, Dhariwal, Radford, and
  Klimov]{schulman2017proximal}
Schulman, J., Wolski, F., Dhariwal, P., Radford, A., and Klimov, O.
\newblock Proximal policy optimization algorithms.
\newblock \emph{arXiv preprint arXiv:1707.06347}, 2017.

\bibitem[Shyam et~al.(2018)Shyam, Jaśkowski, and Gomez]{shyam_sparse}
Shyam, P., Jaśkowski, W., and Gomez, F.
\newblock Model-based active exploration, 2018.
\newblock URL \url{https://arxiv.org/abs/1810.12162}.

\bibitem[Sriperumbudur et~al.(2010)Sriperumbudur, Gretton, Fukumizu,
  Sch{\"o}lkopf, and Lanckriet]{sriperumbudur2010hilbert}
Sriperumbudur, B.~K., Gretton, A., Fukumizu, K., Sch{\"o}lkopf, B., and
  Lanckriet, G.~R.
\newblock Hilbert space embeddings and metrics on probability measures.
\newblock \emph{The Journal of Machine Learning Research}, 11:\penalty0
  1517--1561, 2010.

\bibitem[Sriperumbudur et~al.(2012)Sriperumbudur, Fukumizu, Gretton,
  Sch{\"o}lkopf, and Lanckriet]{sriperumbudur2012empirical}
Sriperumbudur, B.~K., Fukumizu, K., Gretton, A., Sch{\"o}lkopf, B., and
  Lanckriet, G.~R.
\newblock On the empirical estimation of integral probability metrics.
\newblock \emph{Electronic Journal of Statistics}, 6:\penalty0 1550--1599,
  2012.

\bibitem[Valiant(1984)]{Valiant}
Valiant, L.~G.
\newblock A theory of the learnable.
\newblock \emph{Commun. ACM}, 27\penalty0 (11):\penalty0 1134–1142, nov 1984.
\newblock ISSN 0001-0782.
\newblock \doi{10.1145/1968.1972}.
\newblock URL \url{https://doi.org/10.1145/1968.1972}.

\bibitem[Watkins \& Dayan(1992)Watkins and Dayan]{Watkins1992}
Watkins, C. J. C.~H. and Dayan, P.
\newblock Q-learning.
\newblock \emph{Machine Learning}, 8\penalty0 (3):\penalty0 279--292, May 1992.
\newblock ISSN 1573-0565.
\newblock \doi{10.1007/BF00992698}.
\newblock URL \url{https://doi.org/10.1007/BF00992698}.

\bibitem[Weerakoon et~al.(2022{\natexlab{a}})Weerakoon, Chakraborty,
  Karapetyan, Sathyamoorthy, Bedi, and Manocha]{sparse2}
Weerakoon, K., Chakraborty, S., Karapetyan, N., Sathyamoorthy, A.~J., Bedi,
  A.~S., and Manocha, D.
\newblock Htron:efficient outdoor navigation with sparse rewards via heavy
  tailed adaptive reinforce algorithm, 2022{\natexlab{a}}.

\bibitem[Weerakoon et~al.(2022{\natexlab{b}})Weerakoon, Chakraborty,
  Karapetyan, Sathyamoorthy, Bedi, and Manocha]{weerakoon_sparse}
Weerakoon, K., Chakraborty, S., Karapetyan, N., Sathyamoorthy, A.~J., Bedi,
  A.~S., and Manocha, D.
\newblock Htron:efficient outdoor navigation with sparse rewards via heavy
  tailed adaptive reinforce algorithm, 2022{\natexlab{b}}.
\newblock URL \url{https://arxiv.org/abs/2207.03694}.

\bibitem[Yang et~al.(2018)Yang, Liu, Rao, and Neville]{yang18c}
Yang, J., Liu, Q., Rao, V., and Neville, J.
\newblock Goodness-of-fit testing for discrete distributions via stein
  discrepancy.
\newblock In Dy, J. and Krause, A. (eds.), \emph{Proceedings of the 35th
  International Conference on Machine Learning}, volume~80 of \emph{Proceedings
  of Machine Learning Research}, pp.\  5561--5570. PMLR, 10--15 Jul 2018.
\newblock URL \url{https://proceedings.mlr.press/v80/yang18c.html}.

\bibitem[Yang \& Wang(2020)Yang and Wang]{yang2020reinforcement}
Yang, L. and Wang, M.
\newblock Reinforcement learning in feature space: Matrix bandit, kernels, and
  regret bound.
\newblock In \emph{International Conference on Machine Learning}, pp.\
  10746--10756. PMLR, 2020.

\bibitem[Zanette et~al.(2020)Zanette, Brandfonbrener, Brunskill, Pirotta, and
  Lazaric]{zanette2020frequentist}
Zanette, A., Brandfonbrener, D., Brunskill, E., Pirotta, M., and Lazaric, A.
\newblock Frequentist regret bounds for randomized least-squares value
  iteration.
\newblock In \emph{International Conference on Artificial Intelligence and
  Statistics}, pp.\  1954--1964. PMLR, 2020.

\bibitem[Zimmert \& Lattimore(2019)Zimmert and Lattimore]{Zimmert_info}
Zimmert, J. and Lattimore, T.
\newblock Connections between mirror descent, thompson sampling and the
  information ratio, 2019.
\newblock URL \url{https://arxiv.org/abs/1905.11817}.

\end{thebibliography}
\bibliographystyle{icml2022}

\newpage
\onecolumn
\appendix
\addcontentsline{toc}{section}{Appendix} 
\part{Appendix} 
\parttoc 


\section{Detailed Context of Related Works}\label{related_works}

    \textbf{Model based and Model Free RL.} Most RL algorithms fall into two categories: \emph{model-free} \citep{schulman2017proximal, mnih_qlearn, haarnoja2018soft,lillicrap_ddpg} and \emph{model-based} \citep{pmlr-v139-fan21b, deisenroth2011pilco, chua2018deep, janner2019trust}. In model-free approaches, the agent learns direct policy mapping from states to action with approximate dynamic programming methods. In contrast, in model-based approaches, an agent learns the approximate model of the environment itself and trains a policy under the learned dynamics. Recently, probabilistic model-based RL algorithms have shown superior performance in practice relative to their model-free counterparts, despite the strong conceptual guarantees for model-free approaches \citep{janner2019trust}. We focus on model-based RL in this work. 
        
\textbf{Performance Measure.}        To understand why this may be so, it's important to assess the convergence criteria of RL methods: Probably approximate correct (PAC) bounds \citep{Valiant}, Frequentist regret \citep{jin2018q, jin2020provably}, Bayesian regret \citep{osband2013}, Mistake (MB) Bound \citep{Littlestone}, KWIK (Knows What It Knows) \citep{Lihong} $\&$ convergence in various distributional metrics \citep{amortila2020distributional, borkar2002risk, chowdhury2019online, ksrl} all exist. A crucial challenge lies in deciding the optimal selection criteria to evaluate the algorithm's computational and statistical efficiency. Rather than comment on the specific merit of a particular convergence criterion as a motivation for our restriction of focus to Bayesian regret, we note a few of its salient attributes: it imposes minimal requirements on access to a generative model underlying state transitions  \citep{osband17a, osband2013, osband2014model}, and respects the inherent uncertainty associated with the optimal policy, rather than supposing that it can be effectively captured by confidence sets based on a few moment-based estimates of the transition dynamics, which can lead to undesirable behavior in the presence of sparse rewards \citep{jin2020provably, yang2020reinforcement, zanette2020frequentist}. In addition, it has been observed in several practical scenarios that the performance degrades drastically with sparse rewards \citep{sparse1, sparse2, sparse3, sparse4}.
        
\textbf{Information-theoretic Approaches.}         To encapsulate the inherent uncertainty present in the optimal policy, one may augment notions of regret to quantify distance to the optimal occupancy measure or other information-theoretic quantities \citep{russo2018learning,hao2022regret}. That this is advantageous may be seen by honing in on the sparse reward setting: consider the traditional definition of regret $\mathbb{E}[V^{*}(s) - V^{k}(s)]$ for any $k^{th}$ episode in an environment with near-zero rewards. Suppose one policy incorporates exploration based on a distributional estimate of the environment, whereas the other only consider moments of the distribution of returns, such as UCB \citep{lai1985asymptotically}. In this case, the traditional notion of regret may not encourage exploration in a way that yields increased state-space coverage, as the value distribution for this case would be near-null. This issue is well-documented in the bandit setting  \citep{thompson_tutorial}. Hence, there is an intrinsic motivation to consider augmentations of regret that are well-calibrated to the inherent uncertainty of the optimal policy. Motivated by \citep{russo2018learning,hao2022regret}, we consider convergence criteria from information theory and Bayesian inference to define an appropriate notion of regret.
       To understand the exact manner in which these modifications are incorporated into model-based RL (MBRL), we contrast them with the model-free setting. In such settings, incorporating exploration bonuses is well-established \citep{jin2020provably,eysenbach_entropy,liu_entropy,zafarali_entropy}, either in the form of augmenting the reward, the value function \citep{jin2018q,osband2019deep}, or the policy gradient \citep{gelfand1991recursive,raginsky2017non}. However, such methods sample uniformly with respect to a value or policy rather than in pursuit of reducing the estimation error to the optimal transition dynamics, which can yield spurious behavior \citep{weerakoon_sparse, shyam_sparse}. By contrast, information-theoretic regularisation in model-based RL is an active area of research. Empirical advancements based on intrinsic curiosity \citep{burda_curiosity,pathak_curiosity}, i.e., modifying the sampling probabilities driven by uncertainty estimates in the transition dynamics or forward model prediction error, can improve performance in practice but lack conceptual guarantees. 
        
        \textbf{Information Directed Sampling.} To substantiate these approaches conceptually, the resultant algorithms have recently been rewritten in a way that their performance can be quantified by information-theoretic or Bayesian regret \citep{lu_info, vanroy_bit}, under a specific choice of Dirichlet priors for the transition model, inspired by earlier work on bandits \citep{russo_info}. Extensions that alleviate any requirements on the prior for MBRL also exist based upon the development of a surrogate environment estimation procedure via rate-distortion theory   \citep{hao2022regret}. Unfortunately, the resultant algorithm requires estimating the information gain, which is generally intractable. This issue can be partially addressed by instead optimizing the evidence lower-bound (ELBO) \citep{achiam_lb}, but exhibits exponential dependence on the mutual information with respect to the optimal occupancy measure \citep{ozair_wasserstein}. Related approaches replace mutual information by Bregman divergence; however, this necessitates inverting a Fisher information matrix per step which can be computationally costly \citep{Lattimore_minmax, Zimmert_info}.
        Hence, previous efforts to incorporate information-theoretic bonuses in MBRL either impose restrictive assumptions on the prior or yield computationally heavy objectives whose algorithmic solutions exhibit scalability problems.
 \section{Preliminaries: Kernelized Stein Discrepancy}\label{KSD_existing_appenndix}

Consider the notion of Integral probability metric to measure  the deviation between the estimated distribution $q$ and the unknown target distribution $p$ defined as 
	  \begin{align}\label{IPMs_appenndix}
		      d_{\mathcal{F}}(q, p) = \sup_{f \in \mathcal{F}} |\mathbb{E}_{q} [f(X)] - \mathbb{E}_{p} [f(X)]|,
		  \end{align}
		  where the supremum is over a class of real-valued test functions $f\in\mathcal{F}$. By adjusting the function class $\mathcal{F}$, we can recover the well-known metrics such as  Total variation distance, Wasserstein distance \citep{sriperumbudur2010hilbert}, etc. 
	  However, the major challenge in evaluating the IPM in \eqref{IPMs_appenndix} is that it requires an integration under the true distribution {$p$} which is intractable. A seminal idea to alleviate this issue is called Stein's method, which restricts the class of distributions  $\mathcal{F}$ to functions such that {$\mathbb{E}_{p}[f(X)] = 0$}. Building upon this idea, \citep{liu2016kernelized} develops a tractable way to evaluate the IPM by restricting distributions to the Stein class, associated with a reproducing kernel Hilbert space (RKHS) over Stein kernels \citep{berlinet2011reproducing}. In this case, the IPM can be evaluated in terms of the Stein kernel as the kernelized stein discrepancy \citep{gorham2015measuring}. Stein's method provides a generalised framework for studying distributional distances and relies on the fact that two smooth densities $p(x)$ and $q(x)$ are identical iff they satisfy the Stein's identity given by
	  \begin{align}\label{stein_identity_appenndix}
		  \max_{f\in \mathcal F} \big(\mathbb E_p[ s_q(x) f(x) + 
		  \nabla_x f(x)]\big)^2 = 0, 
		  \end{align}
	 where $s_q(x)$ denotes the score function of $q(x)$ given by $s_q(x) =  \nabla_x \log q(x)$. As an example, Stein's identity in \eqref{stein_identity_appenndix} holds for smooth functions $f$ lying in the Stein class of $p$. A function $f$ is in the Stein class of $p$ if it's smooth and satisfies $\int_{x} \nabla_{x} (f(x) p(x)) dx =0$. Hence, for any function $f$ in the Stein class of $p$, we can say $ \mathbb E_p [\mathcal{A}_p f(x)] = 0 $ where $\mathcal{A}_p$ is the Stein operator of $p$ which is a linear operator.
	
	  From here the Stein discrepancy between $p$ and $q$ is defined as \citep{liu2016kernelized}
	  \begin{align}\label{stein_disc_defn_appenndix}
		   \text{KSD}^2 (p,q)  = \max_{f\in \mathcal F} \big(\mathbb E_p[ s_q(x) f(x) + 
		  \nabla_x f(x)]\big)^2, 
		  \end{align}
	  where $\mathcal F$ is a class of smooth functions satisfying Stein's identity \eqref{stein_identity_appenndix}. However, the above definition is computationally intractable as  it requires solving a complex variational optimization. To this end,  \citep{liu2016kernelized} define a computationally tractable modification as 
	  \begin{align}\label{jordan_stein_appenndix}
		   {\KSD}(p,q)  =  \mathbb E_{x,x'\sim p } \big [  u_q(x,x') \big],
		 \end{align}
	  where $ u_q(x,x')$ is the Stein kernel defined as 
	  \begin{align}
		  u_q & (x,x') := s_q(x)^\top \kappa(x,x') s_q(x')  +   s_q(x)^\top \nabla_{x'}\kappa(x,x')  
\nonumber
\\  &+\nabla_{x}\kappa(x,x')^\top   s_q(x') + Tr(\nabla_{x,x'}\kappa(x,x')),\nonumber
		  \end{align}
    where $\kappa(x,x')$ is the base kernel.  
    
    For the setting in this work, we have $p=P^*$ (transition dynamics corresponding to true model $M^*$) and $q=P^{M_k}$ (transition dynamics corresponding to posterior $M_k \sim \phi(\cdot|\mathcal{H}_k)$). Interestingly, KSD empowers us to evaluate the distance $\KSD(P^{M_k},P^*)$.

     \section{Proof of Proposition \ref{lemma_KCDSD}}\label{appendix_lemma_1_proof}
\begin{proof}

	Let us first define the compact notation such that $G_{{x}}:=G{(x,\cdot)}$ and $\xi_{P_{y| x}^{M^k}}({y},\cdot) := s_{P^{M^k}}( y) l( y, \cdot) - \triangle^{\ast} l( y,\cdot)$. 
        Here, we derive our proposed DSD as defined in \eqref{eq:kssd_pop_ustat}. Further, for simplicity of analysis, we denote $P^*(s'|s,a) \rightarrow P^*_{(y|x)} (y|x)$, $P^*(s,a) \rightarrow P^*_{x} (x)$ and $P^*(s,a,s') \rightarrow P^*_{(x,y)} (x,y)$, state-action pair $(s,a) \rightarrow {x} \in \mathcal{X}:=\mathcal{S}\times \mathcal{A}$ and the corresponding next state $s'  \rightarrow {y} \in \mathcal{S}$. Also, for simplicity of notation we denote $P^{M^*} \rightarrow P^*$ and $P^{M^k} \rightarrow P^k$. To start the proof, let us consider ${\text{DSD}(P^k,P^*)}$ and write
	  \begin{align}\label{kcdsd:op_proof}
		  {\text{DSD}(P^k,P^*)} & =\big\|\mathbb{E}_{(\mathbf{x},\mathbf{y})\sim P^*_{x,y}}G_{\mathbf{x}}\xi_{P_{y|x}^k}(\mathbf{y},\cdot)\big\|^2
    \\
		   & =\left\langle \mathbb{E}_{\mathbf{(x, y)} \sim P^*_{(x,y)}}G_{\mathbf{x}}\xi_{P_{y|x}^k}(\mathbf{y},\cdot),\mathbb{E}_{\mathbf{(x', y')} \sim P^*_{(x,y)}}G_{\mathbf{x'}}\xi_{P_{y'|x'}^k}(\mathbf{y}',\cdot)\right\rangle \\
		   & =\mathbb{E}_{\mathbf{(x, y)} \sim P^*_{(x,y)}}\mathbb{E}_{\mathbf{(x', y')} \sim P^*_{(x,y)}}\left\langle G_{\mathbf{x}}\xi_{P_{y|x}^k}(\mathbf{y},\cdot),G_{\mathbf{x'}}\xi_{P_{y'|x'}^k}(\mathbf{y}',\cdot)\right\rangle \\
		   & =\mathbb{E}_{\mathbf{(x, y)} \sim P^*_{(x,y)}}\mathbb{E}_{\mathbf{(x', y')} \sim P^*_{(x,y)}}\left\langle G_{\mathbf{x}'}G_{\mathbf{x}}\xi_{P_{y|x}^k}(\mathbf{y},\cdot),\xi_{P_{y'|x'}^k}(\mathbf{y}',\cdot)\right\rangle\\
		   &=\mathbb{E}_{\mathbf{(x, y)} \sim P^*_{(x,y)}}\mathbb{E}_{\mathbf{(x',y')}\sim P^*_{(x,y)}} [k(\mathbf{x},\mathbf{x}')\kappa_{P}((\mathbf{x},\mathbf{y}),(\mathbf{x}',\mathbf{y}'))]\\
		  &=\mathbb{E}_{\mathbf{(x, y)} \sim P^*_{(x,y)}}\mathbb{E}_{\mathbf{(x',y')}\sim P^*_{(x,y)}} [\kappa_k((\mathbf{x},\mathbf{y}),(\mathbf{x}',\mathbf{y}'))].
		  \end{align}
	Here we have applied the reproducing property of kernels with the linearity of expectations to derive the equations. {where, $G_{\mathbf{x}'}G_{\mathbf{x}} = G(x,x') = k(x,x')I$.}
	  This proves the derivation for an equivalent Stein operator for our scenario. Now, we need to show a version of Stein's identity to complete the derivation of our proposed Kernelized Conditional Discrete Stein Discrepancy. We note that 
	  \begin{align}
		      \mathbb E_{x,y \sim P^*_{(x,y)}} \kappa_k((x,y),\cdot) &= \mathbb  E_{x,y \sim P^*_{(x,y)}} G_x \xi_{P_{y|x}^k}(\mathbf{y},\cdot)\\
		      & =    \mathbb E_{x \sim P^*_{x}} G_x \mathbb E_{y \sim P^*_{(y|x)}} \xi_{P_{y|x}^k}(\mathbf{y},\cdot).
		  \end{align}
	  Now, we show that if $P^k {(y|x)} = P^* {(y|x)}$, $   \mathbb E_{y \sim P^*_{y|x}} \xi_{P_{y|x}^k}(\mathbf{y},\cdot) = 0$ which proves an equivalent notion of Stein's identity for our Discrete conditional case. Replacing $P^k = P^*$.
	  \begin{align}
		        \mathbb E_{y \sim P^*_{(y|x)}} \xi_{P_{y|x}^k}(\mathbf{y},\cdot) &=   \mathbb E_{y \sim P^*_{(y|x)}} [s_{P^*_{(y|x)}}(y) l(y, \cdot) - \triangle^{\ast} l(y,\cdot)]\\
		      &= \sum_{y} [s_{P^*_{(y|x)}}(y) l(y, \cdot) P^*(y|x) -  \triangle^{\ast} l(y, \cdot) P^*(y|x)]\\
		      &= \sum_{y} [\triangle P^*(y|x) l(y, \cdot)   -   \triangle^{\ast} l(y, \cdot) P^*(y|x)]\label{kcdsd_stn_idn_p1}.
		  \end{align}
    Here the first equality holds by denoting $\xi_{P_{y|x}^k} =  \big[s_{P^{k}}( y) l( y, \cdot) - \triangle^{\ast} l( y,\cdot)\big]$ from equation \eqref{our_kernel}. Then expanding upon the expectation and replacing the expression of the score function from equation  \eqref{def_disc_op}, we get the final expression.
	  For each $i$ we can write the first and second part of \eqref{kcdsd_stn_idn_p1} from the definition of difference operators in equation \eqref{def_disc_op} as
	  \begin{align}\label{kcdsd_stn_idn_p2}
		\sum_{y} [\triangle_{y_i}P^*(y|x) l(y,\cdot) ] &= \sum_{y} [l(y,\cdot) P^*(y|x)  - l(y,\cdot) P^*(\lor_i y|x)],\\
          \sum_{y} [\triangle^{\ast} l(y,\cdot) P^*(y|x)] &= \sum_{y} [l(y,\cdot)P^*(y|x) - l(\land_i y, \cdot)P^*(y|x)].
		  \end{align}
%
	 The two equations are equal since $\lor$ and $\land$ are inverse cyclic permutations on $\mathcal{S}$ with $\land_i(\lor_i y) = \lor_i(\land_i y)  = y$ and hence substituting equation \eqref{kcdsd_stn_idn_p2} into equation \eqref{kcdsd_stn_idn_p1} we get $\mathbb E_{y \sim P^*(s'|s,a)} \xi_{{y|x}}(\mathbf{y},\cdot) = 0$. So, this proves an equivalent notion of Stein's identity which completes the proof. 
	\end{proof}
	
    \section{Proof of Theorem \ref{theorem_0}}\label{proof_theorem_0}
	  \begin{proof}
	  	  Let us start with the definition of Bayesian regret defined in \eqref{bayesian_Regret} as follows
		  \begin{align}
			       \BR_K &= \sum_{k=1}^K\mathbb E \left[\mathbb E_{k}\left[V_{1,\pi^*}^{M^*}(s_1^k)-V_{1,\pi^{k}_{\texttt{TS}}}^{M^*}(s_1^k)\right]\right],
			  \end{align}
		  where the inner expectation
		  is over the posterior distribution, and the outer expectation is over the stochastic policy and environment $M^*$. For brevity of notation, let us define {$\mathcal{R}_k:=\mathbb E_k\left[V_{1,\pi^*}^{M^*}(s_1^k)-V_{1,\pi^{k}_{\texttt{TS}}}^{M^*}(s_1^k)\right]$}. From here onwards, we use $\pi^{k}_{\texttt{TS}} \rightarrow \pi^{k}$ to represent the Posterior sampling policy for notation simplicity. Next, we introduce the Stein information ratio via multiplying and dividing by {$\mathbb K_k^{\pi}\left({M^*}; \cH_{k, H}\right)$} as follows
            {
		  \begin{align}
			       \BR_K &= \sum_{k=1}^K \mathbb E\left[\sqrt{\frac{\left(\mathcal{R}_k\right)^2}{\mathbb K_k^{\pi}\left({M^*}; \cH_{k, H}\right)}\mathbb K_k^{\pi}\left({M^*}; \cH_{k, H}\right)}\right]. \nonumber
			  \end{align}}
		  After applying Cauchy–Schwartz inequality,  using the linearity of expectations, and considering the definition of $\Gamma_k^{\text{DSD}}(\pi)$ in \eqref{def:information_ratio}, we can get
		  \begin{align}\label{Bayesian-regret-stein_Ration}
			       \BR_K 
			       &\leq \sqrt{\mathbb E\left[\sum_{k=1}^K\Gamma_k^{\text{\text{DSD}}}(\pi^k)\right]}\sqrt{\mathbb E\left[\sum_{k=1}^K \mathbb K_k^{\pi}\left({M^*}; \cH_{k, H}\right)\right]}\nonumber
			       \\
			       &= \sqrt{\mathbb E\bigg[\sum_{k=1}^K\mathbb K_k^{\pi}\left({M^*}; \cH_{k, H}\right)\bigg] \sum_{k=1}^K\mathbb E[\Gamma_k^{\text{DSD}}(\pi^k)]}\,.
			  \end{align}
		  From definition of $\Gamma_k^{\text{DSD}}(\pi^k)$, we have $\Gamma_k^{\text{DSD}}(\pi^k)\leq {\Gamma}^*$ for any $k\in[K]$. Hence, from \eqref{Bayesian-regret-stein_Ration}, we can write \eqref{final_bayes_regret}. 
%
	  \end{proof}

 \section{Proof of Lemma \ref{lemma_KSD}}\label{proof_lemma_KCSD}
  \begin{proof}
	
	  We run a local optimization procedure by dividing the total number of samples $H$ in an episode into batches of size $Z$ with $Z':=\frac{H}{Z}$ batches and select Stein optimal points per batch using an SPMCMC style update. We begin the analysis by representing the dictionary till the $k^{th}$ episode as $\mathcal{D}_k$ and we expand upon the definition of DSD \eqref{eq:kssd_pop_ustat} as

	  \begin{align}
		          |\mathcal{D}_k|^2\text{DSD}^2(P^k;{\mathcal{D}}_{k}) =& \sum_{(x_i, y_i) \in \mathcal{D}_k}\sum_{(x_j, y_j)\in \mathcal{D}_k}  \kappa_k((x_i, y_i),(x_j, y_j))
		          \label{here_11} 
		          \\
		          =&|{\mathcal{D}}_{k-1}|^2\text{DSD}^2(P^{k-1};{\mathcal{D}}_{k-1})
            \nonumber
            \\
            &+\sum_{z=1}^{Z'}\bigg[\kappa_k((x_k^{z}, y_k^{z}),(x_k^{z}, y_k^{z}))+ 2 \sum_{(x_i, y_i) \in {\mathcal{D}}_{k-1}}\kappa_k((x_i, y_i),(x_k^{z}, y_k^{z}))\bigg]. 
		          \label{here_12}
		          \end{align}

           {In the above expression, equality in \eqref{here_11} comes from the empirical definition of DSD, where $\kappa_{M_k}$ denoted as $\kappa_k$ is the Stein kernel which depend upon the score function of $P^{M_k}$  (cf. \eqref{our_kernel}). }
            Next, for each $z$, we select the sample $(x_k^{z}, y_k^{z})$ from $\mathcal{Y}_z:=\{(x_k^{l}, y_k^{l})\}_{l=1}^Z$ using an SPMCMC style local optimization procedure as in \citep[Appendix A.1]{stein_point_Markov}). Now, from the SPMCMC-based selection, we can write 
	          \begin{align}
		              \kappa_k((x_k^{z}, y_k^{z}),(x_k^{z}, y_k^{z}))+ 2 \sum_{(x_i, y_i) \in {\mathcal{D}}_{k-1}}& \kappa_k((x_i, y_i),(x_k^{z}, y_k^{z})) 
                \nonumber
                \\
                &=  \inf_{(x_k^{z}, y_k^{z}) \in\mathcal{Y}_m} \kappa_k((x_k^{z}, y_k^{z}),(x_k^{z}, y_k^{z}))+  2\sum_{(x_i, y_i)\in {\mathcal{D}}_{k-1}}\kappa_k((x_i, y_i),(x_k^{z}, y_k^{z}))
		              \nonumber
		              \\
		              &\leq  B^2+2\inf_{(x_k^{z}, y_k^{z}) \in\mathcal{Y}_z} \sum_{(x_i, y_i) \in {\mathcal{D}}_{k-1}}\kappa_k((x_i, y_i),(x_k^{z}, y_k^{z})). \label{inf_upper_bound}
		          \end{align}
	        The inequality in \eqref{inf_upper_bound} holds because we restrict our attention to  regions for which it holds that $\kappa_k((x,y),(x,y))\leq B^2$  for all ${(x,y)}\in\mathcal{Y}_k^z$ for all $k$ and $z$. Utilizing the upper bound of \eqref{inf_upper_bound} into the right hand side of \eqref{here_12}, we get
	          \begin{align}
		          |{\mathcal{D}}_k|^2\text{DSD}(P^k;{\mathcal{D}}_{k})^2  \leq & |\mathcal{D}_{k-1}|^2\text{DSD}(P^{k-1};{\mathcal{D}}_{k-1})^2+
		          {Z' B^2}
            \nonumber
            \\
            &+2\sum_{z=1}^{Z'}\inf_{\mat{h}_k^z\in\mathcal{Y}_z} \sum_{(x_i, y_i) \in {\mathcal{D}}_{k-1}}\kappa_k((x_i, y_i),(x_k^{z}, y_k^{z})).
		          \label{here_13}
		  \end{align}
	
	  From the application of Theorem 5 \citep{hawkins2022online} for our formulation with $H$ new samples in the dictionary. 
	  \begin{align}
		      2\inf_{(x_k^{m}, y_k^{m}) \in\mathcal{Y}_m} \sum_{(x_i, y_i) \in {\mathcal{D}}_{k-1}}\kappa_k((x_i, y_i),(x_k^{m}, y_k^{m}))\leq r_k\|f_k\|^2_{\mathcal{K}_0}+\frac{\text{DSD}(P^{k-1};{\mathcal{D}}_{k-1})^2}{r_k},
		      \label{here_14}
		  \end{align}
	  for any arbitrary constant $r_k>0$ and any $f_k = \sum {\kappa_k(x_i, y_i)} , \cdot)$ which lies in RKHS spanned by the kernel $\kappa_k((x_i, y_i),(x_j, y_j))$ can be trivially upper-bounded as $\|f_k\|^2_{\mathcal{K}_0} \leq C_0 S^2 A$, for any choice of kernel where $C_0$ is a positive constant. Hence, we can use the upper bound in \eqref{here_14} to the right hand side of \eqref{here_13}, to obtain 
	  \begin{align}
		           |{\mathcal{D}}_k|^2{\DSD}(P^k;{\mathcal{D}}_{k})^2  \leq|\mathcal{D}_{k-1}|^2\left(1+\frac{Z'}{r_k}\right) {\DSD}(P^{k-1};{\mathcal{D}}_{k-1})^2+Z'{(B^2+r_k C_0 S^2 A)}.
		          \end{align}
	          Next, we divide the both sides by $|{\mathcal{D}}_k|^2=(|{\mathcal{D}_{k-1}}|+ Z')^2$ to obtain
	          \begin{align}
		         {\DSD}(P^k;{\mathcal{D}}_{k})^2 \leq \frac{|\mathcal{D}_{k-1}|^2}{\left(|\mathcal{D}_{k-1}|+  Z' \right)^2}\left(1+\frac{ Z'}{r_k}\right){\DSD}(P^{k-1};{\mathcal{D}}_{k-1})^2+\frac{ Z'\left(B^2+r_kC_0 S^2 A\right)}{\left(|\mathcal{D}_{k-1}|+ Z'\right)^2}\label{thinned}
		  \end{align}
	  It is interesting that a novel aspect in analysis lies in establishing a recursive relationship for the DSD amongst iterations which eventually paves the way to establish the DSD convergence results. After unrolling the recursion in \eqref{thinned}, we can write 
	  \begin{align}
		  {\DSD}(P^k;{\mathcal{D}}_{k})^2 \leq \sum_{i=1}^{k} \left( \frac{ Z'\big(B^2+r_iC_0 S^2 A\big)}{\left(|\mathcal{D}_{i-1}|+ Z'\right)^2}+ \epsilon_i \right)\left(\prod_{j=i}^{k-1} \frac{|\mathcal{D}_j|}{|\mathcal{D}_j|+ Z'}\right)^2 \left( \prod_{j=i}^{k-1} \left(1+\frac{ Z'}{r_{j+1}}\right)\right).  \label{KSD_bound0}
		  \end{align}
	  Applying the log-sum exponential bound $  \prod_{j=i}^{k-1}\left(1+\frac{ Z'}{r_{j+1}}\right)\leq \exp\left( Z'\sum_{j=1}^n \frac{1}{r_j}\right)$, we can write \eqref{KSD_bound0} as
	  \begin{align}
		  \label{eq: initial recursion with r 0}
		 {\DSD}(P^k;{\mathcal{D}}_{k})^2 &
		  \leq \exp\left( Z'\sum_{j=1}^k \frac{1}{r_j}\right)\sum_{i=1}^{k} \left( \frac{ Z'(B^2+r_iC_0 S^2A)}{\left(|\mathcal{D}_{i-1}|+ Z'\right)^2}+ \right)\left(\prod_{j=i}^{k-1} \frac{|\mathcal{D}_j|}{|\mathcal{D}_j|+ Z'}\right)^2.
		  \end{align}
	  Next, we consider the inequality in \eqref{eq: initial recursion with r 0}. {By replacing, $r_j = \frac{k}{ Z'}$}, we get rid of the constant exponential term and obtain
	  \begin{align}
		      {\DSD}(P^k;{\mathcal{D}}_{k})^2& \leq \sum_{i=1}^{k} \left( \frac{ Z'(B^2+r_iC_0 S^2A)}{\left(|\mathcal{D}_{i-1}|+ Z'\right)^2}\right)\left(\prod_{j=i}^{k-1} \frac{|\mathcal{D}_j|}{|\mathcal{D}_j|+ Z'}\right)^2
		       \nonumber
		       \\
		       =&\sum_{i=1}^{k} \left( \frac{ Z'(B^2+r_iC_0 S^2A)}{\left(|\mathcal{D}_{k-1}|+ Z'\right)^2}\right)\left(\prod_{j=i}^{k-1} \frac{|\mathcal{D}_j|}{|\mathcal{D}_{j-1}|+ Z'}\right)^2\label{here1111},
		  \end{align}
        where Equation \eqref{here1111} corresponds to the sampling error and represents the bias incurred at each step of the SPMCMC point selection scheme.
	The equality in the second line holds by rearranging the denominators in the multiplication and pulling $\left(|\mathcal{D}_{k-1}|+ Z'\right)^2$ inside the first term. Next, from the fact that ${|\mathcal{D}_j|} = {|\mathcal{D}_{j-1}|+ Z'}$ which implies that the product will be  $1$, we can upper bound the right hand side of \eqref{here1111} as follows
	  \begin{align}\label{eq: bound computation0}
		       {\DSD}(P^k;{\mathcal{D}}_{k})^2&\leq\sum_{i=1}^{k} \left( \frac{ Z'(B^2+r_iC_0 S^2A)}{\left(|\mathcal{D}_{k-1}|+ Z'\right)^2}\right).
		  \end{align}
	  %
From the dictionary update,  we note that $|\mathcal{D}_{k-1}|+ Z' = |{\mathcal{D}_k}| = \mathcal{O}(k)$, which implies that $1/\left({|{\mathcal{D}_{k-1}}|+ Z'}\right)^2 = 1/k^2$, which we utilize in the right hand side of  \eqref{eq: bound computation0} to write
	  \begin{align}\label{eq: bound computation}
		      {\DSD}(P^k;{\mathcal{D}}_{k})^2 &\leq\sum_{i=1}^{k} \left( \frac{ Z'(B^2+r_iC_0 S^2A)}{ k^2}\right).
		  \end{align}
	We note that the above bound holds for any given $M$. And hence we can conclude that after taking an expectation over posterior $M\sim \phi(\cdot |\mathcal{D}_k)$, it holds that 
	  \begin{align}\label{eq: bound computation2}
		      \mathbb E_k \left[{\DSD}(P^k;{\mathcal{D}}_{k})^2\right] &\leq\sum_{i=1}^{k} \left(\frac{ Z'(B^2+r_iC_0 S)}{k^2}\right)= \mathcal{O}\left(\frac{S^2A}{k}\right),
		  \end{align}
   where we absorb the constants into $\mathcal{O}(\cdot)$ notation.  Hence proved.  We can improve the above bound to $\mathcal{O}\left(\frac{SA}{k}\right)$ for kernels that satisfy some factorization properties. For instance, the above result holds for the Kronecker delta kernel where $\kappa((x,y), (x', y')) = \delta((x,y), (x', y'))$ which implies 
$k((x,y), (x', y')) = \delta(x,x') \delta(y,y')$, and leads to improved upper-bound in DSD of order $\mathcal{O} (\frac{SA}{k})$. This result would also hold for the bounded function $\|G(x,\cdot)\|\leq C$ (cf. \eqref{our_kernel}) for some constant $C$.  This would help us to  upper-bound the term $\|f_k\|^2_{\mathcal{K}_0} \leq C_0 S A$ and subsequently follow the same steps (\eqref{here_14}-\eqref{eq: bound computation2}) to get the final bound of $\mathcal{O}\left(\frac{SA}{k}\right)$.
	  \end{proof}

    \section{Proof of Lemma \ref{lemma_1}}\label{proof_lemma_1}
   
  \begin{proof}
	  \textbf{Proof of statement (1):} 
	  We start by analyzing the regret decomposition of the value function as follows
	  \begin{align}\label{eqn:regret_decomposition}
		            &\mathbb E_k\left[V_{1,\pi^*}^{M^*}(s_1^k)-V_{1,\pi^k}^{M^*}(s_1^k)\right]= \underbrace{\mathbb E_k\left[V_{1,\pi^*}^{M^*}(s_1^k)-V_{1,\pi^k}^{{M}^k}(s_1^k)\right]}_{I_1} +  \underbrace{\mathbb E_k\left[V_{1,\pi^k}^{{M}^k}(s_1^k)-V_{1,\pi^k}^{M^*}(s_1^k)\right]}_{I_2}\,.
		   \end{align}
   From the probability matching principle of PSRL \citep{osband2013, osband2014model, osband2017posterior, osband17a}  we have  $\mathbb E_k\left[V_{1,\pi^*}^{M^*}(s_1^k)-V_{1,\pi^k}^{{M}^k}(s_1^k)\right] = 0$ conditioned on the history $D_k$. At the start of each episode $M^*, M^k$ are identically distributed as detailed in \citep{osband2013}, hence $I_1 = 0$\\
  \textbf{Upper Bound on $I_2$}\\
  Now, we derive the upper bound on $I_2$
  \begin{align}\label{interemediate_Delta}
	      \Delta_h^{k}(s,a) :=\mathbb E_{s'\sim P^{M^k}(\cdot|s,a)}[V_{h+1,\pi^k}^{M^k}(s')]-\mathbb E_{s'\sim P^{ M^*}(\cdot|s,a)}[V_{h+1,\pi^k}^{ M^k}(s')]\,.
	  \end{align}
  Now, using the definition in \eqref{interemediate_Delta}, we can write $I_2$ as 
  \begin{align}
	      I_2
	        &=\mathbb E_{k}\left[\sum_{h=1}^H\mathbb E_{\pi^k}^{M^*}\left[\Delta_h^{k}(s_h^k,a_h^k)\right]\right]
	        \\
	        &=
	        \sum_{h=1}^H\mathbb E_{k}\left[\sum_{(s,a)}d_{h,\pi^k}^{M^*}(s,a)\Delta_h^{k}(s,a)\right]\,,\label{first}
	        \\
	          &=\sum_{h=1}^H\mathbb E_{k}\left[ \sum_{(s,a)}\frac{d_{h,\pi^k}^{ M^*}(s,a)}{(\mathbb E_{k}[d_{h,\pi^k}^{M^*}(s,a)])^{1/2}}(\mathbb E_{k}[d_{h,\pi^k}^{M^*}(s,a)])^{1/2}\Delta_h^{k}(s,a)\right]\,.\label{second}
	  \end{align}
   The equality in \eqref{first} holds by introducing the state action occupancy measure, and in \eqref{second},  we divide and multiply with $d_{h,\pi^k}^{M^*}(s,a)>0$. From the Cauchy–Schwartz inequality and using the fact that $d_{h,\pi^k}^{M^*}(s,a)\leq 1$, we can write 
  \begin{align}\label{eqn:bound_state_occupancy}
	      I_1
	       &\leq \left(\sum_{h=1}^H\mathbb E_{k}\left[\sum_{(s,a)}\frac{(d_{h,\pi^k}^{M^*}(s,a))^2}{\mathbb E_{k}[d_{h,\pi^k}^{M^*}(s,a)]}\right]\right)^{1/2}\left(\sum_{h=1}^H\mathbb E_{k}\left[\sum_{(s,a)}\mathbb E_{k}[d_{h,\pi^k}^{M^*}(s,a)](\Delta_h^{k}(s,a))^2\right]\right)^{1/2}
	       \nonumber
	       \\
	       &\leq \left(\sum_{h=1}^H\mathbb E_{k}\left[\sum_{(s,a)}\frac{d_{h,\pi^k}^{M^*}(s,a)}{\mathbb E_{k}[d_{h,\pi^k}^{M^*}(s,a)]}\right]\right)^{1/2}\left(\sum_{h=1}^H\mathbb E_{k}\left[\sum_{(s,a)}\mathbb E_{k}[d_{h,\pi^k}^{M^*}(s,a)](\Delta_h^{k}(s,a))^2\right]\right)^{1/2}\,.
	  \end{align}
  First, from the first part in the right-hand side of \eqref{eqn:bound_state_occupancy}, we note that
  {
  \begin{align}\label{eqn:I11}
	      \sum_{h=1}^H\mathbb E_{k}\left[\sum_{(s,a)}\frac{d_{h,\pi^k}^{M^*}(s,a)}{\mathbb E_{k}[d_{h,\pi^k}^{M^*}(s,a)]}\right]=\sum_{h=1}^H\left[\sum_{(s,a)}\frac{\mathbb E_{k}[d_{h,\pi^*}^{M^*}(s,a)]}{\mathbb E_{k}[d_{h,\pi^k}^{M^*}(s,a)]}\right] \leq HSA \,.
	  \end{align}
   }

  Then, from the second part on the right-hand side of \eqref{eqn:bound_state_occupancy},  we note that given $\cD_k$, we have $d_{h,\pi^k}^{M^*}(s,a)$ and $\Delta_h^{k}(s,a)$ independent. This implies
      {
      \begin{align}
	          \mathbb E_{k}\left[\sum_{(s,a)}d_{h,\pi^k}^{M^*}(s,a)\right]\mathbb E_{k}\left[(\Delta_h^{k}(s,a))^2\right]&=\mathbb E_{k}\left[\sum_{(s,a)}d_{h,\pi^k}^{M^*}(s,a)(\Delta_h^{k}(s,a))^2\right]\nonumber\\
	        &=\mathbb E_{k}\left[\mathbb E_{\pi^k}^{M^*}\left[(\Delta_h^{k}(s_h^k,a_h^k))^2\right]\right],\label{proof_2}
	      \end{align}
       }
      From the definition in \eqref{interemediate_Delta}, we can write
      {
  \begin{align}\label{eqn:I12}
	      \mathbb E_{k}&\left[\mathbb E_{\pi^k}^{M^*}\left[(\Delta_h^{k}(s_h^k,a_h^k))^2\right]\right]
	  \nonumber
	       \\
	          &=H^2\mathbb E_{k}\left[\mathbb E_{\pi^k}^{ M^*}\left[\left(\mathbb E_{s'\sim P^{M^k}(\cdot|s_h^k,a_h^k)}\left[V_{h+1,\pi^k}^{M^k}(s')/H\right]-\mathbb E_{s'\sim P^{ M^*}(\cdot|s_h^k,a_h^k)}\left[V_{h+1,\pi^*}^{ M^k}(s')/H\right]\right)^2\right]\right]\nonumber
	          \\
	          &= H^2\mathbb E_{k}\left[\mathbb E_{\pi^k}^{ M^*}\|P^{M^k}(\cdot|s_h^k,a_h^k) - P^{M^*}(\cdot|s_h^k,a_h^k)\|^2 \right].
	  \end{align}
   }
  Next, combining \eqref{eqn:I11}, \eqref{proof_2}, and \eqref{eqn:I12}, can upper bound $I_2$ in \eqref{eqn:bound_state_occupancy} as 
  \begin{align}\label{intermed_bound_i1}
	    I_2 &\leq  \sqrt{H^3 SA \sum_{h=1}^H \mathbb E_{k}\left[\mathbb E_{\pi^k}^{ M^*}\|P^{M^k}(\cdot|s_h^k,a_h^k) - P^{M^*}(\cdot|s_h^k,a_h^k)\|^2 \right])}.
	  \end{align}
  {Next, we take square on both sides and introduce the Kernelized Stein discrepancy between probability measures $P^{M}$ and $P^{M^*}$ by upper-bounding the total variation norm in equation \eqref{intermed_bound_i1} \footnote{In (\citep{liu2016kernelized}), the Kernelized Stein discrepancy has been defined as $KSD^2$ and in \citep{gorham2015measuring} as $KSD$, hence it can be used interchangeably. For ease of the analysis, we proceed by considering $KSD^2$ as the Stein discrepancy.} \citep{gorham2015measuring} which is a clear departure from the traditional Bayesian regret analysis as in \citep{osband2013, osband17a, osband2014model, osband2019deep}.  to obtain}
  \begin{align}
	    I_2^2
	    &\leq  H^3 SA \sum_{h=1}^H \mathbb{E}_{k}\left[\mathbb E_{\pi^k}^{ M^*}[DSD^2(P^{M^k}(\cdot|s_h^k,a_h^k),P^{M^*}(\cdot|s_h^k,a_h^k))] \right] \nonumber\\
        &\leq  H^3 SA \sum_{h=1}^H \mathbb E_{k}\left[\mathbb E_{\pi^k}^{ M^*}[DSD^2(P^{k}(\cdot|s_h^k,a_h^k))] \right].
	  \end{align}
  {The second line in the equation is due to the definition of KSD which requires only the unnormalized density/pmf of one and samples from the other. Also, we add the second line for notational consistency} Next, from the definition in \eqref{informayion_gain_KSD}, we can finally write 
  \begin{align}
	    I_2^2 
	    &\leq H^3 SA \cdot \mathbb K_k^{\pi}\left(M^*; \cH_{k, H}\right).
	  \end{align}
  From \eqref{eqn:regret_decomposition}, we can write
  {
  \begin{align}\label{final_step_lemma_1}
	      \left(\mathbb E_k\left[V_{1,\pi^k}^{M^k}(s_1^k)-V_{1,\pi^k}^{M^*}(s_1^k)\right]\right)^2\leq H^3 SA \cdot \mathbb K_k^{\pi}\left( M^*; \cH_{k, H}\right).
	  \end{align}
   }
  Since $I_1 =0$, as detailed above, adding that to equation \eqref{final_step_lemma_1}, we get
  \begin{align}\label{final_step_p2}
	      \left(\mathbb E_k\left[V_{1,\pi^*}^{M^*}(s_1^k)-V_{1,\pi^k}^{M^*}(s_1^k)\right]\right)^2\leq H^3 SA \cdot \mathbb K_k^{\pi}\left( M^*; \cH_{k, H}\right).
	  \end{align}
  {Hence, from equation \eqref{final_step_p2} and using the definition of Stein information ratio in \eqref{def:information_ratio} and inequality in \eqref{final_step_lemma_1},  it is clear that we can upper bound $\mathbb E[{\Gamma}^{DSD}_k]$ as $\mathbb E[{\Gamma}^{DSD}_k]\leq SAH^3$. Now from the definition of $\Gamma^* \leq \Gamma_k^{{DSD}}(\pi)$ we have $\Gamma^* \leq SAH^3$ that which completes the proof.}

  
  \textbf{Proof of statement (2):} 
  {In this section, we derive an upper-bound on the total Stein information for the $K$ episodes given by $\mathbb\sum_{k=1}^K \mathbb E [\mathbb K_k^{\pi}\left( M; \cH_{k, H}\right)]$ in order to compute the second term in the equation \eqref{Bayesian-regret-stein_Ration}. We start by  deriving an upper bound for $\mathbb E [\mathbb K_k^{\pi}\left( M; \cH_{k, H}\right)]$ with an SPMCMC style local optimization method proposed originally in \citep{stein_point_Markov} and later used in sequential decision-making scenarios by \citep{chakraborty2022posterior, cole_koppel}.}\\
  We start with the definition in \eqref{informayion_gain_KSD} to write
  {
  \begin{align}\label{ksd_upperbound}
	   \mathbb E_k [\mathbb K_k^{\pi}\left(M^*; \cH_{k, H}\right)]& = \sum_{h=1}^H \mathbb E\left[DSD^2(P^{k}(\cdot|s_h^k,a_h^k)\right].
	  \end{align}
   }
  From the upper bound in Lemma \ref{lemma_1}, we can write

  {
  \begin{align}\label{ksd_upperbound2}
	   \mathbb E [\mathbb K_k^{\pi}\left(M^*; \cH_{k, H}\right)]
	    & \leq \sum_{h=1}^H \mathcal{O}\left(\frac{S^2 A}{k}\right) = \mathcal{O}\left(\frac{HS^2 A}{k}\right).
	  \end{align}
   }
  Finally, we take summation over $k$ and obtain the upper bound as 
  {
  \begin{align}
	    \mathbb\sum_{k=1}^K \mathbb E [\mathbb K_k^{\pi}\left(M^*; \cH_{k, H}\right)] \leq HS^2 A \int_{1}^{K}\frac{1}{x} dx ={HS^2 A(\log K)}. \label{integral_bound_one} %
	  \end{align}
   }
  Hence Proved. With the Corollary \ref{improvement_corollary}, the following upper-bound can be improved by an order of $S$ under the appropriate assumptions to $HS A(\log K)$.
  \end{proof}

  \section{Proof of Theorem \ref{theorem}}\label{proof_of_main_theorem}
  	  \begin{proof}
		  Consider the expression in \eqref{final_bayes_regret} and we note that the Bayesian regret depends on the Stein information ratio $\mathbb E[{\Gamma}^*]$ and the total Stein information gain $ \sum_{k=1}^K \mathbb E [\mathbb K_k^{\pi}\left( M^*; \cH_{k, H}\right)]$.
		  From Lemma \ref{lemma_1}, we can upper bound the right hand side of \eqref{final_bayes_regret} as
		  \begin{align}\label{final_bayes_regret3}
			      \BR_K\leq & \sqrt{SAH^3\cdot K \cdot HS^2 A(\log K)}
         \\
         = & H^2\sqrt{ S^3 A^2 K  \log K}=\tilde{\mathcal{O}}(H^2\sqrt{ S^3A^2 K }),
			  \end{align}
		 where $\tilde{\mathcal{O}}$ absorbs the log factors and we obtain the final result. 
		  \end{proof}   

  \section{Proof of Theorem  \ref{theorem_regularized_regret}}\label{proof_regularized_regret}
  	  \begin{proof}
		 Following the inequality $2ab\leq a^2+b^2$, for any policy $\pi$, we can write 

	\begin{align} \label{bound_AM_GM}
        &\frac{\mathcal{R}_k}{\sqrt{\lambda \mathbb K_k^{\pi}\left({M^*}; \cH_{k, H}\right)}}\sqrt{\lambda \mathbb K_k^{\pi}\left({M^*}; \cH_{k, H}\right)} \leq 
			 \frac{\mathcal{R}_k^2}{2 \lambda \mathbb K_k^{\pi}\left({M^*}; \cH_{k, H}\right)} + \frac{\lambda}{2} \mathbb K_k^{\pi}\left({M^*}; \cH_{k, H}\right),
			  \end{align}
		  where $\mathcal{R}_k:=\mathbb E \left[V_{1,\pi^*}^{M^*}(s_1^k)-V_{1,\pi^k}^{M^*}(s_1^k)\right]$,.
		  Now, recollecting the definition of Bayesian regret from \eqref{bayesian_Regret}  and after adding and subtracting the regularization term, we can write 
		  \begin{align}\label{eq:reg_regret}
	       \BR_K 
	       &= \mathbb E\Bigg[\sum_{k=1}^K\mathcal{R}_k-\lambda\sum_{k=1}^K\mathbb K_k^{\pi}\left({M^*}; \cH_{k, H}\right))+ \lambda\sum_{k=1}^K\mathbb K_k^{\pi}\left({M^*}; \cH_{k, H}\right))\Bigg]. 
	      \end{align}
	      Utilizing the upper bound in  \eqref{bound_AM_GM}, we can write
	      \begin{align}\label{final_Regret_Regu}
        \BR_K  \leq & \frac{1}{2\lambda}\sum_{k=1}^K \mathbb E\left[\frac{\left(\mathbb E_k\left[V_{1,\pi^*}^{M^*}(s_1^k)-V_{1,\pi}^{M^*}(s_1^k)\right]\right)^2}{\mathbb K_k^{\pi}\left({M^*}; \cH_{k, H}\right))}\right]+\lambda \sum_{k=1}^K \mathbb E [\mathbb K_k^{\pi}\left( {M^*}; \cH_{k, H}\right)]\nonumber\\
        =& \frac{K \mathbb E[\Gamma^*] }{2\lambda} +\lambda \sum_{k=1}^K \mathbb E [\mathbb K_k^{\pi}\left(M^*; \cH_{k, H}\right)].
		\end{align}
 %
		 Now, select $\lambda$ in \eqref{final_Regret_Regu} as
		 %
		     $\lambda = \sqrt{K\mathbb E[\Gamma^*]/\mathbb \sum_{k=1}^K \mathbb E [\mathbb K_k^{\pi}\left(M^*; \cH_{k, H}\right)}$
        to obtain the final expression. 
     \end{proof} 

\section{Detailed Information of Experimental Setup}\label{additonal_experiments}
    \subsection{Description of the Environments} \label{environment_Details}
    \textbf{DeepSea Environment:} The DeepSea exploration environment (a slightly modified version of \citet{osband17a} as used in \citet{markou_mm}) is an extremely challenging environment (see Fig. \ref{fig:architecure}) to test the agent's capability of directed and sustained exploration. As shown in Fig. \ref{fig:architecure}, there are total $N$ states, the agent starts from the left-most state and can swim left or right from each of the $N$ states in the environment. The agent gets a reward of $r = 0$ (red transitions) for the left action. On the other hand, the right action from $s = 1, \cdots, (N - 1)$ succeeds with probability $(1 - 1/N)$, moving the agent to the right and otherwise fails and moving the agent to the left (blue arrows), giving $r\sim \mathcal{N}(-\delta, \delta^2)$ regardless of whether it succeeds. A successful swim-right from $s = N$ moves the agent back to $s = 1$ and gives $r = 1$. Hence, as we increase the number of states $N$, it will increase the amount of sparsity in the environment, making it extremely hard for the agent to explore (we provided this evidence in Fig. \ref{figure_sparse_rewards}). 
    	%
	  \begin{figure}[H]
	  \centering
        \includegraphics[width=0.55\textwidth]{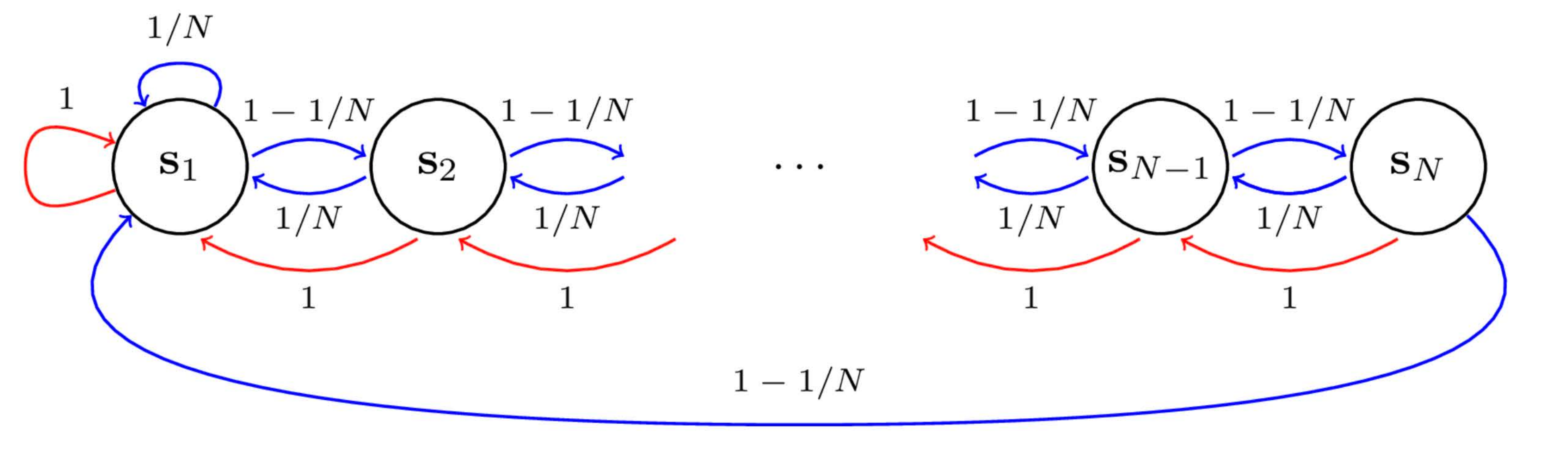}
        \caption{DeepSea Exploration Environment \citep{osband17a, markou_mm}. This environment tests the agent's capability of sustained and directed exploration. This figure is same as Fig.~\ref{fig:new_architecure} and we repeat it here for quick reference. }
        \label{fig:architecure}
     \end{figure}

     \textbf{WideNarrow MDP Environment:} This is another challenging environment (see Fig. \ref{fig:wide_narrow}) presented in \citet{markou_mm}, which has $2N + 1$ states with deterministic transitions. In WideNarrow MDP environment,  odd-numbered states except $s = (2N + 1)$ have $W$ actions, out of which one gives $r \sim \mathcal{N}(\mu_l, \sigma_l^2)$ whereas all other actions result in $r\sim\mathcal{N}(\mu_h, \sigma_h^2)$, with $\mu_l < \mu_h$. Even-numbered states have a single action which results in $r \sim\mathcal{N}(\mu_h, \sigma_h^2)$. In the experiments, we use $mu_h = 0.5, \mu_l = 0, \sigma_l = \sigma_h = 1$. The WideNarrow MDP helps understand and compare the performance of {\algo} under factored posterior approximations.
     \begin{figure}[h]
	  \centering
        \includegraphics[width=0.45\textwidth]{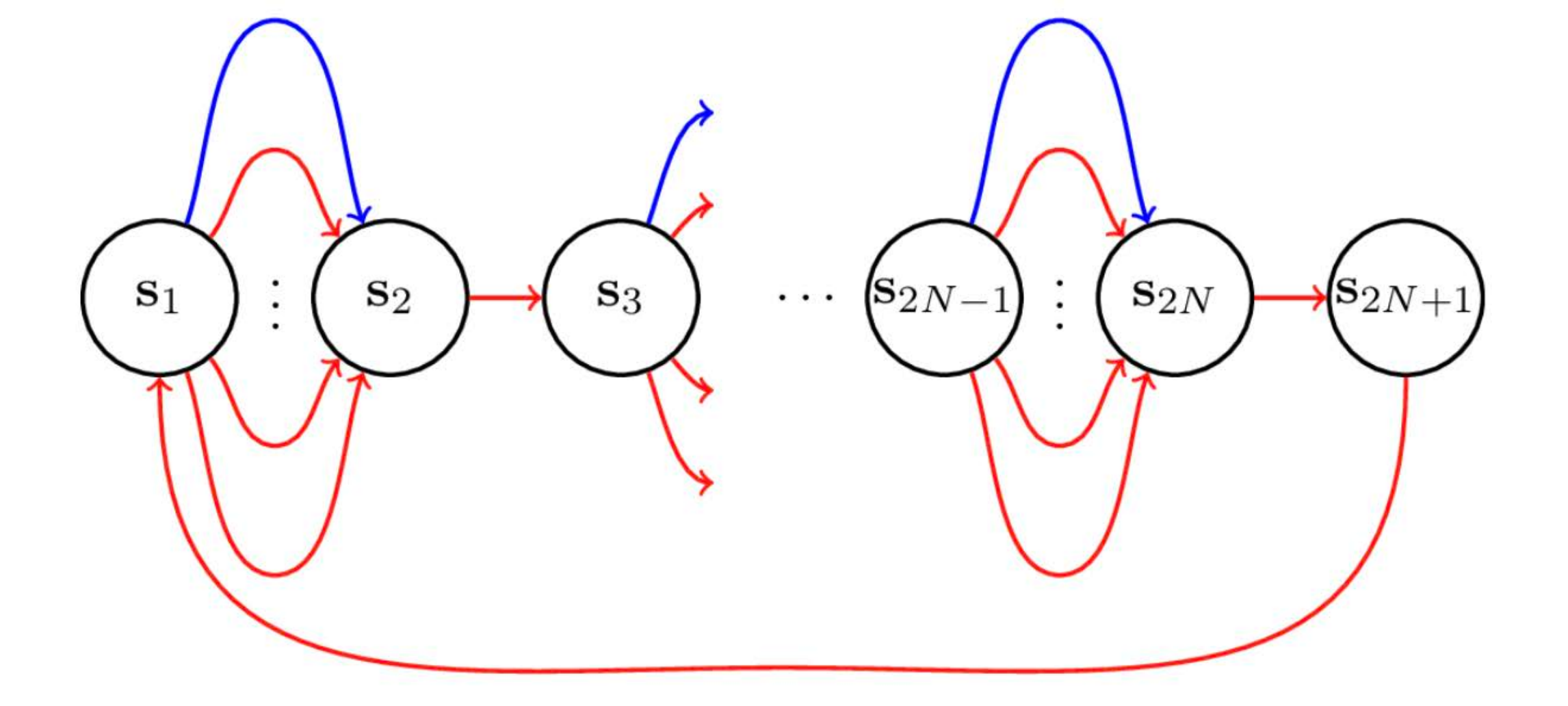}
        \caption{WideNarrow MDP environment \citep{markou_mm}. The purpose of this environment is to test the agent's performance under factored posterior approximations.}
        \label{fig:wide_narrow}
     \end{figure}

    \textbf{PriorMDP Environment}: The previous environments, DeepSea Exploration and WideNarrow MDP, have a special structure within them that tests different aspects of the agent/algorithm. In contrast, the PriorMDP environment of \citet{markou_mm} provides a more general environmental setup to test the algorithms where the dynamics are sampled from a Dirichlet prior with concentration $k = 1$ and reward from Normal prior with mean and precision drew from a Normal-Gamma prior with parameters $\langle0,1,1,4\rangle$.
    \subsection{Baselines and Evaluations} \label{Q_value_compairons}
    We compare our proposed algorithm  {\algo} with various Bayesian/ non-Bayesian and distributional RL baselines. The baselines include 
    \begin{itemize}
        \item \textbf{Q-learning}: Vanilla Q-learning with $\epsilon-$greedy action selection \citep{Watkins1992},
        \item \textbf{BQL}: Bayesian Q-learning \citep{dearden_bql},
        \item \textbf{UBE}: Uncertainty Bellman equation \citep{donoghue_ube}, 
        \item \textbf{MM}: Moment Matching across Bellman equation \citep{markou_mm}, 
        \item \textbf{PSRL}: Posterior Sampling Reinforcement Learning \citep{osband2013}.
    \end{itemize}

    \textbf{1. Bayesian Q-Learning} : BQL introduced in \citet{dearden_bql} models the distribution over state-action returns $Z^*$, which is assumed to follow Gaussian (ergodic MDP) and updates the posterior belief of $P(\theta_{Z^*}|D)$ using Bayesian update rule.  BQL considers Normal-Gamma prior on the parameters (mean and precision) of the Gaussian. However, there is a factored posterior assumption that restricts its generalisability and hinders the performance, as shown in Fig. \ref{fig:sids_s5}
    
    \textbf{2. Uncertainty Bellman Equation}: UBE proposed in \citet{donoghue_ube} is a model-based reinforcement learning approach designed primarily for modeling the epistemic uncertainty in $\mu_z = \E_z[Z|\theta_z]$ but with a strong assumption of MDP being a directed acyclic graph with bounded mean rewards. In other words, each state-action can be visited at most once per episode, which is restrictive and requires sparse design (repeating state-action multiple times) to make it work even for toy problems. Under these assumptions and a suitable Bellman operator, it holds  that UBE has a unique solution which upper-bounds the epistemic uncertainty $\text{Var}_{\theta_T,\theta_R} [\mu_z]$ where $\theta_T, \theta_R$ are the parameters of the transition and rewards model, respectively. However, the strong assumptions restrict the generalisability of the UBE-based methods to various practical scenarios where the inherent structure of the MDP is not acyclic.

    \textbf{3. Moment Matching across Bellman Equation}: An interesting approach proposed recently by \citet{markou_mm} also uses the Bellman equation to estimate the epistemic uncertainty by comparing the moments (first and second moments) across the Bellman equation. While the first-order moments give the standard $V(s)$ \& $ Q(s,a)$, the second-order moments can be decomposed into aleatoric and epistemic terms without the need to compute upper bounds as in UBE-based methods. Similar to prior methods, the policy is optimized w.r.t $P(\theta_T|D), P(\theta_R|D)$ and for the epistemic uncertainty $\mu_z$. However, as with existing methods, MM also approximates with a factored posterior leading to performance loss. 
    
    \textbf{4. Posterior Sampling Reinforcement Learning:} Posterior Sampling reinforcement learning (PSRL) introduced by \citet{osband2013, osband2014model} is primarily built upon Thompson sampling or probability matching principle. PSRL provides an efficient and tractable solution to model-based RL with provable guarantees. The algorithm works by sampling a transition and rewards model from the posterior distribution at any $k^{th}$ episode $\theta_T^k \sim P(\theta_T|D_k), \theta_R^k \sim P(\theta_R|D_k)$ and the optimal policy $\pi^k$ is obtained by solving the Bellman equation under the sampled transition, rewards model. The agent then follows the policy $\pi^k$ to interact with the environment and gather data and update the dictionary $D_k$. The primary advantage of PSRL lies in its computational tractability, as the policy needs to be optimized under a single sampled transition and rewards model. For PSRL, state-of-the-art Bayesian regret bounds exist under minor assumptions \citep{osband2014model, osband2017posterior}. However, even the performance of PSRL degrades for complex environments such as under sparse reward scenarios, as empirically proved in Fig. \ref{fig:psrl_sparse}.

    We evaluate all the algorithms by comparing their performance in terms of cumulative regret. Further, for the Bayesian methods, we also evaluate the algorithms in terms of the posterior representation and concentration on the true $Q^*$ values.

    \subsection{Implementation Details of {\algo}}  
    Here we present the implementation details of {\algo}, as outlined in Algorithm \ref{alg:MPC-PSRL}. Our approach begins by sampling transition and reward models from the posterior distribution using Categorical-Dirichlet and Normal-Gamma distributions for the transition and rewards model,  respectively, as in \citet{markou_mm}. To ensure a fair comparison, we considered the same setup for all categorical and continuous distributions for the other baselines. In the second phase, our algorithm represents a distinct deviation from prior PSRL and IDS-based methods by quantifying the distributional distance between the true MDP and the current estimated MDP through the use of Stein discrepancy. Specifically, we utilize the regularized Stein information sampling as outlined in Theorem \ref{theorem_regularized_regret} for empirical analysis. Next, we compute the Stein-discrepency for each $(s,a)$ pair using the formulae for DSD as defined in \eqref{eq:kssd_pop_ustat}. 
    \subsection{Hyperparameters}
    
   For all Dirichlet priors in the algorithms we use hyperparameters $\eta_{(s,a)} =1$ and for Normal-Gamma priors we use $(\mu, \Lambda, \alpha, \beta)_{(s,a)} = (0,4,3,3)$ as in \citet{markou_mm}. For both {\algo} and Var-IDS we use the same regularization constant (IA) $\lambda = 0.5$. For all the environments, we run the algorithms for $T = 5000$ timesteps with a buffer length (max) of size $N$, where $N$ denotes the number of states and run the policy iteration for $2N$ iterations. We have also run PSRL, Var-IDS and {\algo} for $T = 10000$ in Figure \ref{fig:conc_pp} to observe the effect of prior with convergence. For the baseline implementation of the algorithms including Q-Learning, UBE, BQL, MM, PSRL we leverage \footnote{\url{https://github.com/stratisMarkou/sample-efficient-bayesian-rl}}. We utilize \footnote{\url{https://github.com/jiaseny/kdsd}} and \footnote{\url{https://github.com/colehawkins/KSD-Thinning}} for the DSD computation. We thank the authors for the open-source repositories.

    \section{Additional Experimental Results \& Discussions (Intuitive Insights)} \label{experimental _Study_appendix}
    \subsection{Evolution of Posterior Representations for DeepSea Environment}\label{convergence_Q}
    To gain more insights about the improvements and directed exploration behavior achieved by {\algo}, we perform a detailed ablation study to analyze the evolution of  posterior representation learned over  iterations for all the algorithms. As a metric to plot, we plot the mean and variance of $Q$ values calculated from the learned posterior and compare them against the true value denoted by $Q^*$ (see Fig. \ref{fig:sids_s0000} for {\algo}). We remark that as the posterior concentrates with the progress in training, estimated $Q$ values would also concentrates to the true $Q^*$ (shown by dotted red line in Fig. \ref{fig:sids_s0000}). Since $N$ denotes the number of states for DeepSea environment, and we obtain an estimated $Q$ value for each state action pair, we choose to plot the evolution of the last $4$ states -action pairs (left action in top row and right action in bottom row in Fig. \ref{fig:sids_s0000}). 
            \begin{figure}[h]
	  \centering
          \subfigure[]{ \includegraphics[width=0.46\textwidth]{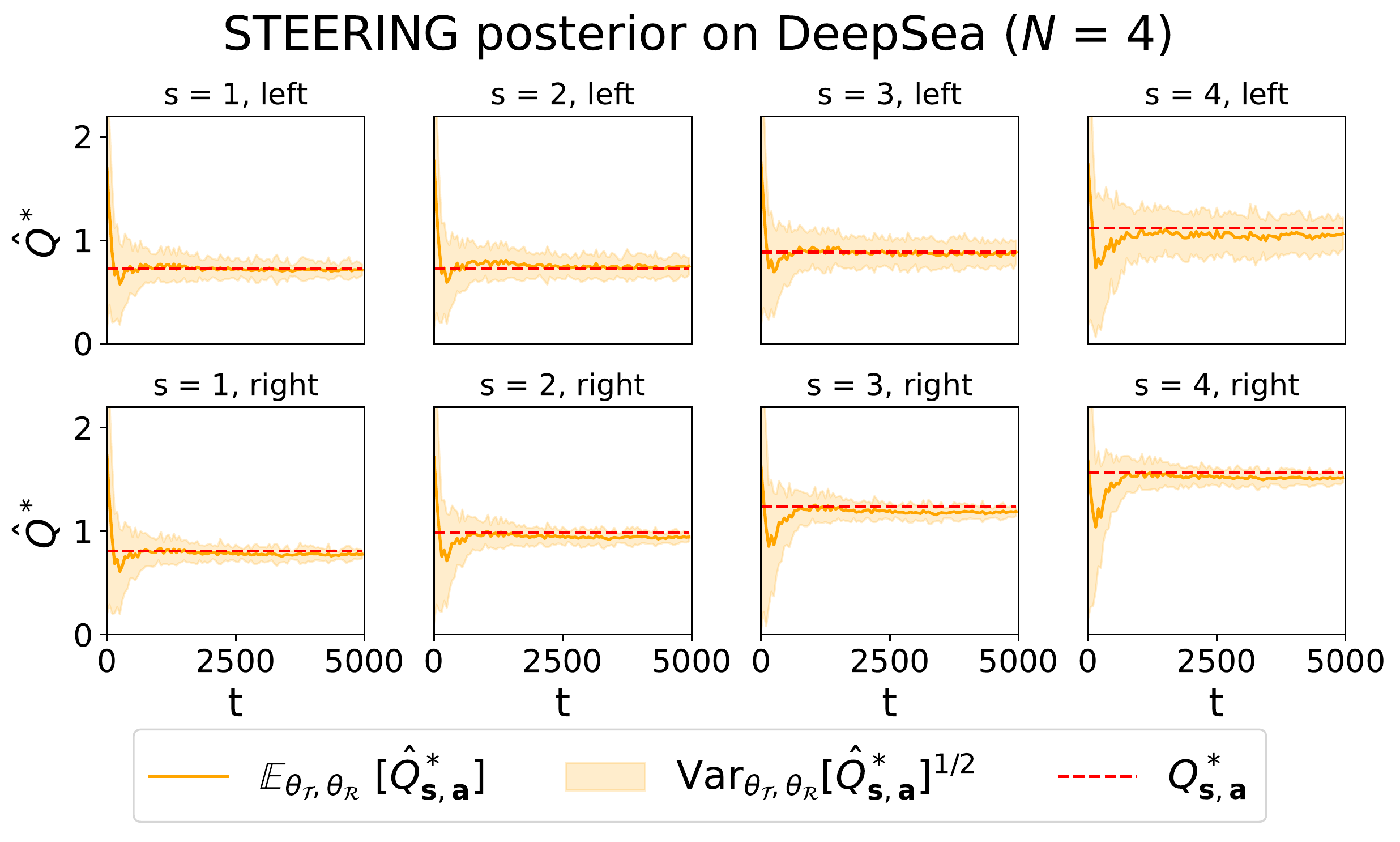}        }       \hspace{0mm}         
          \subfigure[]{\includegraphics[width=0.46\textwidth]{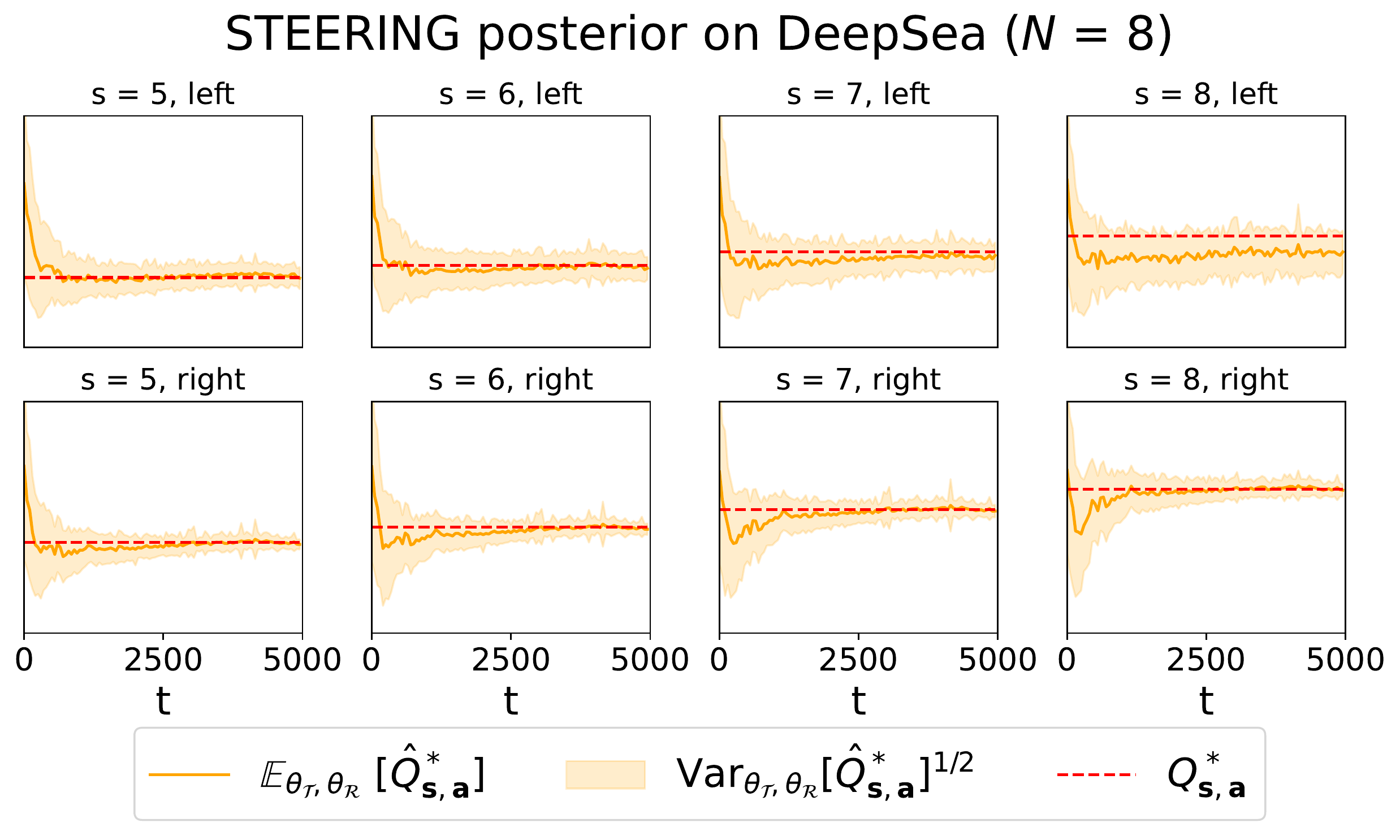}        }
        \caption{This plot analyzes the concentration of the predicted $\hat{Q}$ to the true $Q^*$ (ground-truth) with iterations for {\algo}. \textbf{(a)} This figure shows the performance for DeepSea environment with $N=4$. \textbf{(b)} This figure shows the performance for DeepSea environment with $N=8$.         
       }
        \label{fig:sids_s0000}
     \end{figure}

Fig.~\ref{fig:sids_s0000} is of critical importance, as it gives a clear understanding of whether the agent is over-exploring or under-exploring actions based on its sub-optimality. To make the advantage clear as compared to existing approaches, we plot similar figures for all the existing algorithms in Fig.~\ref{fig:sids_s3} to Fig.~\ref{fig:sids_s1}. \textbf{Interestingly, in all the comparison plots from Fig.~\ref{fig:sids_s3} to Fig.~\ref{fig:sids_s1}, {\algo} exhibits a superior posterior representation which results in directed exploration, even with increased sparsity.} The most interesting aspect of {\algo} over baselines is that it does not over-explore actions once it is confident that those are sub-optimal which is an important feature helps in achieving directed exploration.
        \begin{figure}[h]
	  \centering
       \subfigure[]{  \includegraphics[width=0.46\textwidth]{figures_new/steering_4.pdf} }         \subfigure[]{\includegraphics[width=0.46\textwidth]{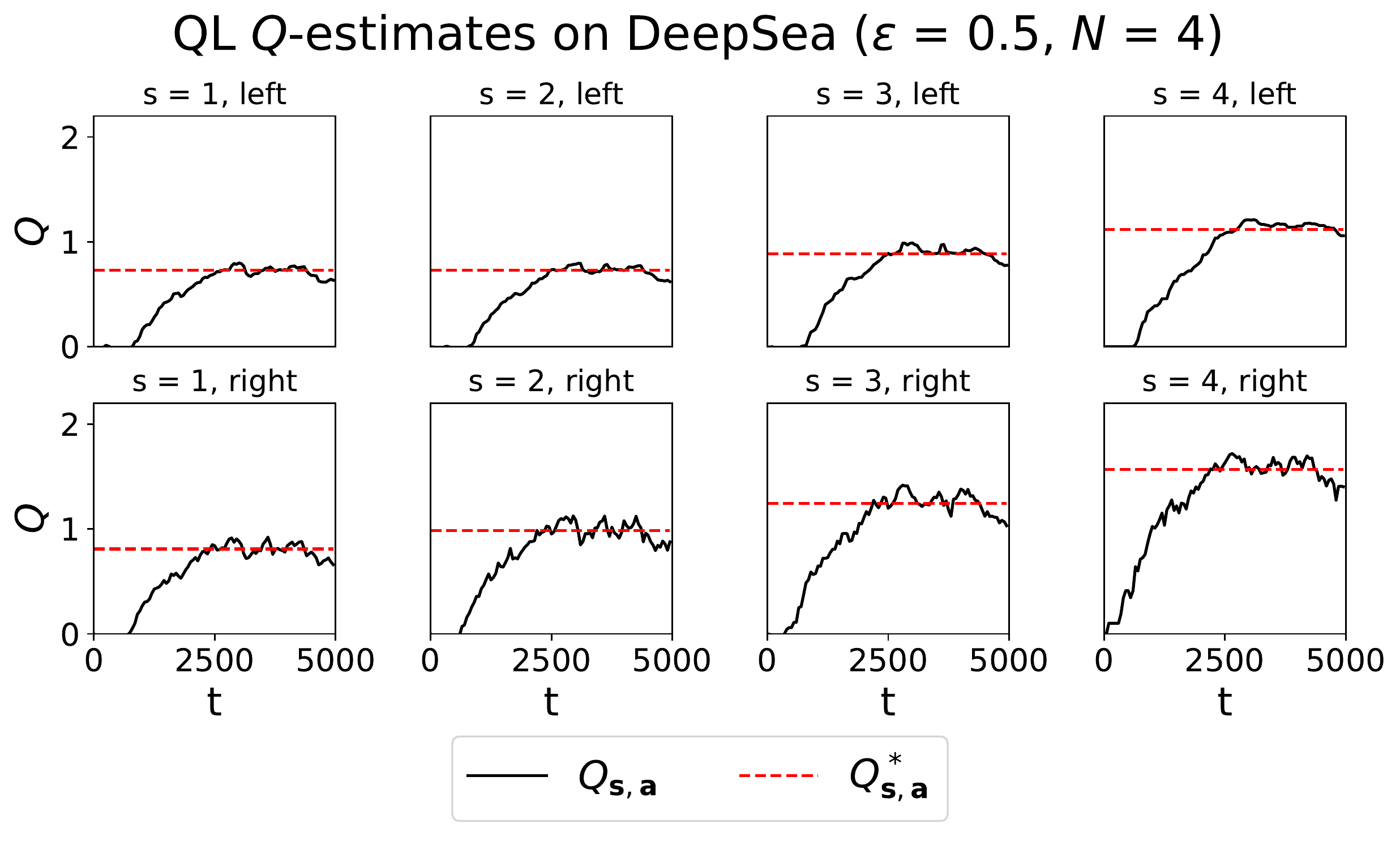}}
        \caption{Comparison of {\algo} and $Q$-learning on  DeepSea Exploration with sparsity $N = 4$. This plot shows the concentration of the predicted $\hat{Q}$ to the true $Q^*$ (ground-truth) versus iterations. It is evident that {\algo} converges to true $Q^*$ much faster than Q-learning.
        \textbf{Remark:} As right actions are optimal for DeepSea environment, {\algo} stops exploring left actions beyond a point, leading to a comparatively higher variance in $\hat{Q}$ for left actions, thus providing directed exploration.
       }
        \label{fig:sids_s3}
     \end{figure}
     \begin{figure}[t]
	  \centering
        \subfigure[]{\includegraphics[width=0.47\textwidth]{new_conv/Steering-0_0-4_0-3_0-3_0-posterior-deepsea-8_2.pdf}        }
        \subfigure[]{\includegraphics[width=0.47\textwidth]{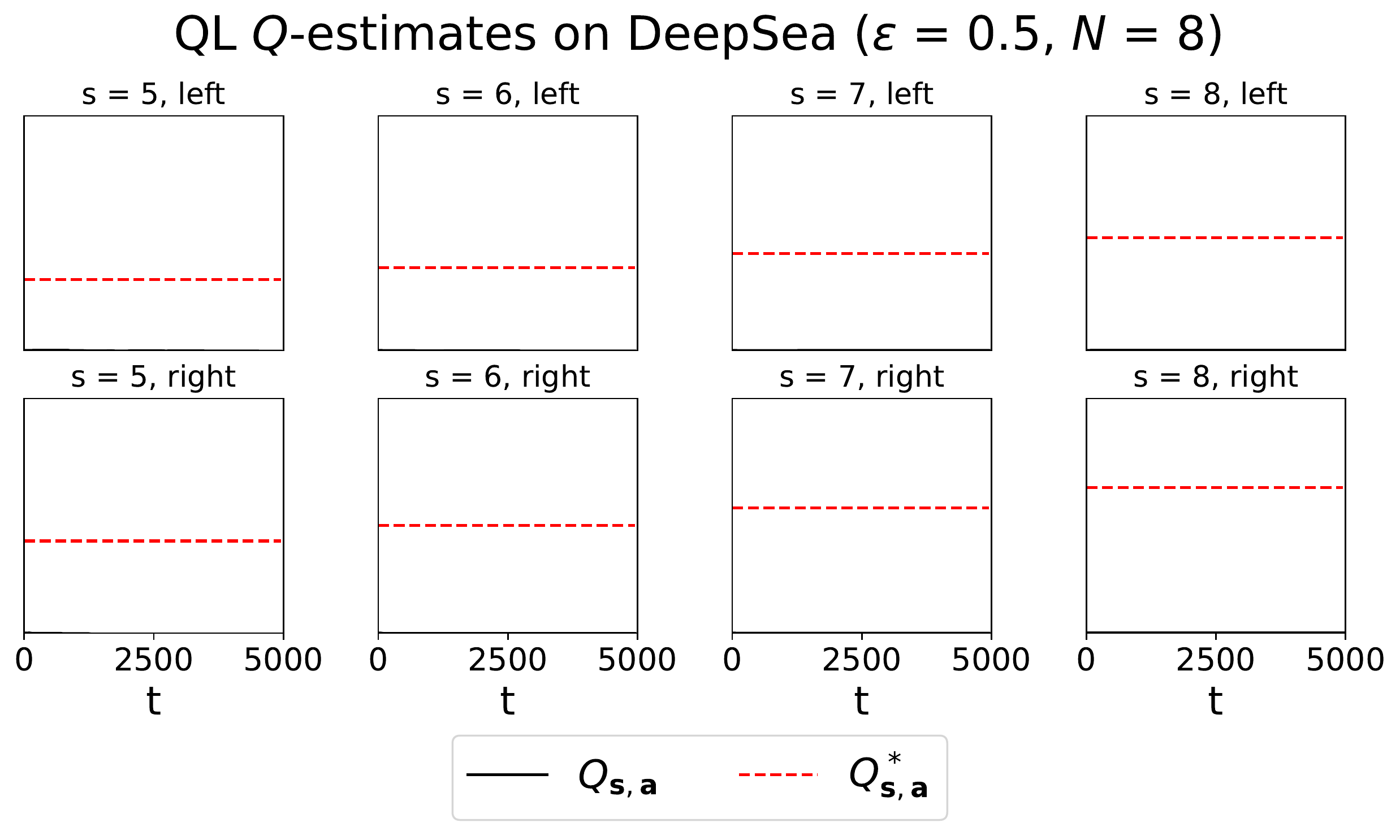}}
        \caption{Comparison of {\algo} and $Q$-learning on DeepSea Exploration with sparsity $N = 8$. This plot shows the concentration of the predicted $\hat{Q}$ to the true $Q^*$ (ground-truth) with iterations. \textbf{Remark:} As the sparsity is increased, the performance of Q-learning degrades drastically (potentially due to lack of efficient exploration) whereas 
        {\algo} converges to true $Q^*$ efficiently. Further, we note that since right actions are optimal for DeepSea environment, {\algo} stops exploring left actions beyond a point leading to a comparatively higher variance in $\hat{Q}$ for left actions, thus providing directed exploration.}
        \label{fig:sids_s4}
     \end{figure}
        \begin{figure}[t]
	  \centering
       \subfigure[]{  \includegraphics[width=0.49\textwidth]{figures_new/steering_4.pdf}   }      %
       \subfigure[]{\includegraphics[width=0.49\textwidth]{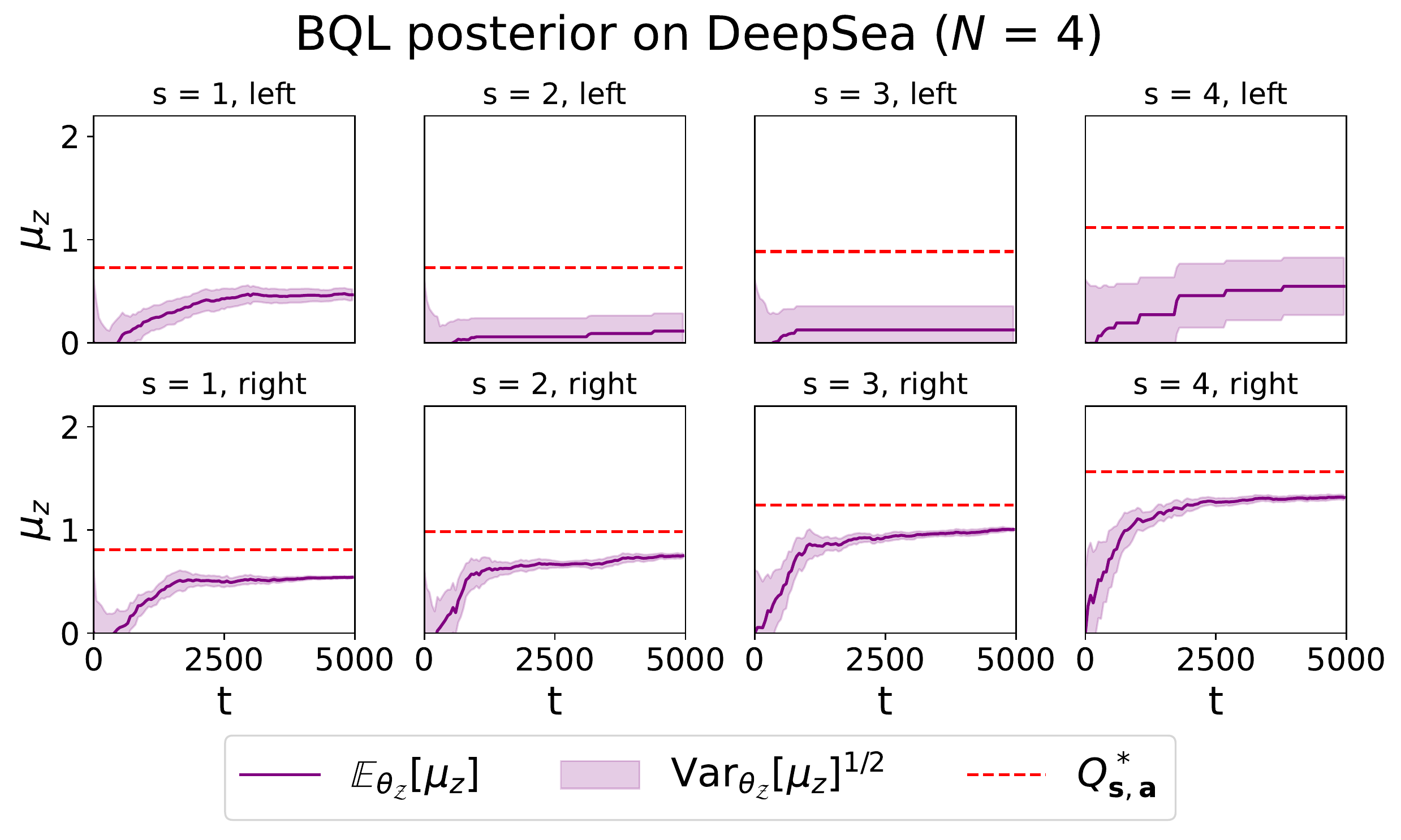}}
        \caption{Comparison of {\algo} and Bayesian-$Q$ learning (BQL) on DeepSea Exploration with sparsity $N = 4$. This plot shows the concentration of the predicted $\hat{Q}$ to the true $Q^*$ (ground-truth) versus iterations. \textbf{Remark:} {\algo} converges to true $Q^*$ much faster than Bayesian-$Q$ learning which fails to concentrate on the true $Q^*$. This is because BQL does not have an efficient forgetting mechanism in its update rule leading to high dependence on inaccurate past updates. This leads to the observation that the posterior is overconfident about incorrect predictions in BQL. However, {\algo} stops exploring left actions beyond a point as right actions are optimal for DeepSea environment, leading to a comparatively higher variance in $\hat{Q}$ for left actions, but low variance for right actions, thus providing directed exploration.}
        \label{fig:sids_s5}
     \end{figure}
     \begin{figure}[t]
	  \centering
        \subfigure[]{\includegraphics[width=0.49\textwidth]{new_conv/Steering-0_0-4_0-3_0-3_0-posterior-deepsea-8_2.pdf} }       
      \subfigure[]{  \includegraphics[width=0.49\textwidth]{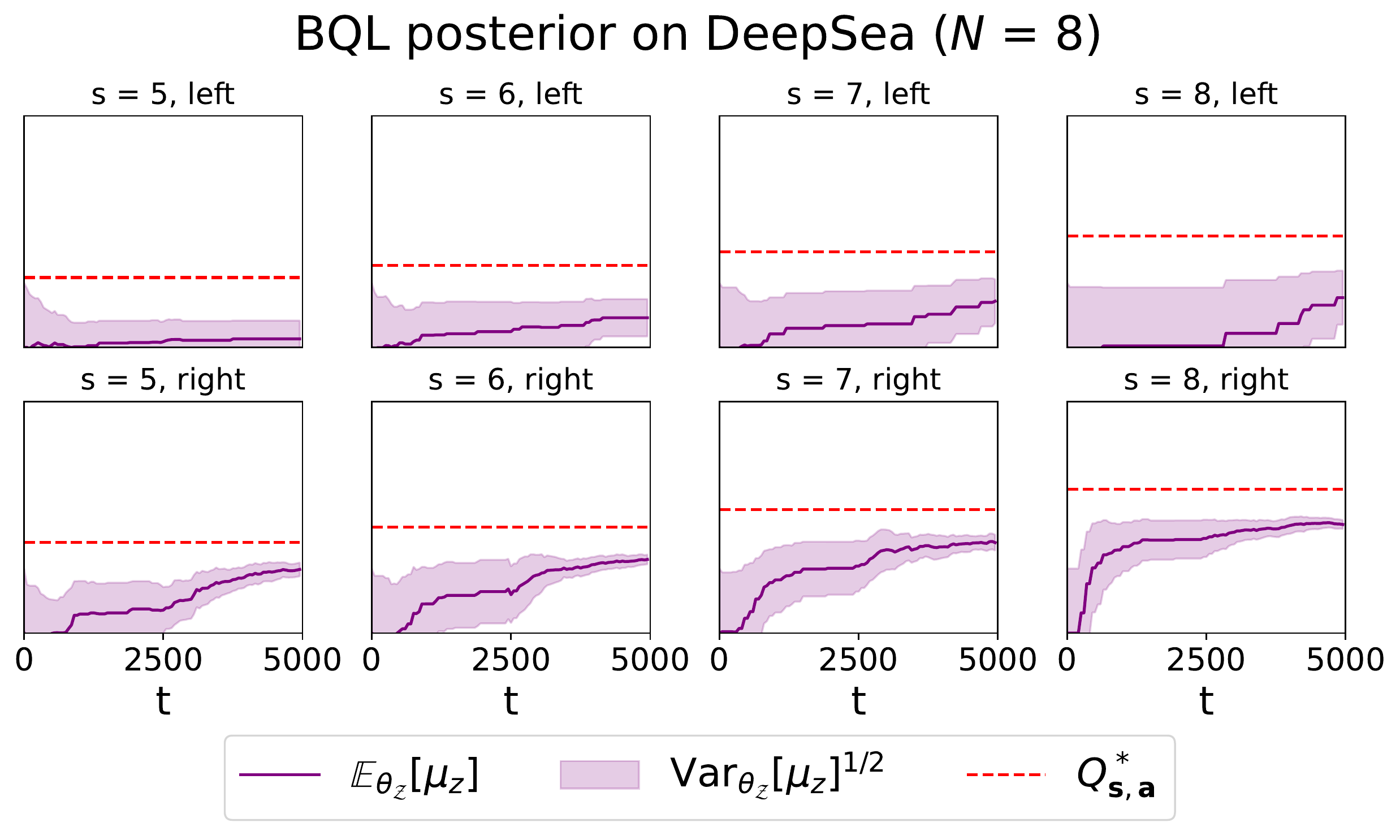}}
        \caption{Comparison of {\algo} and Bayesian-$Q$ learning (BQL)  on DeepSea Exploration with sparsity $N = 8$. This plot shows the concentration of the predicted $\hat{Q}$ to the true $Q^*$ (ground-truth) versus iteration with more sparsity. \textbf{Remark:} {\algo} converges to true $Q^*$ much faster than Bayesian-$Q$ learning which fails to concentrate on the true $Q^*$. A major reason can be that BQL doesn't have an efficient forgetting mechanism in its update rule leading to high dependence on inaccurate past updates. Hence, we observe that the posterior for BQL is overconfident about incorrect predictions. In contrast, {\algo} stops exploring left actions beyond a point as right actions are optimal for DeepSea Exploration environment, leading to a comparatively higher variance in $\hat{Q}$ for left actions, thus providing directed exploration.}
        \label{fig:sids_s6}
     \end{figure}
             \begin{figure}[t]
	  \centering
   \subfigure[]{        \includegraphics[width=0.48\textwidth]{figures_new/steering_4.pdf}  }       %
    \subfigure[]{ \includegraphics[width=0.48\textwidth]{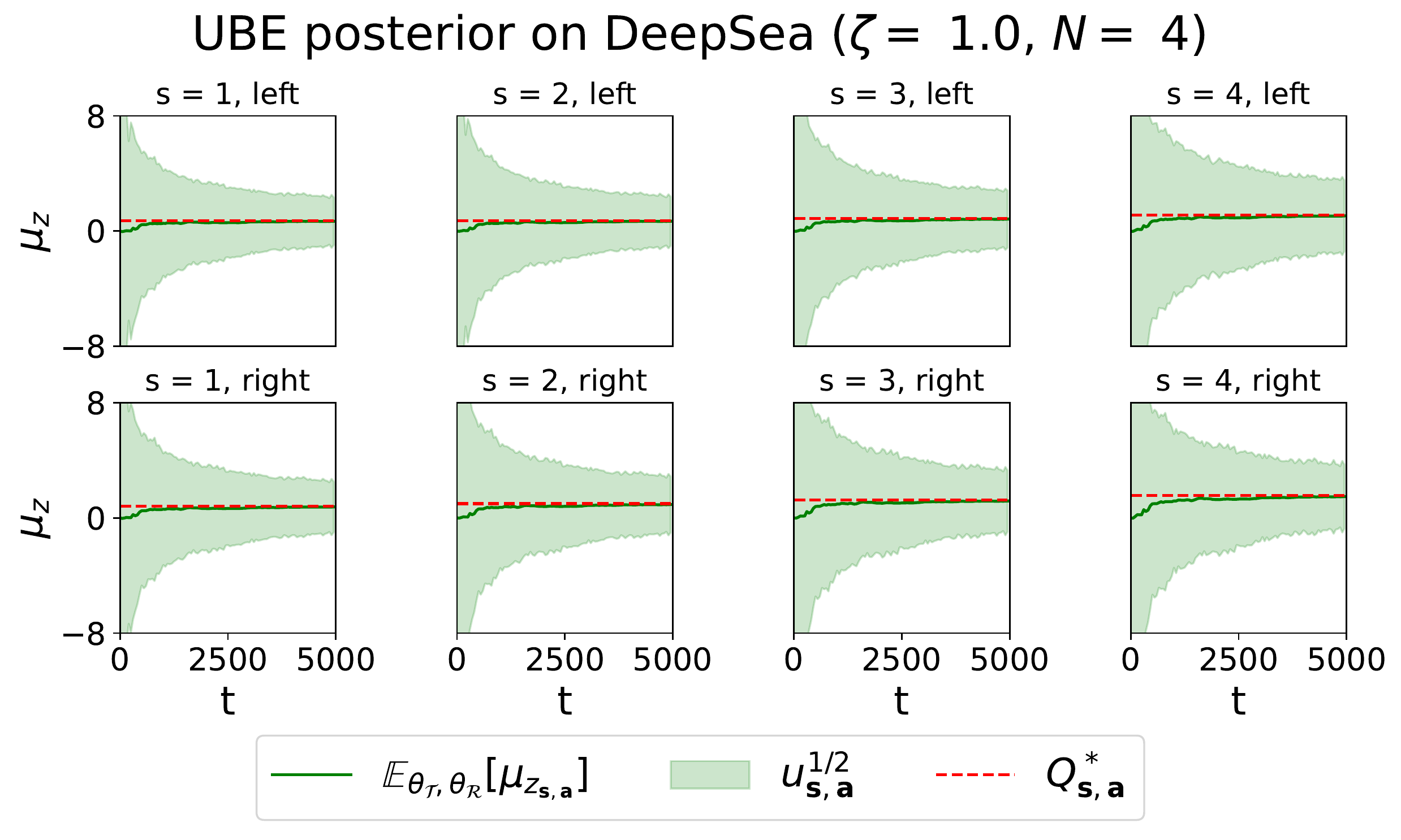}       } 
        \caption{Comparison of {\algo} and uncertainty Bellman equation (UBE) on DeepSea Exploration with sparsity $N = 4$. This plot shows the concentration of the predicted $\hat{Q}$ to the true $Q^*$ (ground-truth) versus iterations. \textbf{Remark:} {\algo} converges much more efficiently to true $Q^*$ (or $\mu_z^*$) compared to UBE which fails to concentrate properly. For UBE, even if the predicted mean is closer to optima, the variance is too high which leads to sub-optimal exploration. In contrast, {\algo} stops exploring left actions beyond a point as right actions are optimal for the DeepSea Exploration environment, leading to a comparatively higher variance in $\hat{Q}$ for left actions, thus providing directed exploration.}
        \label{fig:sids_s7}
     \end{figure}
     
     \begin{figure}[t]
	  \centering
      \subfigure[]{  \includegraphics[width=0.48\textwidth]{new_conv/Steering-0_0-4_0-3_0-3_0-posterior-deepsea-8_2.pdf}    }   
      \subfigure[]{  \includegraphics[width=0.48\textwidth]{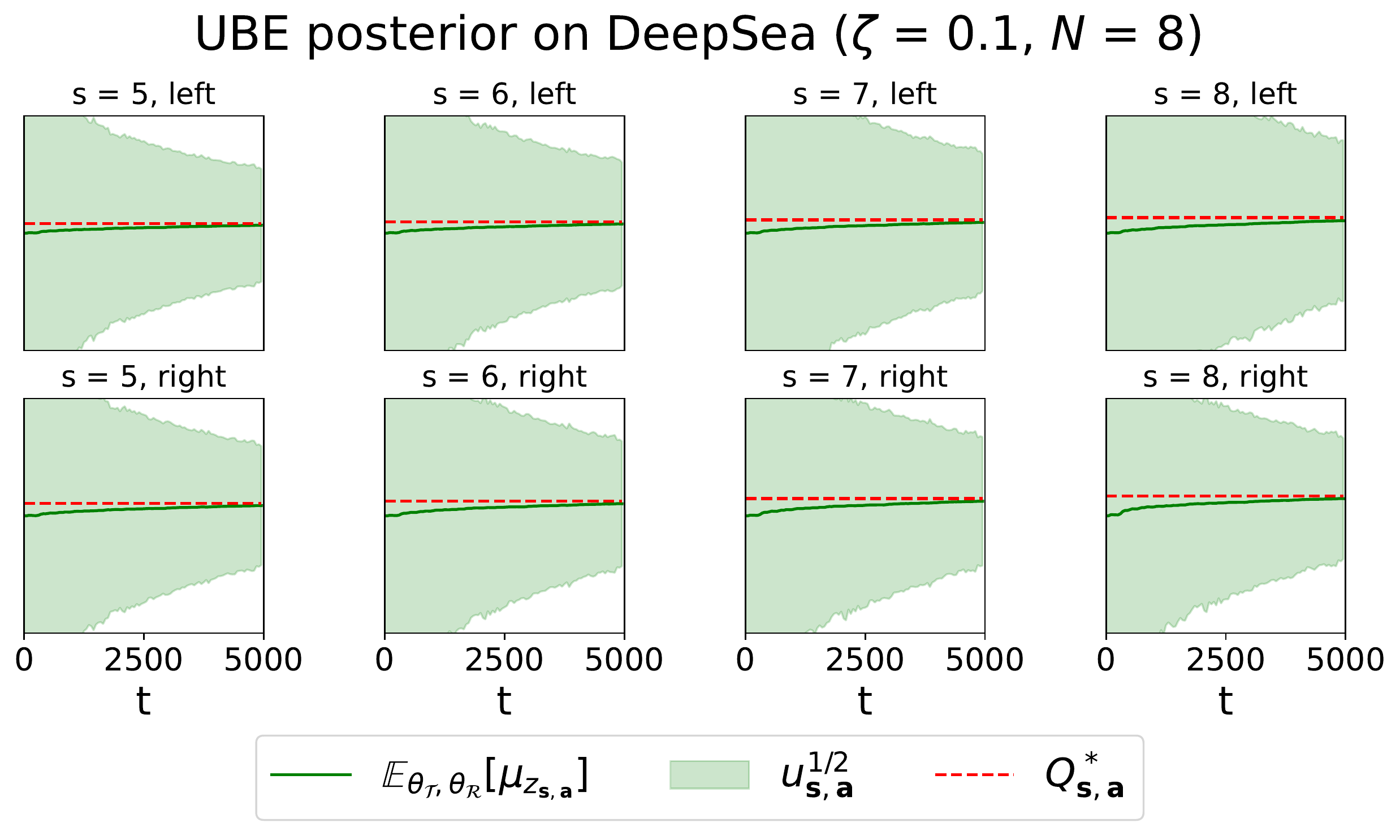}}
        \caption{Comparison of {\algo} and uncertainty Bellman equation (UBE) on DeepSea Exploration with sparsity $N = 8$. This plot analyzes the concentration of the predicted $\hat{Q}$ to the true $Q^*$ (ground-truth) versus iterations with more sparsity. \textbf{Remark:} As the sparsity is increased ($N = 8$), the variance in estimated $\mu_z^*$ for UBE increases significantly and the performance degrades drastically leading to random action selection. In contrast, {\algo} converges much more efficiently to true $Q^*$ (or $\mu_z^*$). Also, {\algo} stops exploring left actions beyond a point as right actions are optimal for DeepSea Exploration environment, leading to a comparatively higher variance in $\hat{Q}$ for left actions, thus providing directed exploration.}
        \label{fig:sids_s8}
     \end{figure}

             \begin{figure}[t]
	  \centering
      \subfigure[]{  \includegraphics[width=0.48\textwidth]{figures_new/steering_4.pdf} }      
   \subfigure[]{     \includegraphics[width=0.48\textwidth]{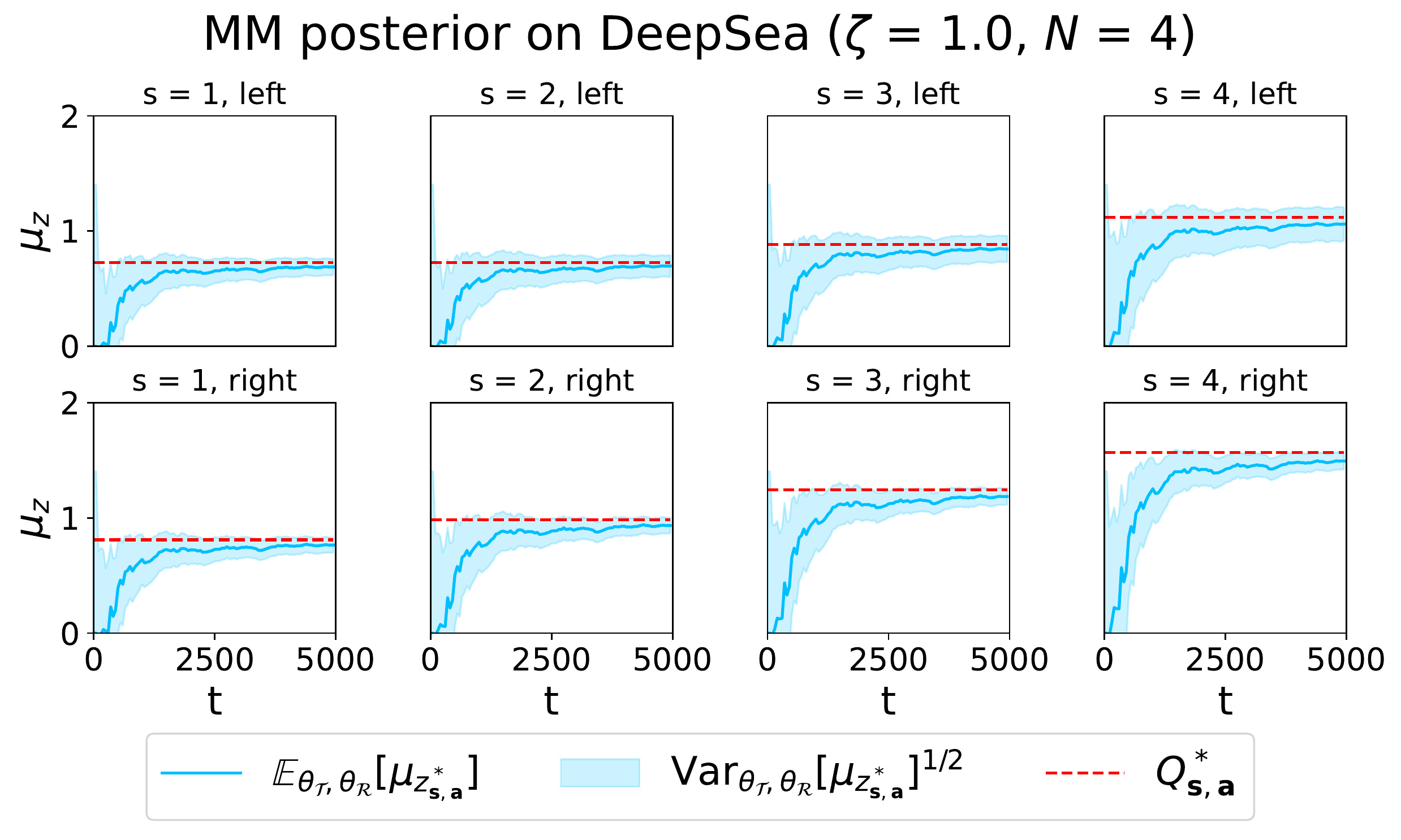}}
        \caption{Comparison of {\algo} and moment matching (MM)  on DeepSea Exploration with sparsity $N = 4$. This plot analyzes the concentration of the predicted $\hat{Q}$ to the true $Q^*$ (or $\mu_z^*$) versus iterations. \textbf{Remark:} Although MM performs well in this setting, {\algo} converges more efficiently and faster to true $Q^*$. Also, {\algo} unlike MM stops exploring left actions beyond a point as right actions are optimal for DeepSea Exploration environment, leading to a comparatively higher variance in $\hat{Q}$ for left actions and low variance for right actions, thus providing directed exploration.}
        \label{fig:sids_s9}
     \end{figure}
     \begin{figure}[t]
	  \centering
       \subfigure[]{ \includegraphics[width=0.46\textwidth]{new_conv/Steering-0_0-4_0-3_0-3_0-posterior-deepsea-8_2.pdf}}    
     \subfigure[]{   \includegraphics[width=0.46\textwidth]{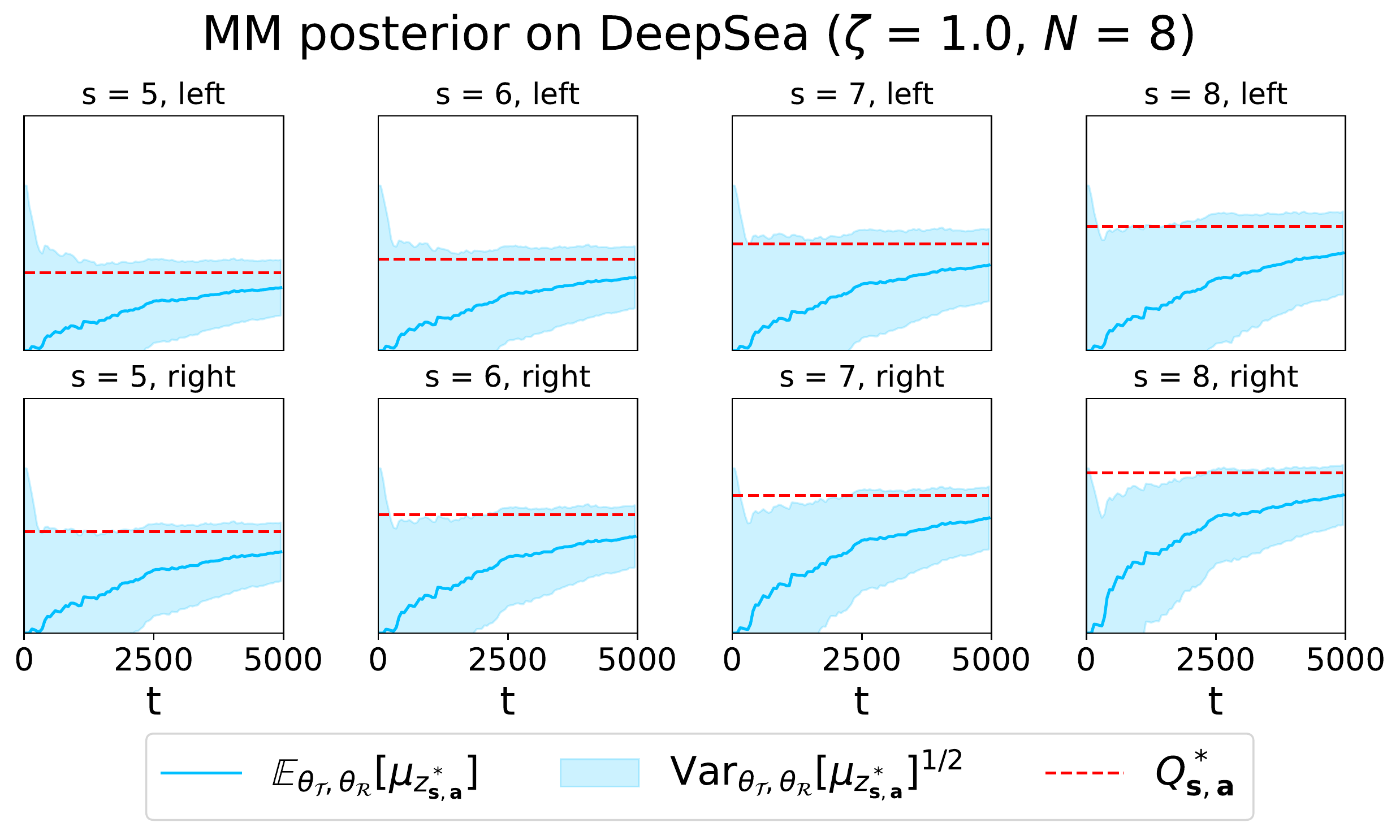}}
        \caption{Comparison of {\algo} and moment matching  on DeepSea Exploration with sparsity $N = 8$. This plot analyzes the concentration of the predicted $\hat{Q}$ to the true $Q^*$ (or $\mu_z^*$) (ground-truth) versus iterations. \textbf{Remark:} As the sparsity is increased ($N = 8$), the performance of MM degrades with increased variance of predicted $\mu_z$ and also converges to sub-optimal $mu_z$. While {\algo} converges to true $Q^*$  efficiently. Also, since right actions are optimal for DeepSea Exploration environment, {\algo} stops exploring left actions beyond a point leading to a comparatively higher variance in $\hat{Q}$ for left actions, thus providing directed exploration.
        }
        \label{fig:sids_s10}
     \end{figure}
     \begin{figure}[t]
	  \centering
       \subfigure[]{ \includegraphics[width=0.46\textwidth]{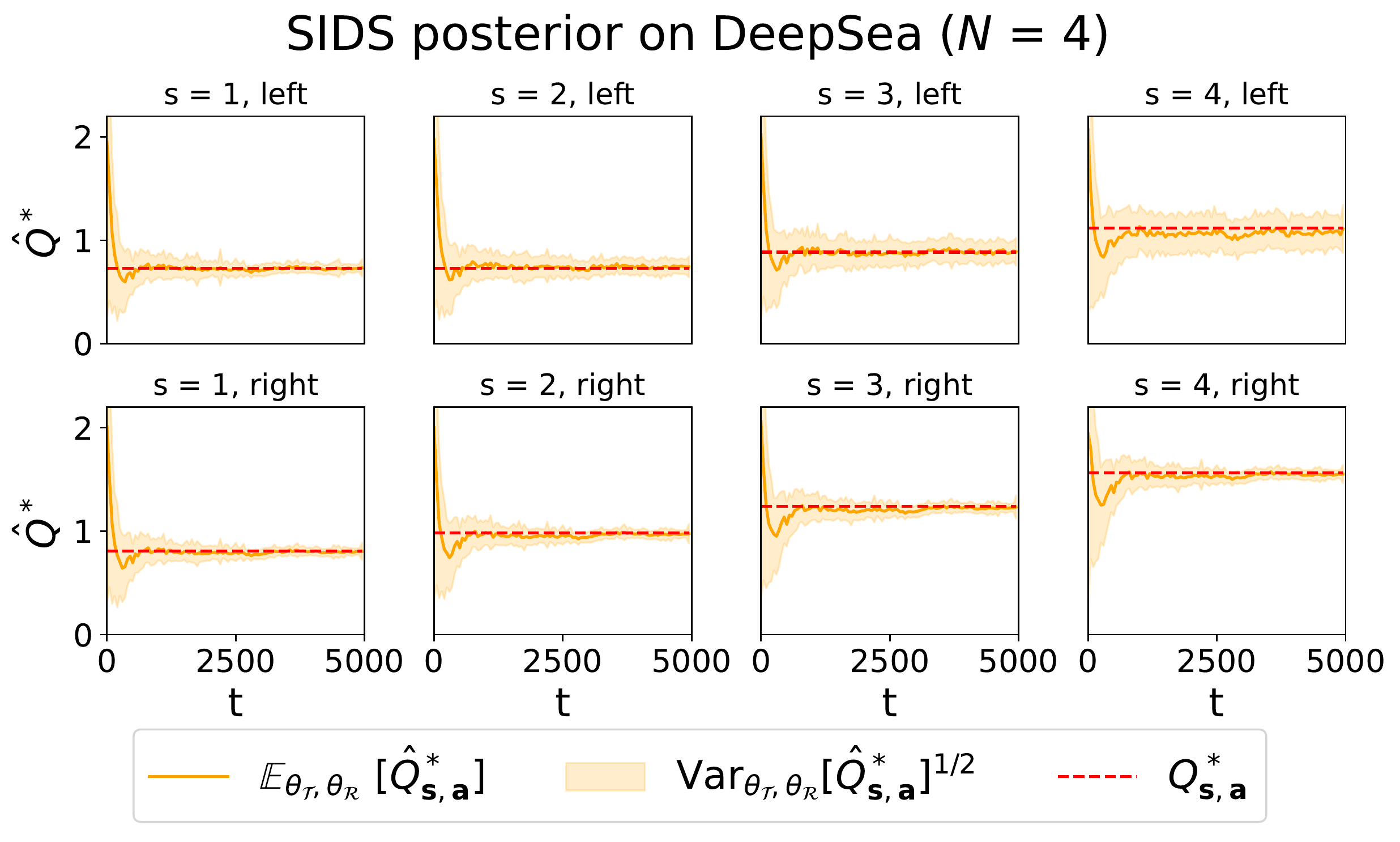}}         
       \subfigure[]{\includegraphics[width=0.46\textwidth]{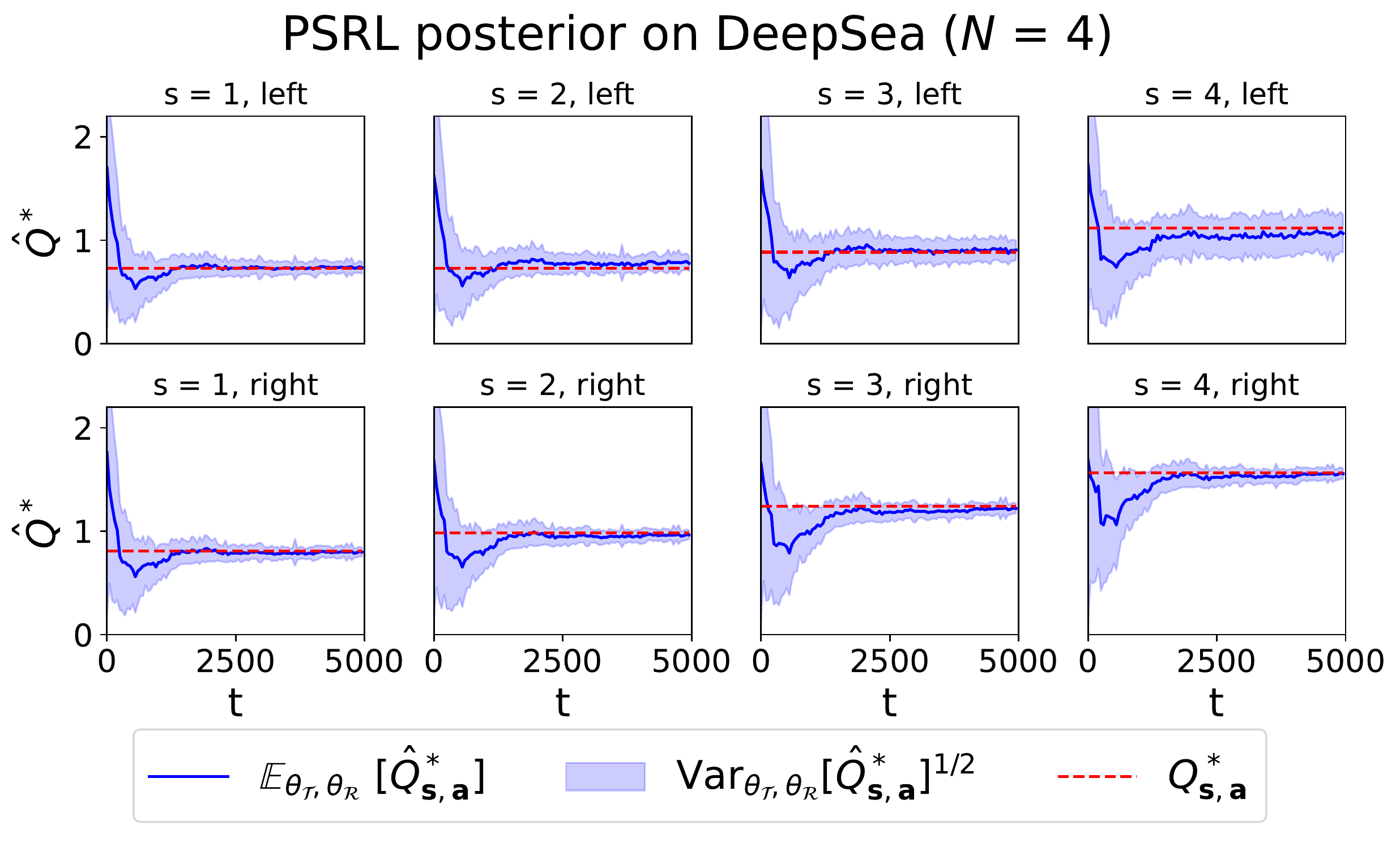}}
        \caption{Comparison of {\algo} and posterior sampling reinforcement learning (PSRL) on DeepSea Exploration with sparsity $N = 4$. This plot analyzes the concentration of the predicted $\hat{Q}$ to the true $Q^*$ versus iterations. \textbf{Remark:} The performance of PSRL is comparable to {\algo}, but still the convergence is faster with lesser variance for {\algo}. We also note that the sparsity level for this setting is low ($N=4$).}
        \label{fig:sids_s1}
     \end{figure}
     \begin{figure}[t]
	  \centering
        \subfigure[]{\includegraphics[width=0.46\textwidth]{new_conv/Steering-0_0-4_0-3_0-3_0-posterior-deepsea-8_2.pdf}}        
        \subfigure[]{\includegraphics[width=0.46\textwidth]{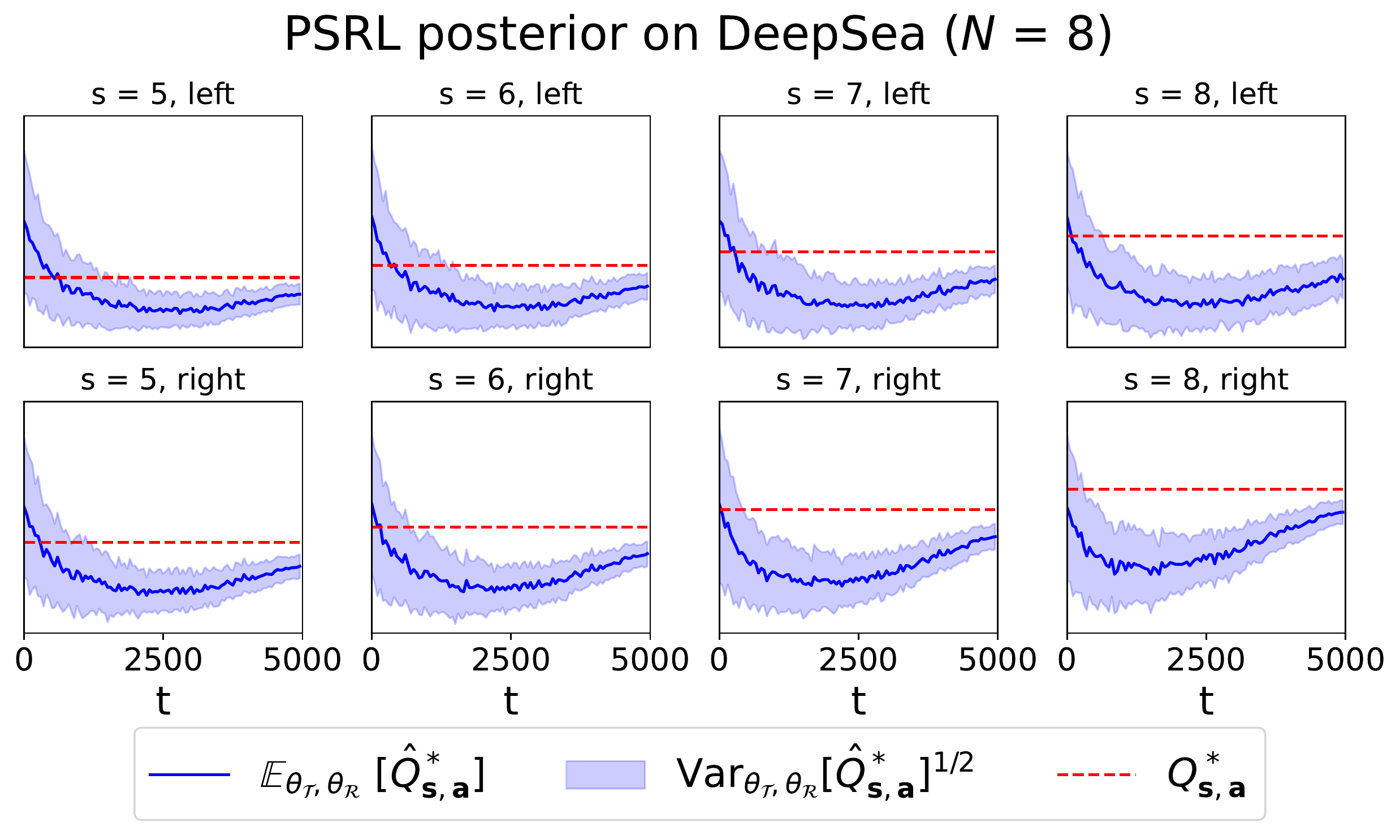}}
        \caption{Comparison of {\algo} and posterior sampling reinforcement learning (PSRL)  on DeepSea Exploration with sparsity $N = 8$. This plot analyzes the concentration of the predicted $\hat{Q}$ to the true $Q^*$ (ground-truth) versus iterations. \textbf{Remark:} As the sparsity is increased ($N = 8$), the performance of PSRL degrades with slower convergence to $Q^*$ with higher variance in $\hat{Q}$ prediction. In contrast, {\algo} converges much more efficiently to true $Q^*$. Also, {\algo} stops exploring left actions beyond a point as right actions are optimal for DeepSea Exploration environment, leading to a comparatively higher variance in $\hat{Q}$ for left actions, thus providing directed exploration.}
        \label{fig:sids_s2}
     \end{figure}

   \clearpage
\subsection{Convergence Plots for DSD}    
Finally, we show the correctness of our approach by observing the convergence of DSD in Fig.~\ref{fig:ksd_convergence}. We note that for the optimal actions (orange), the DSD converges to much lower values than sub-optimal actions (blue) which in-turn implies the effectiveness of our Stein-information based proposed approach.    
            
\begin{figure}[h]
     \begin{center}
     \hfil
\subfigure[]{\includegraphics[width=.245\columnwidth,clip = true]{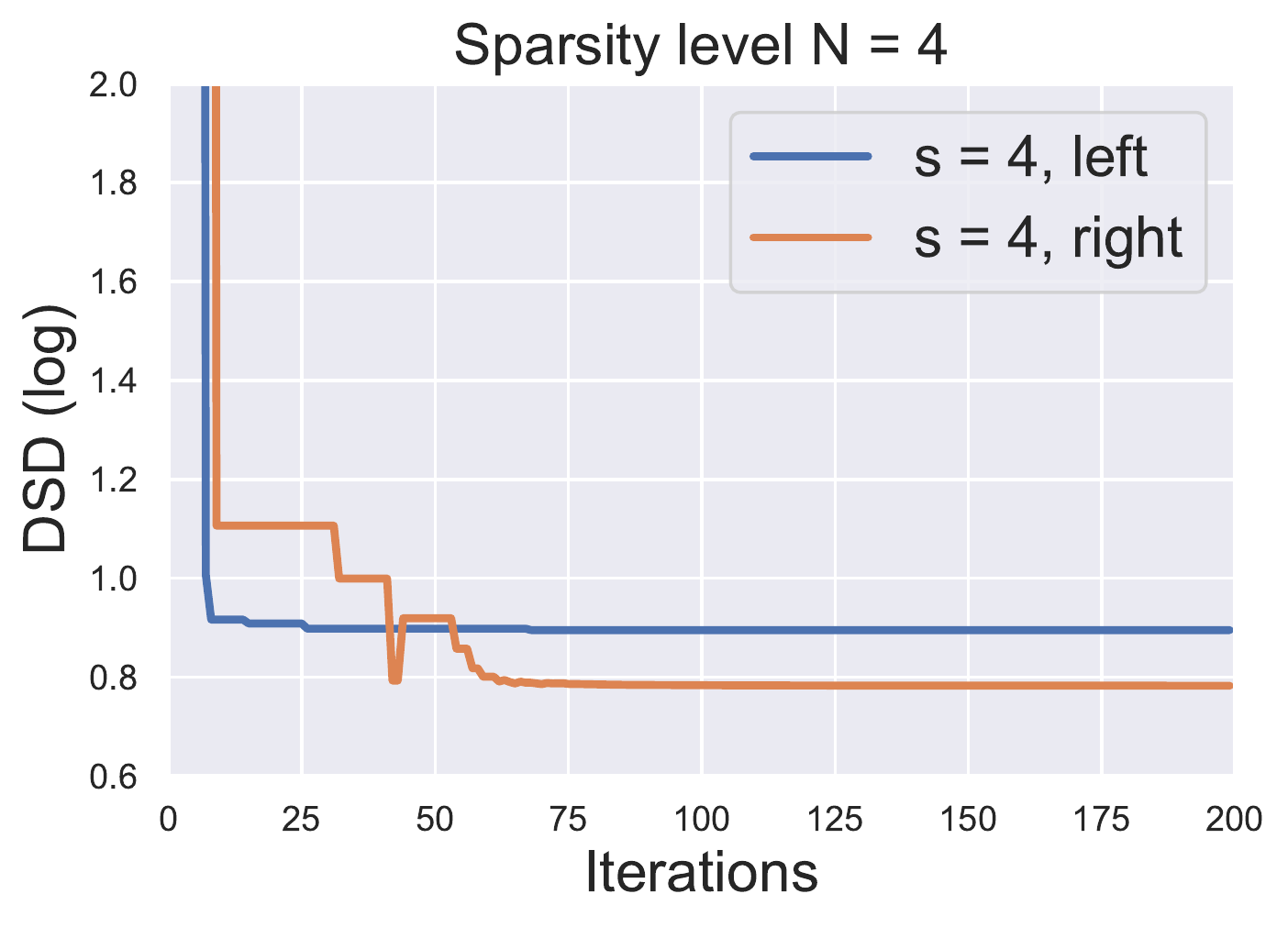}         \label{fig:4_s1}}
\hfil
     \subfigure[]{
\includegraphics[width=.245\columnwidth,clip = true]{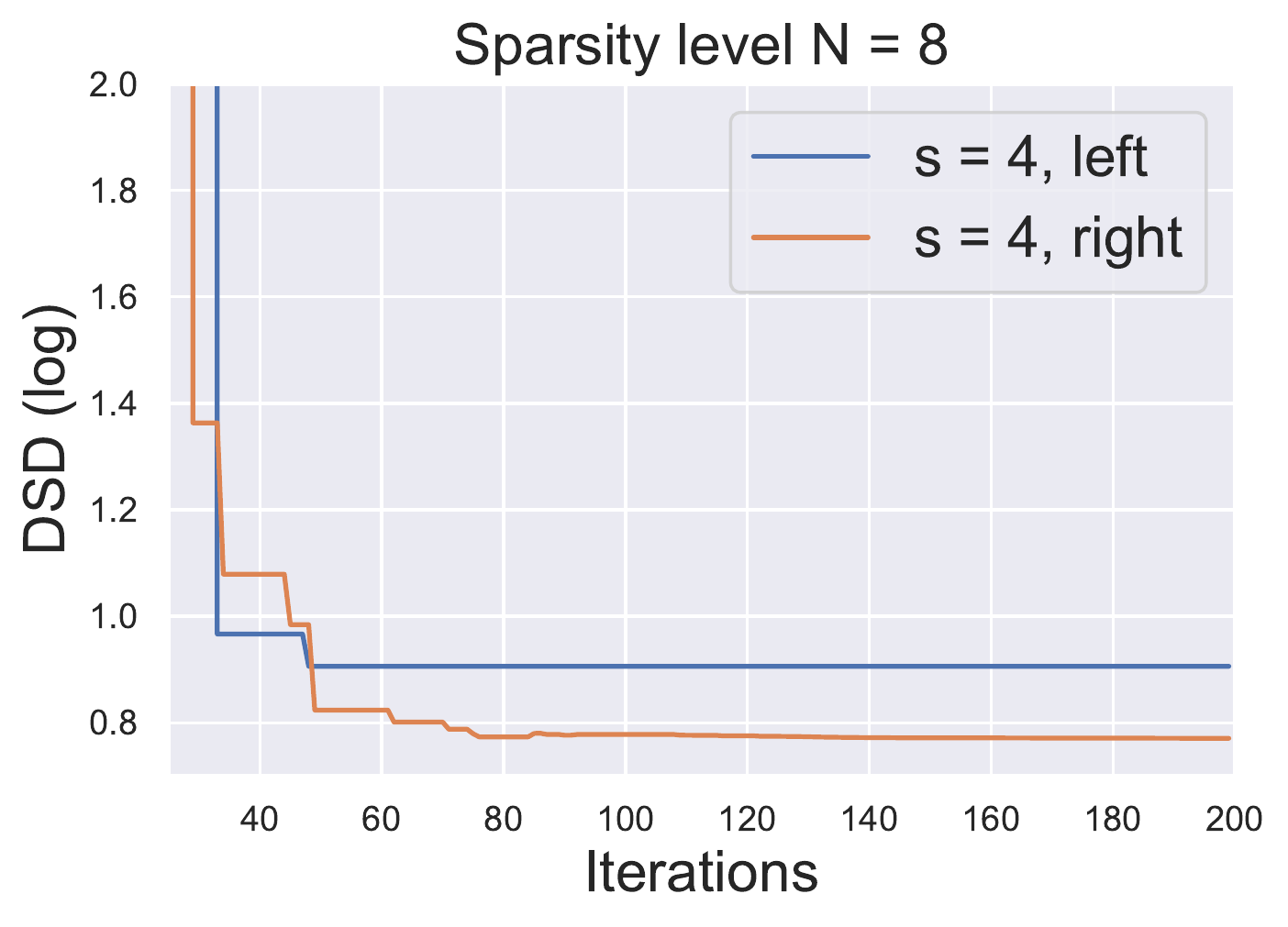}\label{fig:4_s2}}\\
         \hfil
     \subfigure[]{\includegraphics[width=.245\columnwidth,clip = true]{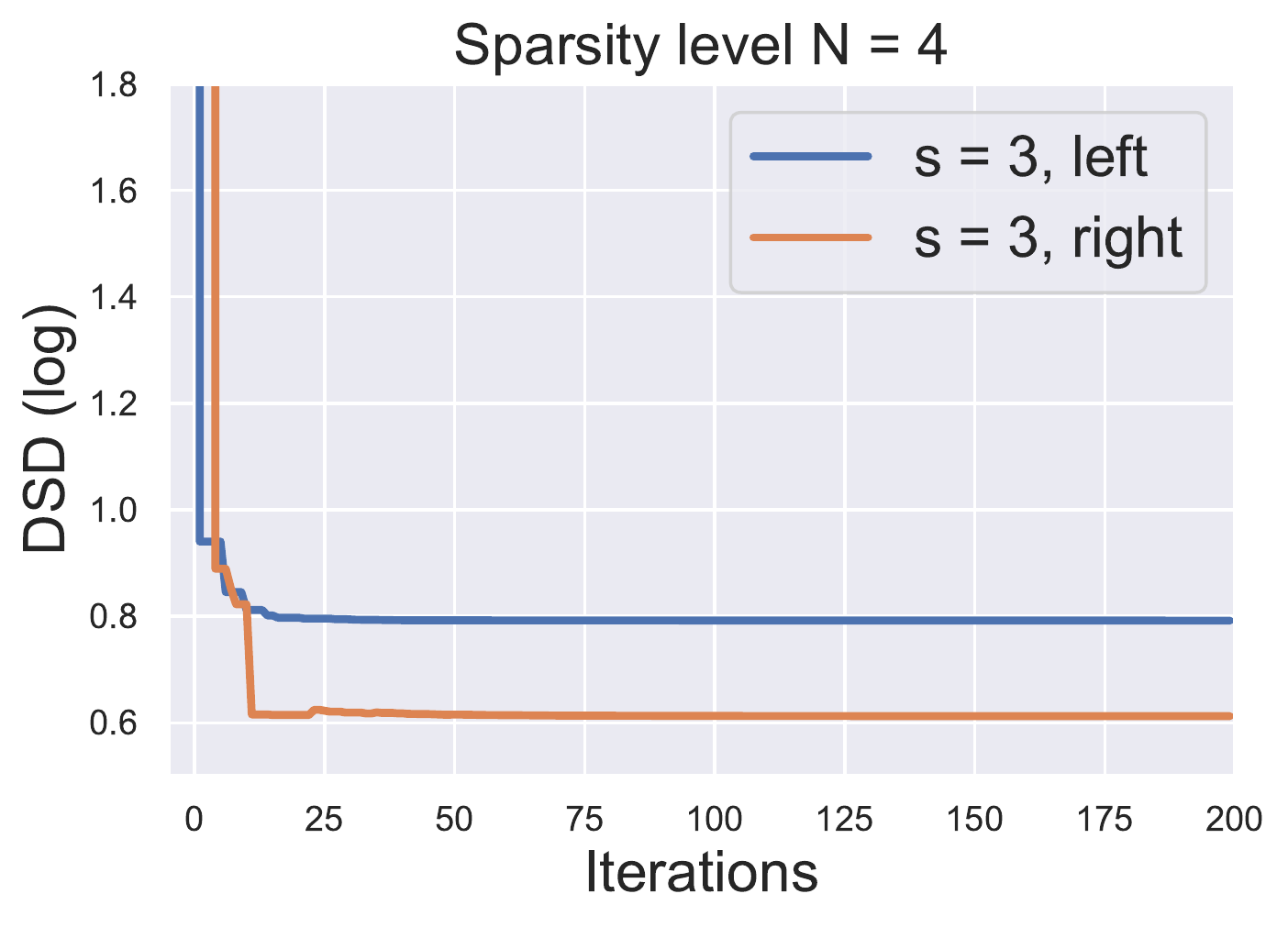}
         \label{fig:3_s1}}
         \hfil
     \subfigure[]{\includegraphics[width=.245\columnwidth,clip = true]{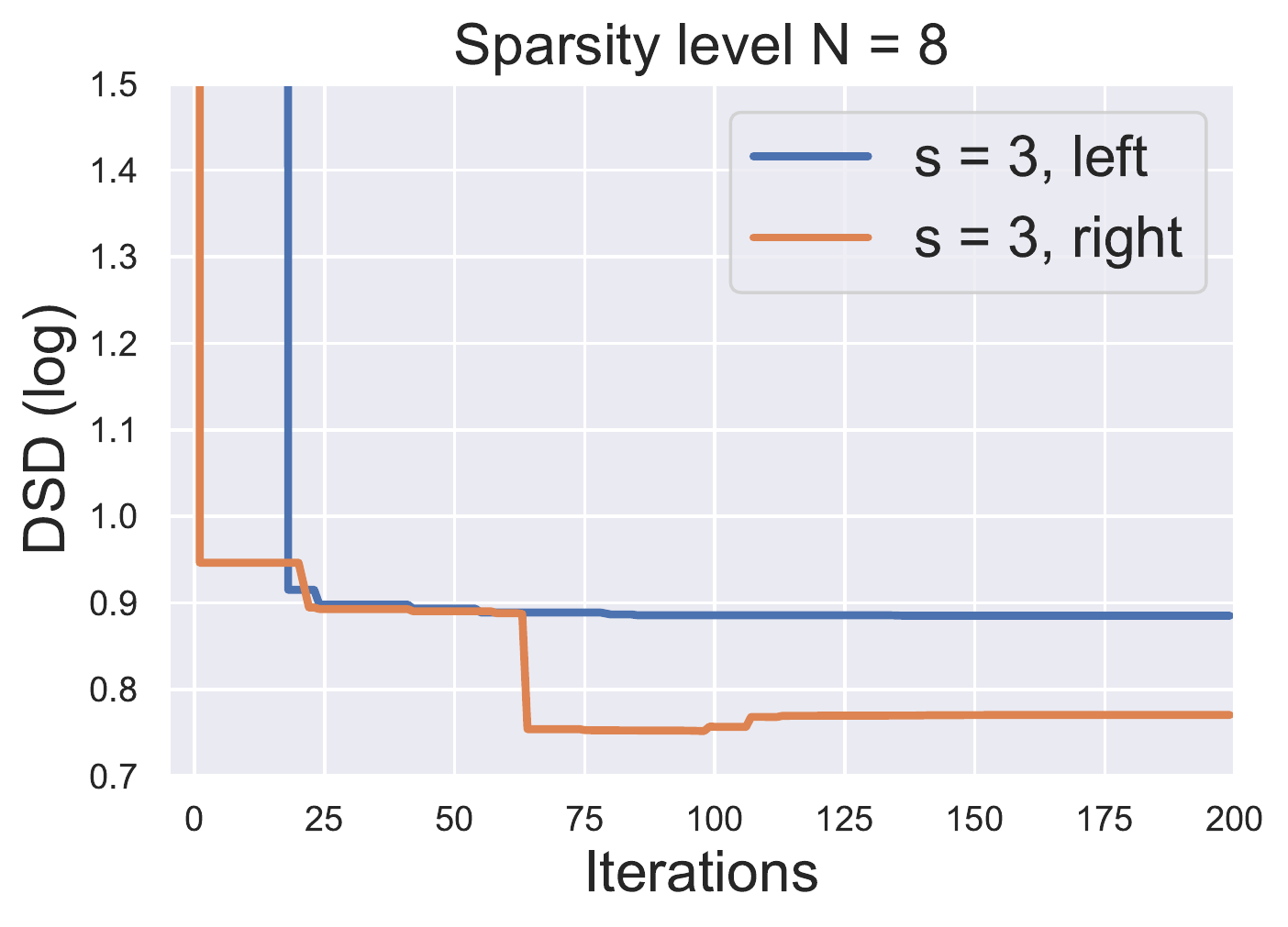}              \label{fig:3_s2}}
              \hfil
              \end{center}
        \caption{This figure provides evidence of DSD convergence (mean plot over 5 seeds) for different state-action pairs and sparsity levels with iterations for the DeepSea environment. Additionally, the plot also provides an indication of directed exploration through the DSD convergence to lower values for right actions which moves agent towards goal than left in states $s = 3$ and $s = 4$.}
        \label{fig:ksd_convergence}
\end{figure}

\subsection{Additional Comparisons for WideNarrow MDP and PriorMDP}
Here in Fig. \ref{final_figure_sota_new_envs},  we also compares the performance of {\algo} on WideNarrow MDP and PriorMDP environments \citep{markou_mm, osband17a} against the existing RL baselines : vanilla Q-learning with $\epsilon-$greedy action selection \citep{Watkins1992}, Bayesian Q-learning (BQL) \citep{dearden_bql}, Uncertainty Bellman Equation (UBE) \citep{donoghue_ube}, Moment matching (MM) across Bellman equation \citep{markou_mm}, Posterior Sampling RL (PSRL) \citep{osband2013}, and IDS \citep{hao2022regret}. We approximated information gain with variance to implement IDS, hence denoted by Var-IDS in Fig. \ref{final_figure_sota_new_envs}. %

\begin{figure}[ht]
                \centering
                \hfil
            \subfigure[WideNarrow MDP]{\includegraphics[width=0.29\columnwidth,clip = true]{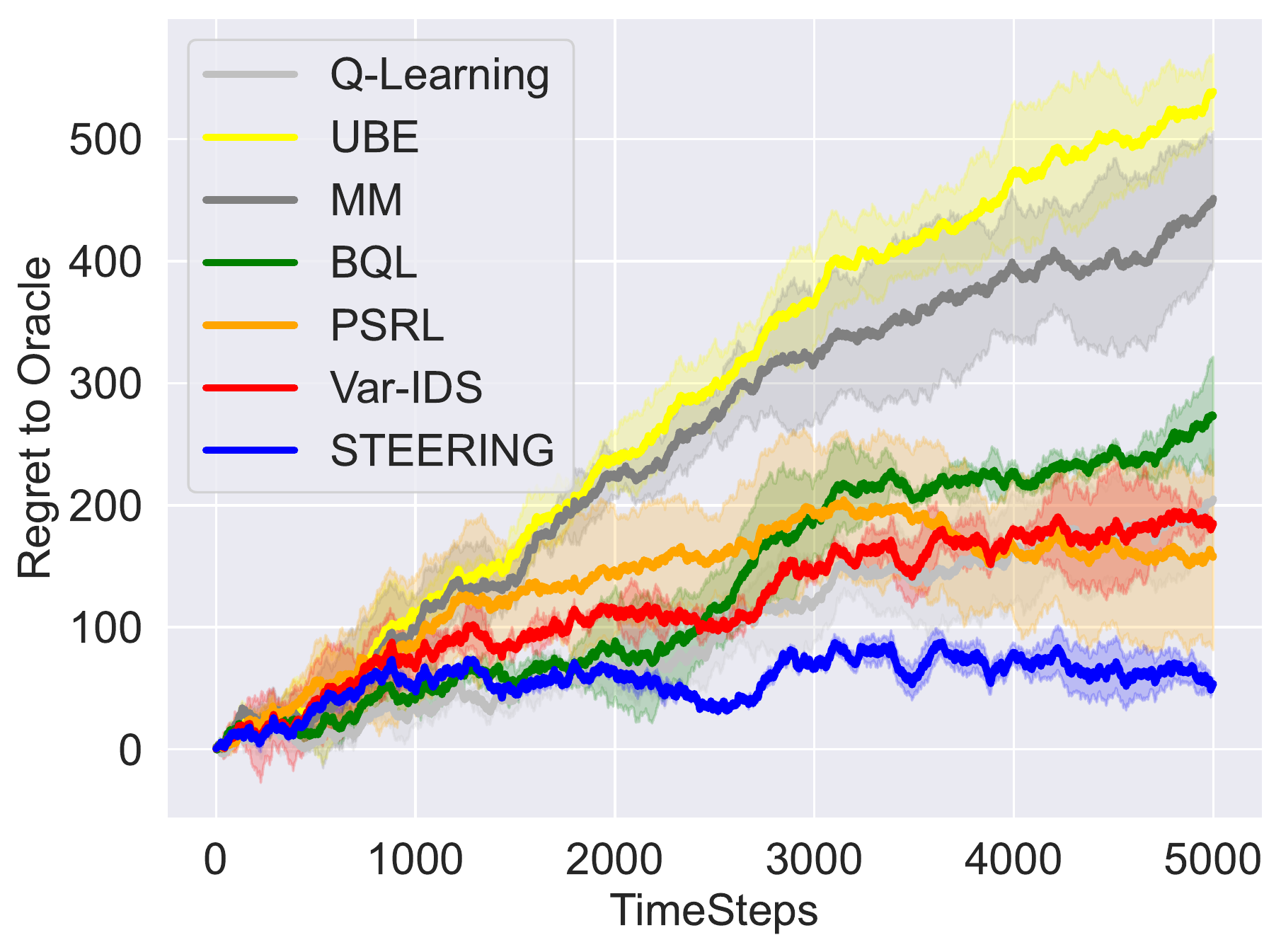}}
            \hfil
        \subfigure[PriorMDP]{\includegraphics[width=0.29\columnwidth,clip = true]{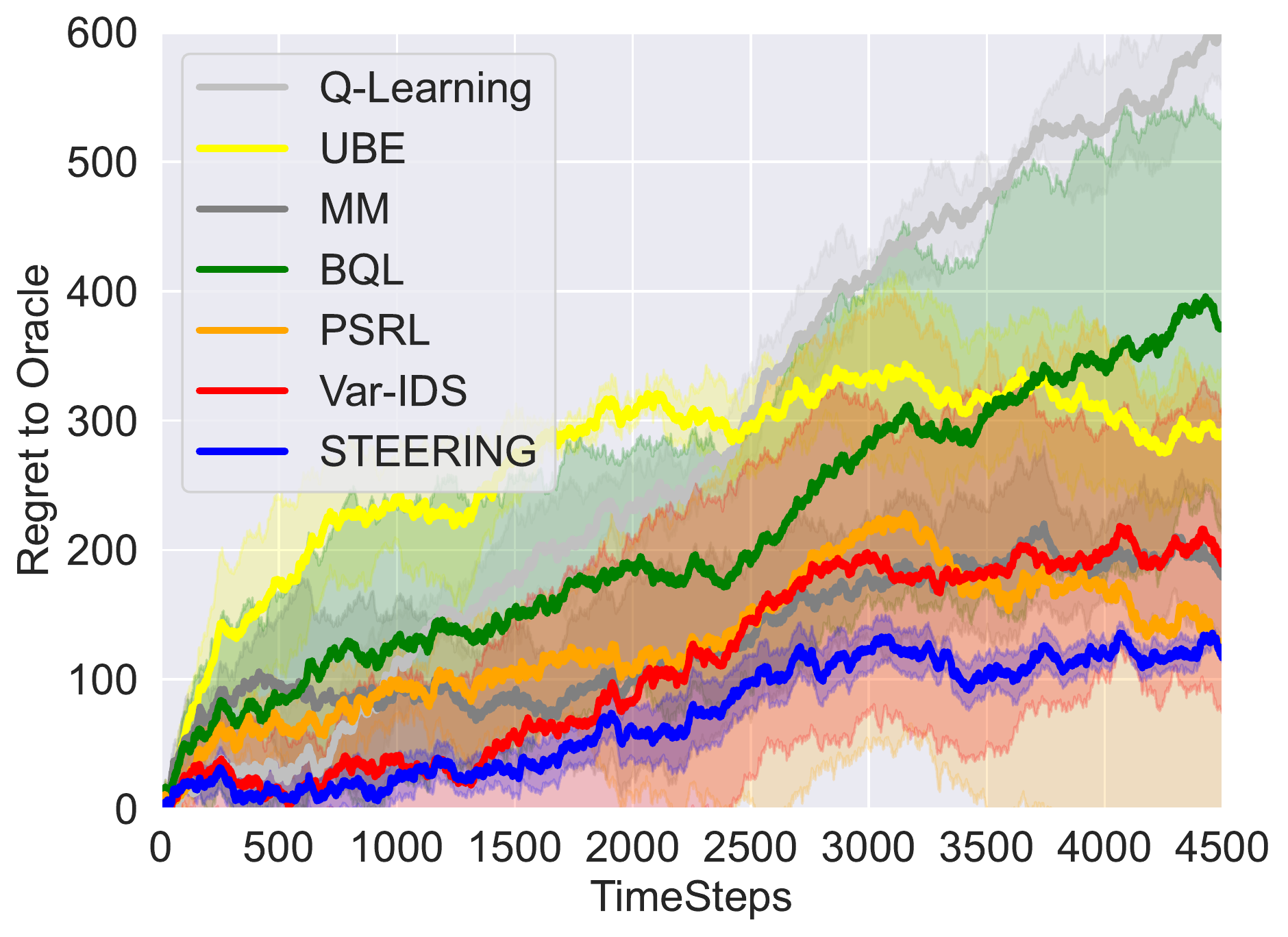}}
        \hfil
            \caption{This figure compares the performance of {\algo} on  (a) WideNarrow MDP and (b) PriorMDP environments \citep{markou_mm, osband17a}. \textbf{Remark:} WideNarrow MDP tests the algorithm's ability under factored posterior approximations, whereas PriorMDP tests the algorithm's ability to more general environments without specific structures. We note  that 
            {\algo} outperforms existing baselines in both the environments.}
            \label{final_figure_sota_new_envs}
 \end{figure}

\end{document}